%% file: 0-paper.tex
\documentclass[conference]{IEEEtran}
\usepackage{graphicx}
\usepackage{xspace}
\usepackage{color}
\usepackage{paralist}
\usepackage{wrapfig}
\usepackage{multirow}
\usepackage{listings}
\usepackage{makecell}
\usepackage{boxedminipage}
\usepackage{booktabs}
\usepackage{subcaption}
\usepackage{caption}
\usepackage{multirow}
\usepackage{multicol}
\usepackage{diagbox}
\usepackage{adjustbox}
\usepackage{url}
\usepackage{cite} 
\usepackage{tabularx}
\usepackage{pifont}
\usepackage{textcomp}
\usepackage{balance} 
\usepackage{upquote}
\usepackage[T1]{fontenc}
\usepackage{algorithm}
\usepackage{algpseudocode}

\usepackage{pifont}
\usepackage{balance}
\usepackage[htt]{hyphenat}
\usepackage{mdframed}
\usepackage{amsmath}
\usepackage{amssymb}
\usepackage[table,xcdraw]{xcolor} 
\usepackage[colorlinks=true,linkcolor=blue,citecolor=red,urlcolor=blue]{hyperref} 

\usepackage[inline]{enumitem}
\newcommand{\coloneqq}{\mathrel{\mathop:}=}
\usepackage{tcolorbox}

\tcbset{
    colback=gray!10, 
    colframe=black,  
    arc=3mm,         
    boxrule=0.5pt,   
    width=\linewidth-2mm, 
    left=3pt, right=3pt, top=3pt, bottom=3pt, 
}

\makeatletter

\newcommand{\refappendix}[1]{\hyperref[#1]{Appendix~\ref*{#1}}}

\input{macro}

\newcommand{\topone}[1]{\cellcolor{green!50}\textbf{#1}} 
\newcommand{\default}[1]{\cellcolor{green!10}#1} 

\newcommand{\mpg}{model poisoning\xspace}
\newcommand{\mpa}{model poisoning attack\xspace}

\newcommand{\umpas}{untargeted model poisoning attacks\xspace}
\newcommand{\mpas}{model poisoning attacks\xspace}

\newcommand{\MPAs}{Model Poisoning Attacks\xspace}

\newcommand{\FG}{GeminiGuard\xspace}
\newcommand{\GG}{GeminiGuard\xspace}
\newcommand{\NIID}{non-iid\xspace}
\newcommand{\NIIDs}{non-iids\xspace}

\begin{document}

\title{Two Heads Are Better than One: Model-Weight and Latent-Space Analysis for Federated Learning on Non-iid Data against Poisoning Attacks}

\author{
    \IEEEauthorblockN{
          {Xingyu Lyu\IEEEauthorrefmark{1}},
          {Ning Wang\IEEEauthorrefmark{2}},
          {Yang Xiao\IEEEauthorrefmark{3}},
          {Shixiong Li\IEEEauthorrefmark{1}},
          {Tao Li\IEEEauthorrefmark{4}},
          {Danjue Chen\IEEEauthorrefmark{5}},
          {Yimin Chen\IEEEauthorrefmark{1}}
    }
    
    \IEEEauthorblockA{\IEEEauthorrefmark{1} Miner School of Computer and Information Sciences, University of Massachusetts Lowell, USA, \\ \IEEEauthorrefmark{2} Department of Computer Science and Engineering, University of South Florida, USA,\\ 
    \IEEEauthorrefmark{3}  Department of Computer Science, University of Kentucky,\\
    \IEEEauthorrefmark{4} Department of Computer and Information Technology, Purdue University, USA, \\ \IEEEauthorrefmark{5}Department of Civil, Construction, and Environmental Engineering, North Carolina State University, USA \\
    \IEEEauthorrefmark{1}\{xingyu\_lyu, shixiong\_li, ian\_chen\}@uml.edu, \IEEEauthorrefmark{2} ningw@usf.edu, \IEEEauthorrefmark{3}litao@purdue.edu, \IEEEauthorrefmark{4}dchen33@ncsu.edu
    }
}

\maketitle

\begin{abstract}
Federated Learning (FL) is a popular paradigm that enables remote clients to jointly train a global model without sharing their raw data. However, FL has been shown to be vulnerable towards model poisoning attacks (MPAs) due to its distributed nature. Particularly, attackers acting as participants can upload arbitrary model updates that effectively compromise the global model of FL. While extensive research has been focusing on fighting against these attacks, we find that most of them assume data at remote clients are under independent and identically distributed (iid) while in practical they are inevitably \NIID. Our benchmark evaluations reveal that existing defenses thus generally fail to live up to their reputation when applied to various \NIIDs. In this paper, we propose a novel approach, \textbf{GeminiGuard}, that aims to address such a significant gap. We design {GeminiGuard} to be \textit{lightweight}, \textit{versatile}, and \textit{unsupervised} so that it well aligns the practical requirements of deploying such defenses. The key challenge from \NIIDs is that they make benign model updates look more similar to malicious ones. \GG is mainly built on two fundamental observations: (1) existing defenses based on either model-weight analysis or latent-space analysis face limitations in covering different MPAs and \NIIDs and (2) model-weight and latent-space analysis are sufficiently different yet potentially complimentary methods as MPA defenses. We hence incorporate a novel model-weight analysis component as well as a custom latent-space analysis component in \GG, aiming to further enhancing the defense performance of \GG. 
We conduct extensive experiments to evaluate our defense across various settings, demonstrating its effectiveness in countering multiple types of untargeted and targeted MPAs, even including an adaptive one. Our comprehensive evaluations show that {GeminiGuard} consistently outperforms state-of-the-art (SOTA) defenses under various settings.


\end{abstract}


\input{1-intro}
\input{2-system-adversary-model}
\input{3-method}

\input{4-evaluation}
\input{6-related}
\input{7-conclusion}

{
\small
\bibliographystyle{ieeetr}
\bibliography{paper}
}

\end{document}

%% file: macro.tex
\newcounter{note}[section]

\newcommand{\ie}{\emph{i.e.}\xspace}
\newcommand{\eg}{\emph{e.g.}\xspace}

\newcommand{\ignore}[1]{}

\newcounter{packednmbr}

\newcommand{\cmark}{\ding{51}}%
\newcommand{\xmark}{\ding{55}}%

\newcounter{lessoncount}

%% file: 1-intro.tex
\section{Introduction}
\label{sec:intro}
Federated Learning (FL) enables multiple remote devices to collaboratively train a Deep Neural Network (DNN) on a central server while keeping their local data private~\cite{mcmahan2017communication}. This privacy-preserving capability makes FL ideal for various real-world applications where security is essential, such as autonomous driving, IoT devices, and healthcare.
While FL has been widely recognized as one major framework for training DNN models distributively, one main hurdle is its vulnerability towards model poisoning attacks (MPAs). MPAs have been demonstrated capable of degrading the overall performance (i.e., untargeted attacks~\cite{fang2020local}) or manipulating the targeted predictions (i.e., targeted or backdoor attacks~\cite{bagdasaryan2019differential}) of the victim model. 

To mitigate \mpas in FL, researchers have introduced novel defenses~\cite{cao2021fltrust,wang2022flare,zhangfldetecotr,zhang2023flipprovabledefenseframework,MESAS,CrowdGuard,fereidooni2023freqfed,ali2024adversarially}. Most existing defenses are \textit{designed} under the assumption of independent-and-identically-distributed (iid) settings, neglecting the fundamental non-iid characteristic of FL. That is, the data distributions at different FL clients are inevitably different in practical. Some defenses provide evaluations for non-iid scenarios while evaluated under only one or two types of \NIIDs as illustrated in Table~\ref{table:flag-position}, which limits their applicability in real-world scenarios. Our experimental results suggest that the performance of these defenses degrade severely under other \NIID settings. This paper aims to address the gap by proposing solutions that work effectively across diverse \NIID scenarios.

\begin{table*}[t]
    \centering
    \scriptsize
    \begin{tabular}{p{2.4cm}|p{1.3cm}|p{1.5cm}|p{1.3cm}|p{1.0cm}|p{2.0cm}|p{1.3cm}|p{3.4cm}} 
        \toprule
        Defense & Built-in \NIID design & Model-weight analysis & Latent-space analysis  & Auxiliary data  & Types of MPAs & Dataset(s) & Non-iid types evaluated  \\ 
        \hline
        \texttt{FLTrust}~\cite{cao2021fltrust}  & \xmark & \cmark & \xmark & \cmark & untargeted, targeted & Image  & prob \\
        \hline  
        \texttt{FLGuard}~\cite{lee2023flguard} & \xmark & \cmark & \xmark & \xmark  & untargeted, targeted & Image & prob \\
        \hline  
     \texttt{FLAME}~\cite{nguyen2022flame} & \xmark & \cmark & \xmark & \xmark & untargeted, targeted  & Image, Text & prob \\
     \hline  
      \texttt{FreqFed}~\cite{fereidooni2023freqfed}& \xmark & \cmark & \xmark & \xmark & untargeted, targeted & Image, Text  & prob \\
      \hline  
        \texttt{FLDetector}~\cite{zhangfldetecotr} & \xmark & \cmark & \xmark & \xmark & untargeted, targeted  & Image & prob \\
        \hline       \texttt{FLIP}~\cite{zhang2023flipprovabledefenseframework}& \cmark & \xmark & \xmark & \xmark &  targeted & Image  & dir\\
        \hline
        \texttt{FLARE}~\cite{wang2022flare} & \cmark & \xmark & \cmark & \cmark  & untargeted, targeted & Image &  qty\\
        \hline
        \texttt{MESAS}~\cite{MESAS} & \xmark & \cmark & \xmark & \xmark  & untargeted, targeted & Image  & dir, qty \\
        \hline
        \texttt{CrowdGuard}~\cite{CrowdGuard} & \cmark & \xmark & \cmark & \cmark  &  targeted & Image & dir, prob \\
        \hline
        \texttt{FLShield}~\cite{kabir2024flshield} & \cmark  & \cmark & \xmark & \cmark    & untargeted, targeted & Image &  dir, prob \\
        \hline
        \texttt{AGSD}~\cite{ali2024adversarially} & \xmark   & \cmark & \xmark & \cmark  &  targeted & Image & prob, qty \\
        \hline
        \texttt{GeminiGuard (Ours)} & \cmark  & \cmark  & \cmark  & \cmark  & untargeted, targeted & Image, Text & prob, dir, qty, noise, quantity skew\\
        \bottomrule
    \end{tabular}
    \caption{Comparison between \FG and recent STOA defenses considering datasets, attack types, and non-iid types.} 
    \label{table:flag-position}
    \vspace{-10pt}
\end{table*}

We aim to propose a \textit{lightweight}, \textit{widely applicable}, and \textit{unsupervised} defense for FL against MPAs, which works under a broad range of MPAs and non-iid scenarios. We aim for it to be unsupervised, making it less data-dependent and therefore more practical for deployment.
As illustrated in Figure~\ref{fig:intuition-weights}, the main challenges of defending against MPAs under non-iid settings consists of two aspects. The primary challenge is that non-iid data inherently brings benign updates closer to malicious updates, making it harder to differentiate malicious clients from benign ones. Meanwhile, the impacts of different \NIID settings and MPAs vary significantly, and we aim to design a single defense that can effectively address most of them.



To this end, we propose \textit{GeminiGuard}, a novel framework to defend FL against model poisoning attacks under \NIIDs by combining model weights and latent space analysis. 
Our intuition comes from a fundamental summary of earlier defense solutions: they rely on either analyzing model weights or inspecting representations in latent space for identifying malicious model updates. That is, the first perspective focuses on model weights to identify malicious updates~\cite{cao2021fltrust,zhangfldetecotr,nguyen2022flame,MESAS,lee2023flguard,ali2024adversarially} while the second perspective emphasizes latent space behavior and examining hidden layers of local models to detect anomalies~\cite{wang2022flare,CrowdGuard}. This thus intrigues us to raise a question: \textit{what if we design a defense combining the two perspectives?} The intuition behind this is that the two perspectives are distinct and therefore can be complementary to each other, i.e., they address the impacts of MPAs and \NIIDs that are not completely overlapping. 

As a result, we design a model-weight analysis module following the first perspective in GeminiGuard. It employs an adaptive clustering method based on weights and key metrics, including cosine similarity and Euclidean distance, extracted from both the global model and local model updates. This module aims to distinguish malicious updates from benign ones in an initial filtering stage. Then, we design a latent-space analysis module following the second perspective in multi-layer settings. Specifically, it computes the average distance of activations at multiple layers of a received model update from those of other model updates for indicating how trustworthy the received model update is among all before aggregation of the global model.

We anticipate that combining these two perspectives will enable GeminiGuard to effectively address the challenges posed by \NIIDs and diverse MPAs. Our code is available at \url{https://anonymous.4open.science/r/GeminiGuard-1314}. We further summarize our main contributions as follows.






\begin{itemize}
    \item We propose \FG, an effective and robust defense against model poisoning attacks on FL under five types of challenging and comprehensive \NIID settings. To the best of our knowledge, \FG is the first model poisoning defense focusing on addressing \NIID challenges in an unsupervised, lightweight, and comprehensive way.
    \item {We propose the design of \FG following our intuition that model-weight and latent-space analysis are different and can complement each other when combined together for defending against MPAs.}
    \item We evaluate the performance of \FG through comprehensive experiments that encompass four untargeted MPAs, five backdoor attacks, and nine recent defenses across four datasets (\ie, MNIST, Fashion-MNIST, CIFAR-10, and Sentiment-140). Compared to the baselines including state-of-the-art (STOA), \FG consistently outperforms them on these datasets. For instance, on CIFAR-10 under IBA attack, \FG achieves an attack success rate (ASR) as low as 0.18\%, whereas the attack success rate of baseline defenses exceeds 10\% under qty-based \NIID settings (see Table~\ref{tab:noniid_backdoor}). We also conduct comprehensive ablation studies to analyze the impact of various factors on the robustness of \FG, including the degree of non-iid, the number of malicious clients, the data poisoning rate, the choice of metrics, the use of auxiliary datasets, and the number of layers. Finally, we evaluate \FG against a novel adaptive attack as well.

\end{itemize}

%% file: 2-system-adversary-model.tex
\section{System and Adversary Model}
\label{sec:system_adversary_model}
\subsection{System Model}
\label{sec:system-model}
\subsubsection{\NIID dataset settings} \label{sec:noniid}
Our target FL systems consist of two types of entities: a parameter server (PS) and $n$ remote clients. As discussed in Section~\ref{sec:intro}, in this paper we focus on \NIID settings, where the local dataset $\mathcal{D}_i$ of a remote client $u_i$ follows a certain type of non-identical distribution. In a general \NIID setting, the data across clients can vary in terms of both the number of data classes and the amount of data samples. As shown below, we explore five types of \NIIDs: \textit{Dirichlet-based (dir), probability-based (prob), quantity-based (qty), noise-based (noise), and quantity skew (qs)}, as proposed in Li et al.~\cite{li2022federated}. Through comprehensive investigations into the impacts of different \NIIDs on FL performance, the authors concluded that label distribution skew (e.g., Dirichlet-based, probability-based, and quantity-based), feature distribution skew (e.g., noise-based), and quantity skew occurred most often in practical, hence deserving attention from the research community.

\begin{figure}
    \centering
    \includegraphics[width=0.96\linewidth]{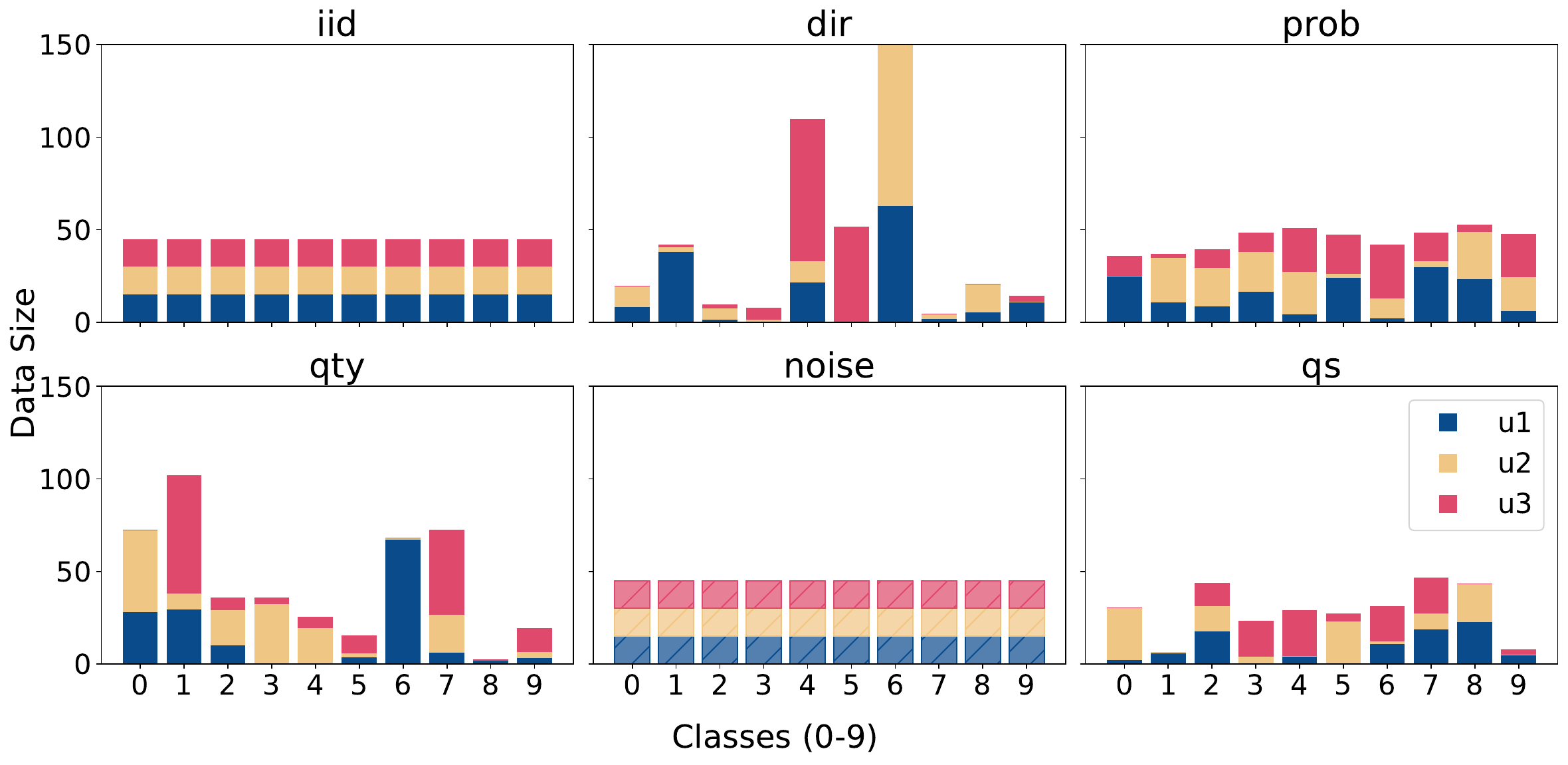}
    \vspace{-5pt}
   \caption{Illustrations of iid and five non-iids (dir-based, prob-based, qty-based, noise, qs) for $u_1, u_2$, and $u_3$, and 10 label classes.}
    \label{fig:iid-non-iid-examples}
    \vspace{-5pt}
\end{figure}

Fig.~\ref{fig:iid-non-iid-examples} provides illustrations of iid and the five \NIIDs we focus on, assuming a simple scenario with three clients ($u_1$, $u_2$, and $u_3$), 10 data classes (from label `0' to label `9'), and 150 samples per client. At first glance, we can observe that the six distributions lead to significantly different sample sets across the three clients. First, iid refers to the case where different clients have the same number of data classes and the same number of samples per class. Second, we define the five investigated \NIID types below.




\begin{itemize}
    \item \textbf{Type-I \NIID (dir)}: Dirichlet distribution-based label imbalance, i.e., dir-based. We can see from Fig.~\ref{fig:iid-non-iid-examples} that this type allocates each client ($u_1, u_2$, and $u_3$) a certain number of samples per label (can even be 0) according to a given Dirichlet distribution, resulting at a diverse and imbalanced data distribution among clients. Several existing defenses against MPAs in FL~\cite{zhang2023flipprovabledefenseframework,CrowdGuard,MESAS} have explored dir-based \NIID.
    \item \textbf{Type-II \NIID (prob)}: Probability-based label imbalance, i.e., prob-based. Clients ($u_1, u_2$, and $u_3$) are first divided into groups based on the main label class(es) they have. In Fig.~\ref{fig:iid-non-iid-examples}, the main label class of $u_1$ is `0' while that of $u_2$ and $u_3$ is `1' and `6', respectively. After that, a given data sample is assigned to the corresponding primary group with a probability $q$ and then to other groups with a probability of $\frac{1-q}{G-1}$ where $G$ is the number of groups. Consequently, a higher $q$ leads to more severe skewed prob-based \NIIDs. Likewise, this \NIID type has been explored in~\cite{fang2020local,cao2021fltrust,zhangfldetecotr,CrowdGuard}.
    \item \textbf{Type-III \NIID (qty) }: Quantity-based label imbalance, i.e., qty-based. In this case, each client only has training samples from the same fixed number of labels. In Fig.~\ref{fig:iid-non-iid-examples}, $u_1, u_2$, and $u_3$ only have samples from only a subset of labels. Likewise, this \NIID type has been explored in~\cite{wang2022flare,mcmahan2017communication}.
    \item \textbf{Type-IV \NIID (noise)}: Noise-based feature distribution skew detailed as follows. After dividing the whole dataset equally among $u_1, u_2$, and $u_3$, Gaussian noise is added to the samples of each client $u_i$ to achieve feature diversity~\cite{li2022federated}. Given a defined noise level $\sigma$, the noise follows the Gaussian distribution $\mathcal{N}\left(0, \sigma \times \frac{i}{N}\right)$, where $N$ is the total number of clients and $\sigma$ is the degree of feature dissimilarity among the clients.
    
    \item \textbf{Type-V \NIID (quantity-skew, qs)}: Under quantity skew, the size of each client's local dataset ($|D_i|$) varies across clients~\cite{zhu2021federated}. In specific, we use a Dirichlet distribution to allocate different number of samples to each client as in~\cite{li2022federated,luo2021no}. We achieve this by sampling $q \sim \text{Dir}(\beta)$, where $\beta$ corresponds to the level of quantity imbalance. Then we can assign a proportion of $q$ from the whole set of samples to a specific client. 
\end{itemize}
In brief, we find that rarely an existing defense against MPAs had considered more than two \NIIDs, indicating the necessity of defense solutions effective under more comprehensive \NIID settings.

\begin{figure}[htbp]
    \centering
    \includegraphics[width=\linewidth]{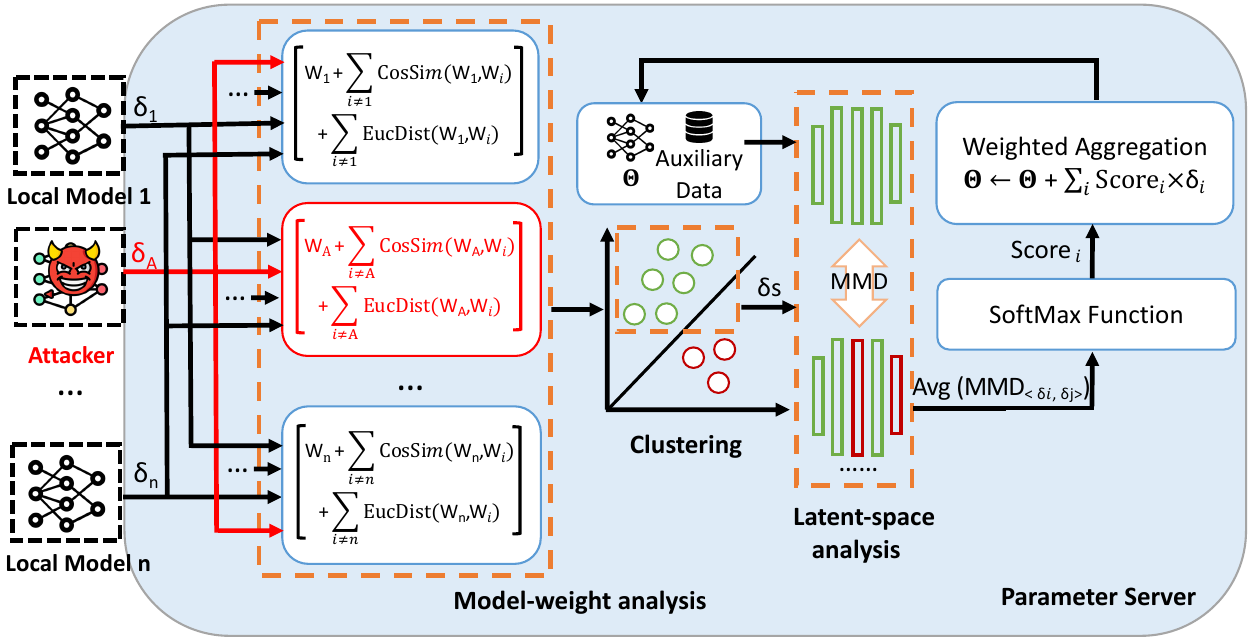}
    \caption{Workflow of FL with \textbf{\FG}.}
    \label{fig:flag-flow}
    \vspace{-5pt}
\end{figure}


\subsubsection{FL Workflow}
Fig.~\ref{fig:flag-flow} illustrates the overall workflow of our targeted FL system as explained below. To start with, each remote client trains and manages a local model while at the cloud site PS distributes and aggregates model updates from selected remote clients. For convenience, we denote the model parameter of client $u_i$ by $\mathbf{\theta}_i\in \mathcal{W}\subseteq \mathbb{R}^d$, where $\mathcal{W}$ is the parameter space and $d$ is the dimensionality of the FL model. Similarly, we further denote the global model by $\Theta\in \mathcal{W}$. As a result, the model update from client $u_i$ can be written as $\delta_i=\mathbf{\theta}_i - {\Theta}$, which awaits to be aggregated into $\Theta$. Assume that the entropy-based loss for a local model $\mathbf{\theta}_i$ is $\mathcal{L}(\theta_i, \mathcal{D}_i)$. Therefore, the total loss function $\mathcal{F}(\Theta)$ can be computed as $\mathcal{F}(\Theta) \coloneqq \sum^{k}_{i=1} \mathcal{L}(\theta_i, \mathcal{D}_i)$, where $k$ is the number of selected clients for participating in the current training round. We also assume PS has a small auxiliary dataset for validation purposes (as in ~\cite{cao2021fltrust,wang2022flare,CrowdGuard,kabir2024flshield,ali2024adversarially} and explained in Section~\ref{sec:method}). The overall training goal is to jointly minimize $\mathcal{F}(\Theta)$ by optimizing $\Theta$. 

The training process of $\Theta$ is as follows. After PS first initializes $\Theta$, each training round follows the three steps: (1) PS first sends $\Theta$ to the randomly selected $k$ remote clients. (2) Assuming $u_i$ is selected, $u_i$ will initialize $\mathbf{\theta}_i\!=\!\Theta$ and then train $\mathbf{\theta}_i$ with its own local training samples and eventually upload the corresponding model update $\delta_i$ to PS. PS invokes \FG and computes a trust score for $\delta_i$. (3) Finally, PS aggregates local model updates together using their trust scores as the weight and updates $\Theta$.

\subsection{Adversary Model}
\textbf{Attackers' goals.} We adopt the attack objectives of untargeted MPAs from~\cite{fang2020local,shejwalkar2021manipulating} and targeted ones from~\cite{xie2019dba, zhang2022neurotoxin,nguyen2024iba}. For untargeted MPAs, the attacker aims to craft model updates that degrade the performance of the global model. For targeted MPAs, we focus on backdoor attacks. On one hand, a backdoor attacker aims to insert defined backdoors into the global model $\Theta$ so that $\Theta$ is misled to classify input samples with the backdoor `trigger' into a target label. On the other hand, the attacker aims that $\Theta$ under attack still maintains high prediction accuracy on `clear' inputs, rendering their attack difficult to detect (\ie, stealthy).

\textbf{Attackers' capabilities.} As in the literature, we assume that our attackers are also legitimate clients of FL and have a local dataset that they can use to train a local model. Furthermore, they can arbitrarily manipulate (modify, add, or delete) their local data, such as inserting triggers into samples or altering the labels. Additionally, they can craft their model updates sent to PS arbitrarily in order to achieve the above attack goals. As a client, the attackers have white-box access to the global model $\Theta$ but no local models of other clients. We assume the number of malicious clients (i.e., attackers) to be less than half of all clients. PS is assumed to be trusted and to deploy defense mechanisms including \FG against MPAs to properly protect $\Theta$. 

%% file: 3-method.tex
\section{GeminiGuard: MPA Defenses under Non-iids}
\label{sec:method}

\subsection{Intuition}
\label{subsec:design-intuition}


As mentioned in Sect.~\ref{sec:intro}, our main intuition is that model-weight analysis and latent-space representation inspection offer complementary perspectives in defending against MPAs. Upon receiving $\delta_i$ from $u_i$, PS performs analysis on $\delta_i$ to determine how suspicious it is. To this end, PS aggregates $\delta_i$ into the current global model $\Theta$ and obtains $\Theta_i$. Then the analysis on $\delta_i$ is done through analyzing $\Theta_i$. Specifically, given a $\Theta_i$, PS can either directly extract statistical features or first feed a small fixed set of samples to $\Theta_i$ and then use the activations from different layers for analysis. We refer to the former approach as \textbf{model-weight analysis} and the latter as \textbf{latent-space analysis}. Our design is based on the following findings: 
\begin{enumerate}[label=\textbf{{{\arabic*.}}}]
    \item \textit{the two approaches are different, and}
    \item \textit{they can complement each other for mitigating MPAs.}
\end{enumerate} 

\begin{figure}[t]
    \centering
    \begin{subfigure}[b]{0.15\textwidth}
        \includegraphics[width=\textwidth]{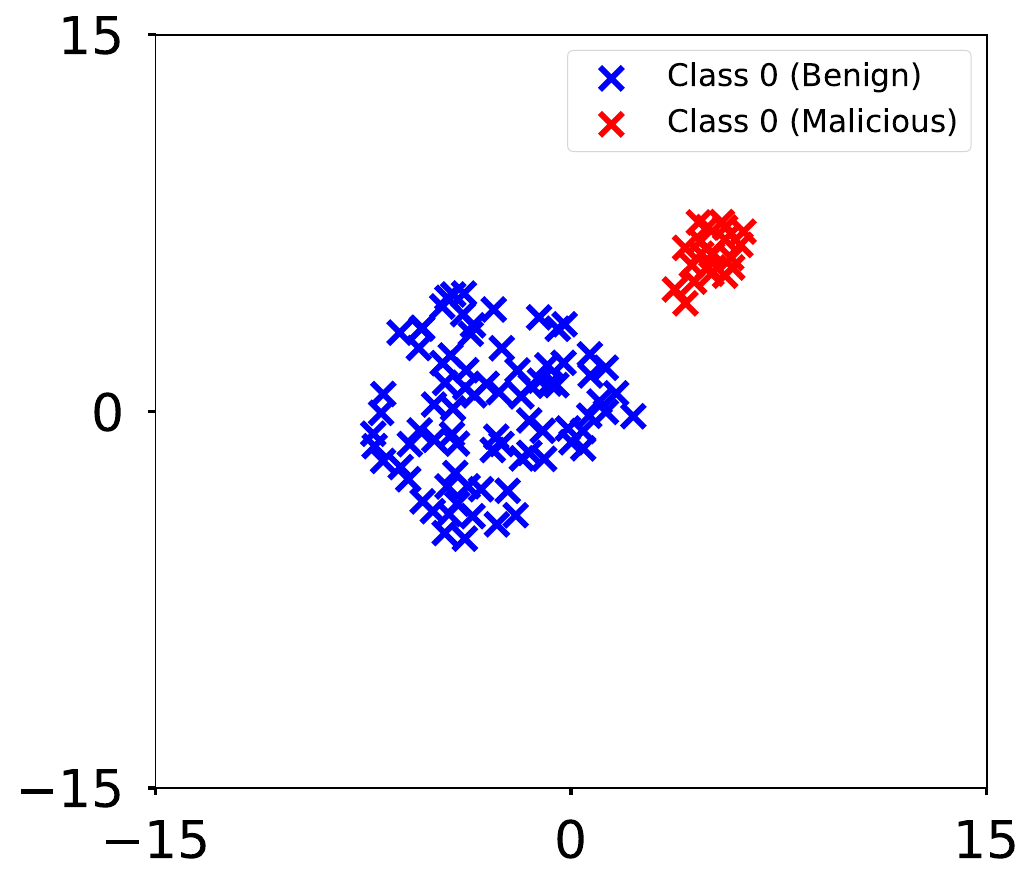}
        \caption{iid}
        \label{fig:noniid-aaa}
    \end{subfigure}
    \begin{subfigure}[b]{0.15\textwidth}
        \includegraphics[width=\textwidth]{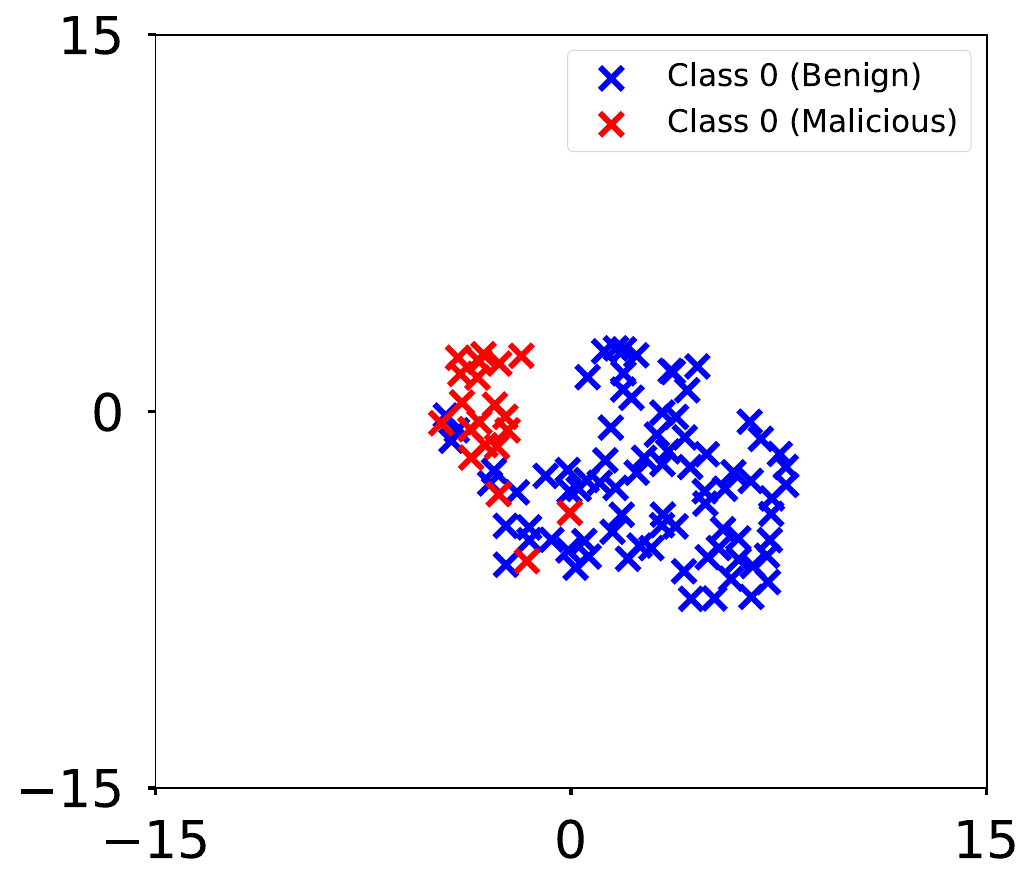}
        \caption{\NIID (qty)}
        \label{fig:noniid-bbb}
    \end{subfigure}
    \begin{subfigure}[b]{0.15\textwidth}
        \includegraphics[width=\textwidth]{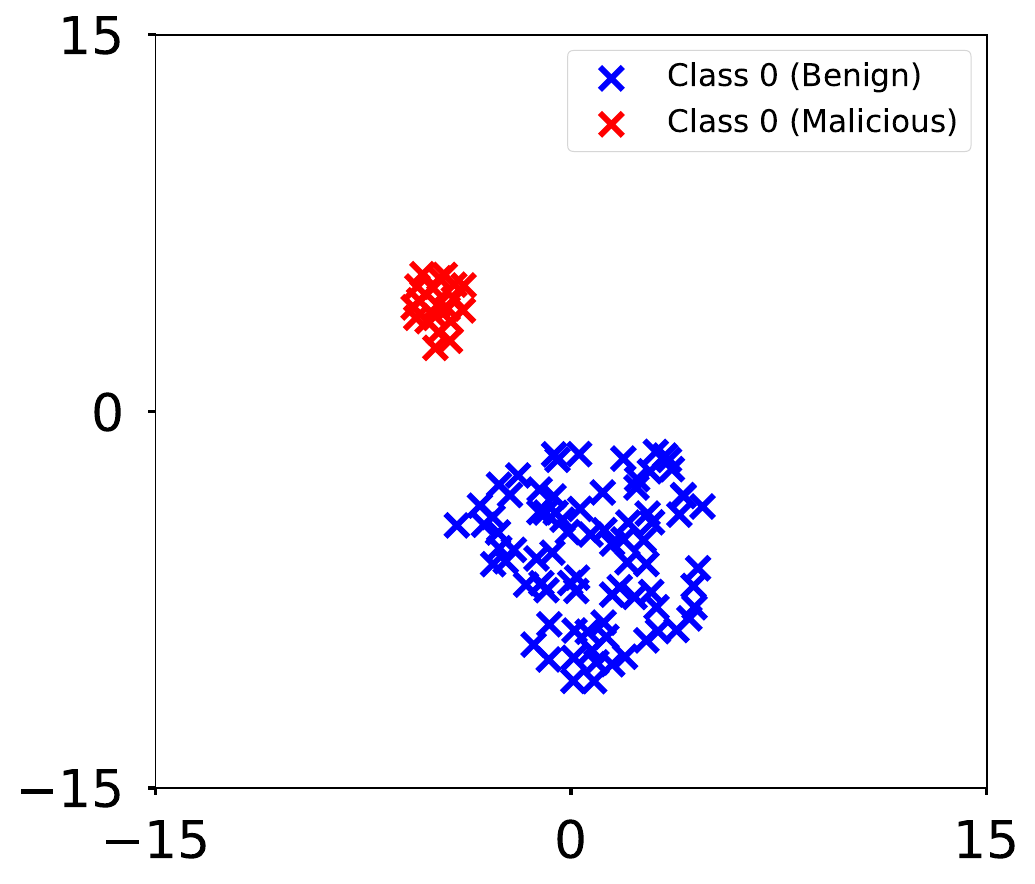}
        \caption{\NIID (with GG)}
        \label{fig:noniid-ccc}
    \end{subfigure}
    \vspace{-5pt}
    \caption{Model updates in t-SNE (BadNets, CIFAR-10) under different scenarios: iid, \NIID (qty), and with \FG(GG).}
    \label{fig:intuition-weights}
    \vspace{-10pt}
\end{figure}

\begin{figure}[t]
    \centering
    \begin{subfigure}[b]{0.15\textwidth}
        \includegraphics[width=\textwidth]{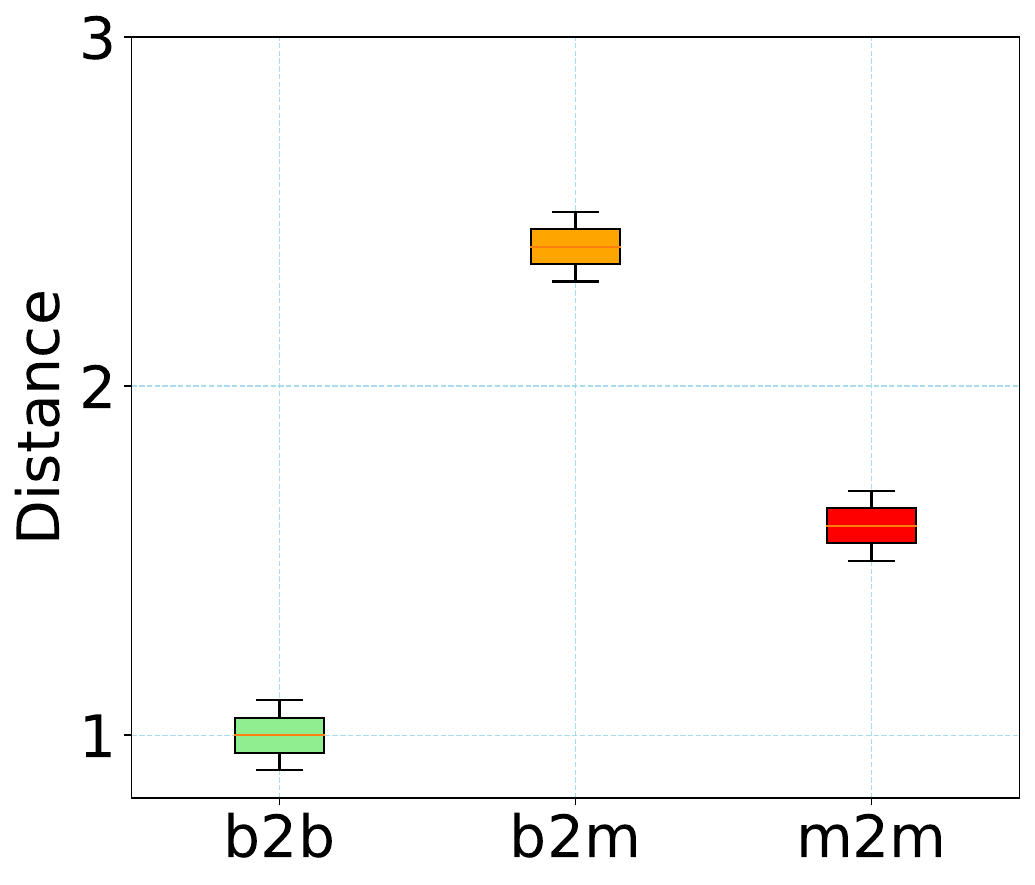}
        \caption{iid}
        \label{fig:noniid-a}
    \end{subfigure}
    \begin{subfigure}[b]{0.15\textwidth}
        \includegraphics[width=\textwidth]{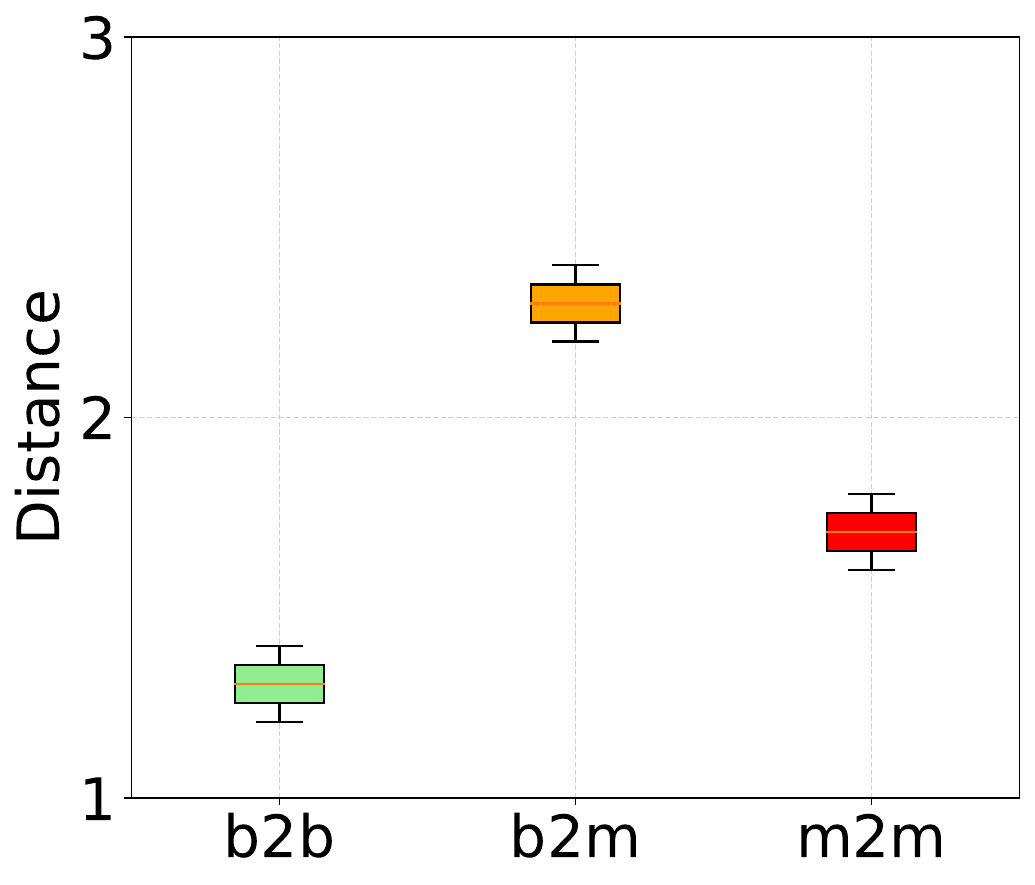}
        \caption{\NIID (dir)}
        \label{fig:noniid-b}
    \end{subfigure}
    \begin{subfigure}[b]{0.15\textwidth}
        \includegraphics[width=\textwidth]{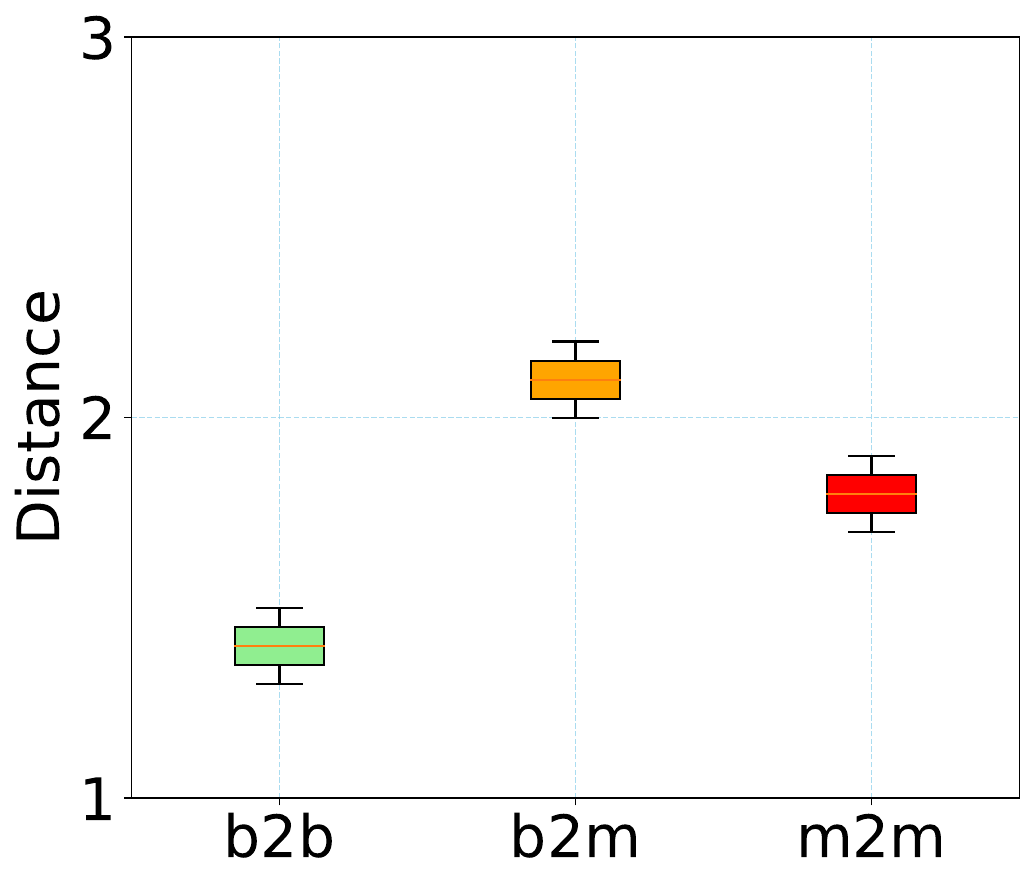}
        \caption{\NIID (qty)}
        \label{fig:noniid-c}
    \end{subfigure}
    \vspace{-5pt}
    \caption{Distance of layers between Benign-to-Benign (b2b), Benign-to-Malicious (b2m), and Malicious-to-Malicious (m2m).}
    \label{fig:intuition-layers}
    \vspace{-10pt}
\end{figure}

Here we show the preliminary results illustrating the necessity for combining model-weight and latent-space analysis together. Fig.~\ref{fig:noniid-aaa} and Fig.~\ref{fig:noniid-bbb} show the t-SNE results of benign and malicious model updates, i.e., different $\Theta_i$s, under iid and one non-iid (i.e., qty). We can see from the two figures that they become more difficult to distinguish under \NIID when compared to iid, as benign updates under non-iids tend to expand toward malicious ones. Fig.~\ref{fig:intuition-layers} illustrates the layer-wise average Euclidean distance between activations of two benign model updates, a pair of benign and malicious updates, and finally two malicious model updates. We further denote the three scenarios as \textit{benign-2-benign (i.e., b2b), benign-2-malicious (i.e., b2m), and malicious-2-malicious (i.e., m2m)}. Fig.~\ref{fig:intuition-layers} shows that under iid, most likely benign-2-benign distances are smaller than malicious-2-malicious ones, which are further smaller than benign-2-malicious distances. Such an observation is consistent to the literature. Moving towards non-iids such as dir or qty, the distinguishability among the three distance metrics tends to decrease while in general they can be separable. Our idea is to combine analysis based on model-weight features and latent-space representations together, enabling them to complement each other effectively so as to enhance the overall model robustness against different MPAs and non-iids combined. As Fig.~\ref{fig:noniid-ccc} shows, our proposed method can effectively increase the distance between the benign and malicious clusters, indicating its better performance. 

\subsection{Design of \FG}
\label{subsec:design-flag}
\textbf{Overview.} Fig.~\ref{fig:flag-flow} shows the workflow of GeminiGuard, consisting of model-weight and latent-space analysis. Specifically, upon receiving a model update from a client, \FG proceeds to first construct the model weights and then compute pair-wise cosine similarity and Euclidean distance. Then \FG performs adaptive clustering based on a combined feature vector consisting of the model weights and the two distance metrics while filtering out $\Theta_i$s that fail to fit in one of the clusters. Next, in the latent-space analysis module, \FG computes a trust score for each model update that survives by calculating multi-layer discrepancy. Finally, PS aggregates the remaining model updates using their trust scores for updating the global model. We detail the above pipeline in Algorithm~\ref{alg:FLAG} and as what follows.

\textbf{Step 1:} In the typical FL process, a client $u_i$ sends its model update to the PS after local training. Specifically, $u_i$ first receives the global model $\Theta$ from the PS and uses it to initialize its local model $\mathbf{\theta}_i$. The client then trains $\mathbf{\theta}_i$ using its local data and uploads the model update  \( \delta_i \) to the PS.  


\textbf{Step 2:}  
Next, \FG performs \textit{model-weight analysis} to effectively exclude `highly suspicious' malicious models based on statistic features. The specific features of a received  \( \delta_i \) we look into include the weight vector and its cosine similarity and Euclidean distance from other weight vectors. For each \( u_i \), we first compute the weight vector \( \mathbf{w}_i \) based on the model updates \( \delta_i \) (by \( \mathbf{w}_i = \Theta + \delta_i \)). Then, \FG computes pairwise cosine similarity \( \text{CosSim}(\mathbf{w}_i, \mathbf{w}_j) \) (to check the direction difference) and Euclidean distance \( \text{EucDist}(\mathbf{w}_i, \mathbf{w}_j) \) (to check the magnitude difference), between each pair of updates \( \mathbf{w}_i \) and \( \mathbf{w}_j \) (\( i, j \in \{1, \ldots, n\} \), where \( n \) is the number of  \( \delta_i \)s received by the PS). In the end, \FG constructs a feature vector for \( \delta_i \):
\[
\mathbf{z}_i = [\mathbf{w}_i; \sum\nolimits_{j \neq i}\text{CosSim}(\mathbf{w}_i, \mathbf{w}_j); \sum\nolimits_{j \neq i}\text{EucDist}(\mathbf{w}_i, \mathbf{w}_j)],
\]


Following that, \FG applies K-Means clustering (as in~\cite{kabir2024flshield}) to the set of \( \{\mathbf{z}_1, \mathbf{z}_2, \ldots, \mathbf{z}_n\} \) of all selected clients to dynamically group updates into clusters. We determine the number of clusters \( k^* \) using the Silhouette Coefficient~\cite{rousseeuw1987silhouettes}, which is to evaluate clustering results based on intra-cluster and inter-cluster distances. After clustering, we filter out those model updates that do not belong to any cluster while keeping the ones resulting at a distance smaller than a given threshold $\tau$:
\[
\mathcal{B} = \{ \mathbf{z}_i \mid \min_{j} \| \mathbf{z}_i - \mathbf{c}_j \| \leq \tau \},
\]
where \( \mathcal{B} \) represents the benign cluster and \( \mathbf{c}_j \) denotes the centroid of cluster \( j \).


\textbf{Step 3:} In \textit{latent-space module}, \FG focuses on obtaining the multi-layer activations for each received \( \delta_i \in \mathcal{B} \) for subsequent analysis.
We assume that PS has a small set of auxiliary iid data as in recent STOA defenses~\cite{cao2021fltrust,CrowdGuard,wang2022flare,ali2024adversarially}, which is for obtaining activations from each neural layer. For convenience, we denote such a dataset by $D_A$. Similarly, {\FG first obtains $\Theta_i$ from the current $\Theta$ and $\delta_i$.} The corresponding sequence of multi-layer activations of $\delta_i$ can be obtained by sending $D_A$ to $\Theta_i$, extracting per-sample activations at all layers (excluding the output layer), and concatenating them together. As shown in Sect.~\ref{sec:evaluation:layers}, we investigate the impact of different layers for latent-space analysis as well. Denote the mapping relationship with $\Theta_i$ by $g_{\Theta_i}:x\in \mathcal{R}^{l \times w \times h} \rightarrow r \in \mathcal{R}^{L}$ where $l$, $w$, and $h$ denote the width, height, and channels of an image input, respectively, and $L$ the size of layers. Assume that $D_A$ has $m$ (we set 50 as~\cite{wang2022flare,ali2024adversarially}) samples. As a result, the sequence of multi-layer activations of $\delta_i$ is thus $\text{LSeq}_i =: \{g_{\Theta_i}(x_1),\ldots,g_{\Theta_i}(x_m)\}$. 

\textbf{Step 4:} In this step, \FG is to compute a trust score for each $\delta_i$ by first using Maximum Mean Discrepancy~\cite{ArthurMMD12} (MMD) to obtain the distance between any two $\text{LSeq}_i$ and $\text{LSeq}_j$ and then estimating the trust score through the average distance. That is, denote the distance between \( \text{LSeq}_i \) and \( \text{LSeq}_j \) (the $\text{LSeq}$ for $\delta_i$ and $\delta_j$) as below:
\begin{equation}
\fontsize{8}{8}\selectfont
    \begin{aligned}
        \text{MMD}(\text{LSeq}_i, \text{LSeq}_j) = \frac{1}{m(m-1)} \Bigg[ &\sum_{a \in \text{LSeq}_i} \sum_{b \in \text{LSeq}_i, b \neq a} k(a, b) \\
        + \sum_{a \in \text{LSeq}_j} \sum_{b \in \text{LSeq}_j, b \neq a} k(a, b) 
        &- 2 \sum_{a \in \text{LSeq}_i} \sum_{b \in \text{LSeq}_j} k(a, b) \Bigg]
    \end{aligned}
\end{equation}
where \( k(\cdot) \) is a Gaussian kernel function. Then, we compute the trust score of $\delta_i$ following the below steps:
\begin{itemize}
    \item For each $\delta_j, j\neq i$ in current round, \FG computes $\text{MMD}(\text{LSeq}_i, \text{LSeq}_j)$. This average distance of $\delta_i$, denoted by \( \text{AvgDist}_i \), is thus calculated as $\text{mean}(\text{MMD}(\text{LSeq}_i, \text{LSeq}_j))$ and serves to reflect the average similarity between $\delta_i$ and other model updates.
    \item \FG then compute the trust score \( \text{Score}_i \) of $\delta_i$ by the following equation:
    \begin{equation}
        \text{Score}_i = \frac{\exp(1 / \text{AvgDist}_i / \tau)}{\sum_{k=1}^n \exp(1 / \text{AvgDist}_k / \tau)},
    \end{equation}
    where \( \tau \) is a temperature parameter to adjust the sensitivity of trust score.
\end{itemize}

\textbf{Step 5:} Finally, \FG aggregates all remaining $\delta_i$s using their trust scores as weights and updates $\Theta$.


\begin{algorithm}
\caption{\texttt{GeminiGuard} Algorithm}
\label{alg:FLAG}
\begin{algorithmic}[1]
\Require $G_0$, $T$ $\triangleright$ $G_0$: Initial global model, $T$: Number of training iterations
\Ensure $G_T$ $\triangleright$ $G_T$: Final global model after $T$ iterations

\For{each training iteration $t \in [1, T]$}
    \State \texttt{Step 1: Local Training}
    \For{each client $i \in [1, n]$}
        \State $\delta_i \gets ClientUpdate(G_t)$ $\triangleright$ Collect model updates
        \State $w_i \gets G_0 + \delta_i$ $\triangleright$ Collect model weights
    \EndFor

    \State \texttt{Step 2: Model-weight Analysis}
    \State $\mathcal{M} \gets \{\sum_{j \neq i}\text{CosSim}(w_i, w_j); \sum_{j \neq i}\text{EucDist}(w_i, w_j)\}$
    \State $\mathcal{C}, \mathcal{O} \gets  Kmeans (\{w_i\}_{i=1}^n + \mathcal{M})$ $\triangleright$ Cluster clients into inliers $\mathcal{C}$ and outliers $\mathcal{O}$ 
    
    \State \texttt{Step 3: Latent-space Feature Extraction}
    \State $\text{\texttt{Layers}} \gets ExtractFeatures(\mathcal{C}, D_A)$ $\triangleright$ Extract all hidden layer features using a auxiliary dataset
    
    \State \texttt{Step 4: Trust Score Calculation}
    \State $\text{\texttt{Scores}} \gets ComputeScores(\text{\texttt{Layers}})$ $\triangleright$ Compute trust scores for each model update of inliers
    
    \State \texttt{Step 5: Aggregation}
    \State $G_{t+1} \gets Aggregate(G_t, \mathcal{C}, \text{\texttt{Scores}})$ $\triangleright$ Aggregate weighted updates
\EndFor

\State \Return $G_T$
\end{algorithmic}
\end{algorithm}

%% file: 4-evaluation.tex
\section{Performance Evaluation}
\label{sec:evaluation}
In this section, we evaluate the performance of \FG across five Non-IID scenarios, comparing it with SOTA baselines under both untargeted \mpa and backdoor attacks. Then, we perform an ablation study to analyze the impact of various parameters on \FG's effectiveness. Finally, we evaluate \FG's robustness against an adaptive attack.

\subsection{Experimental Settings}
\label{eval:setup}

\subsubsection{Implementation} 
Our experiments were implemented in Python using PyTorch. They were executed on a server with an Intel(R) Core(TM) i9-10900X CPU @ 3.70GHz, two NVIDIA RTX 3090 GPUs, and 181 GB main memory. 


\subsubsection{Datasets} 
\label{non-iid-datasets}
We evaluated \FG on three image datasets and one text dataset. For image classification task, we use three popular benchmark datasets MNIST, Fashion-MNIST (FMNIST), and CIFAR-10. Each of the three datasets contains samples from 10 classes and are frequently used for evaluations of backdoor attacks as well as defenses ~\cite{wang2022flare,zhangfldetecotr,CrowdGuard}. We use the Sentiment-140~\cite{go2009twitter} for the sentiment analysis task. This dataset consists of 1.6 million tweets, annotated with positive or negative sentiments, and is commonly used for training and evaluating sentiment classification models. We adopt the same attack settings on the Sentiment-140 dataset as described in~\cite{wang2020attack}. 

We considered IID and five \NIIDs in this paper. For IID scenarios, we divide the data equally among clients. For non-IID scenarios, we consider five \NIID types with the following default parameters: \textbf{1)} prob-based with a non-IID degree of 0.3, \textbf{2)} dir-based with a non-IID degree of 0.3,  \textbf{3)} qty-based where each client only has two classes, \textbf{4)} noise-based feature distribution skew with $\sigma=0.5$, and \textbf{5)} quantity skew with $\beta=0.3$ for Dirichlet distribution.  Aside from the above, we also explore different \NIID degrees in ablation study to evaluate their impacts on the effectiveness of \FG.

\subsubsection{FL Settings} 
In each FL round, we select 10 clients with \textbf{random selection} out of a total of 100 clients to participate FL training. By default, the malicious client ratio is set to 0.2, and the data poisoning rate is 0.5. Eeach selected client trains its local model using a batch size of 64 for five local epochs before aggregating the updates.  For MNIST and FMNIST datasets, we adopt cross-entropy as the loss function, stochastic gradient descent (SGD) optimizer with a learning rate of 0.001, and a momentum of 0.9 to train models. For CIFAR-10 dataset, we use the same loss function, optimizer, and momentum decade but adopt a learning rate of 0.01 for SGD instead. Finally, for Sentiment-140 dataset, we adopt cross-entropy with logit as the loss function and Adam optimizer with a learning rate of 0.001.

\subsubsection{Models.} Table~\ref{tab:models} summarizes the models used in our experiments. For the MNIST and FMNIST datasets, we employ a CNN model, while ResNet-50 is used for the CIFAR-10 dataset. For the Sentiment-140 dataset, we utilize an LSTM network designed to process sequential text data effectively.

\begin{table}[!ht]
    \renewcommand{\arraystretch}{1.1}
    \centering
     \fontsize{6.6}{7}\selectfont
    \caption{Datasets and models used in \FG.}
    \label{tab:models}
    \begin{tabular}{p{1.7cm}|p{1cm}p{1cm}p{1cm}p{1.7cm}}
        \hline
        \textbf{Dataset (\#categories)} & \textbf{\#training samples}  & \textbf{\#testing samples} & \textbf{\#input size} & \textbf{Model $\Theta$} \\ 
        \hline
        MNIST (10) & 60,000 & 10,000 & (28, 28, 1) & 3 Blocks + 6 Conv\\
        FMNIST (10) & 60,000 & 10,000 & (28, 28, 1) & 3 Blocks + 6 Conv\\
        CIFAR-10 (10) & 50,000 & 10,000 & (32, 32, 3) & ResNet-50\\
        Sentiment-140 (2) & 1,280,000 & 320,000 & (50,) & LSTM\\
        \hline
    \end{tabular}
\end{table}


\subsubsection{Metrics to Evaluate \FG}
\begin{itemize}
    \item \textbf{Attack Success Rate (\texttt{ASR})}: This measures the percentage of backdoor test samples that are misclassified as the targeted label by the victim  model. A lower \texttt{ASR} indicates better defense performance in all defenses.
    
    \item \textbf{Model Accuracy (\texttt{Acc})}: This represents the testing accuracy of the victim FL model on the test set. For untargeted model poisoning, a lower \texttt{Acc} signifies a more effective attack. In the case of backdoor attacks, a higher \texttt{Acc} suggests the attack is more stealthy while maintaining a higher \texttt{ASR}.
\end{itemize}

\subsubsection{\MPAs and Defense Baselines} 
\label{sec:attacks-and-defenses-used}
\begin{itemize}

\item \textbf{Attacks:} We evaluate \FG against two
categories of attacks:  untargeted \mpg and
targeted (backdoor) ones. Specifically, we consider four  untargeted \mpas: krum-attack~\cite{fang2020local}, trimmean-attack~\cite{fang2020local}, min-max attack~\cite{shejwalkar2021manipulating} and min-sum attack~\cite{shejwalkar2021manipulating}. For backdoors, we select five powerful and stealthy attacks: Visible Trigger Backdoor (\textit{BadNets}~\cite{gu2017badnets}); Invisible Trigger Backdoors (\textit{Edge-case}~\cite{wang2020attack}); Distributed Backdoor (\textit{DBA}~\cite{xie2019dba}); Durable Backdoor (\textit{Neurotoxin}~\cite{zhang2022neurotoxin}); and Irreversible Backdoor (\textit{IBA}~\cite{nguyen2024iba}).
    
\item \textbf{Defenses:} We compare our proposed defense with 9 SOTA \mpg defenses — FLTrust~\cite{cao2021fltrust}, FLARE~\cite{wang2022flare}, FLAME~\cite{nguyen2022flame}, FLDetector~\cite{zhangfldetecotr}, FLIP~\cite{zhang2023flipprovabledefenseframework}, FLGuard~\cite{lee2023flguard}, MESAS~\cite{MESAS}, FreqFed~\cite{fereidooni2023freqfed}, and AGSD~\cite{ali2024adversarially}. These methods were selected based on their prominence and recent contributions to the field.
\end{itemize}



\subsection{Comparison with the SOTA defenses}


\begin{table}[!ht]
    \renewcommand{\arraystretch}{1.3}
    \centering
    \footnotesize
    \caption{Model acc without poisoning attacks in FL.}
    \label{tab:acc_all_benigh}
    \begin{tabular}{p{1.7cm}|p{0.6cm}p{0.6cm}p{0.6cm}p{0.6cm}p{0.6cm}p{0.6cm}}
        \hline
         \centering
      \textbf{Dataset} & \textbf{iid} & \textbf{dir} & \textbf{prob} & \textbf{qty} & \textbf{noise} & \textbf{qs} \\ 
        \hline
         \centering
     MNIST & 98.55 & 94.60 & 93.69 & 94.45 & 94.48 & 94.83  \\
         \centering
    FMNIST & 90.98 & 83.11 & 80.70 & 81.19 & 87.75 & 80.76 \\
         \centering
    CIFAR-10  & 68.84 & 65.60 & 65.68 & 63.20 & 67.63 & 64.20  \\
         \centering
    Sentiment-140 & 78.69 & 73.12 & 72.56 & 70.98 & 73.66 & 71.61 \\
        \hline
    \end{tabular}
    \vspace{-5pt}
\end{table}

\textbf{Performance of the global models.}
Table~\ref{tab:acc_all_benigh}  presents the accuracy of clean global models without attacks across four datasets, evaluated under both IID and five \NIIDs settings. The results indicate that the performance decreases in \NIIDs due to differences in data distribution. In particular, the accuracy gap between \NIIDs and IID exceeds 5\% in most cases. It is worth noting that the noise in \NIIDs has a minimal impact on the model's performance.

\begin{figure*}[ht]
    \centering
    \begin{subfigure}[b]{0.19\textwidth}
        \includegraphics[width=\textwidth]{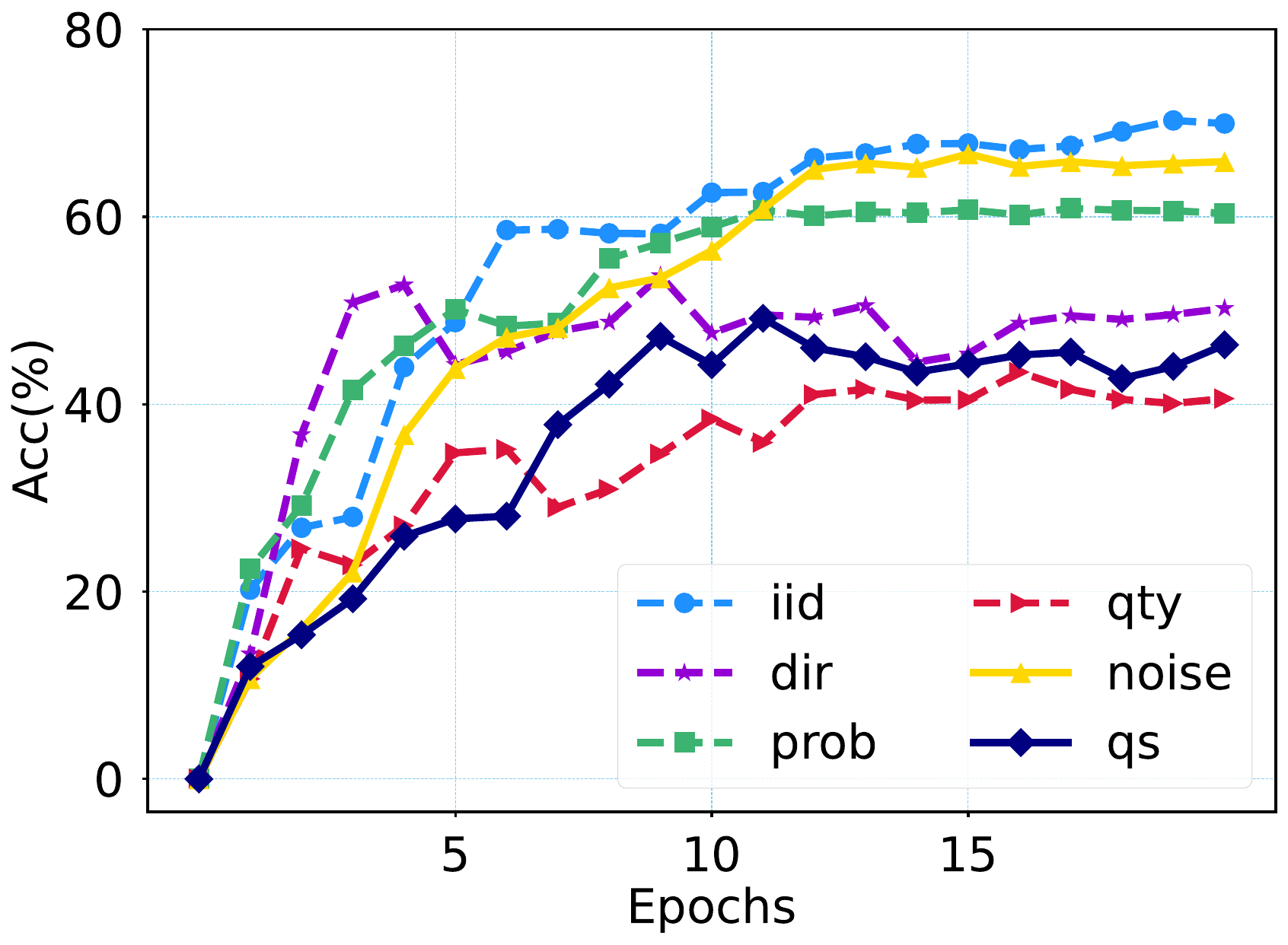}
        \caption{FLTrust~\cite{cao2021fltrust}}
        \label{fig:noniid_degree_FLTrust}
    \end{subfigure}
    \begin{subfigure}[b]{0.19\textwidth}
        \includegraphics[width=\textwidth]{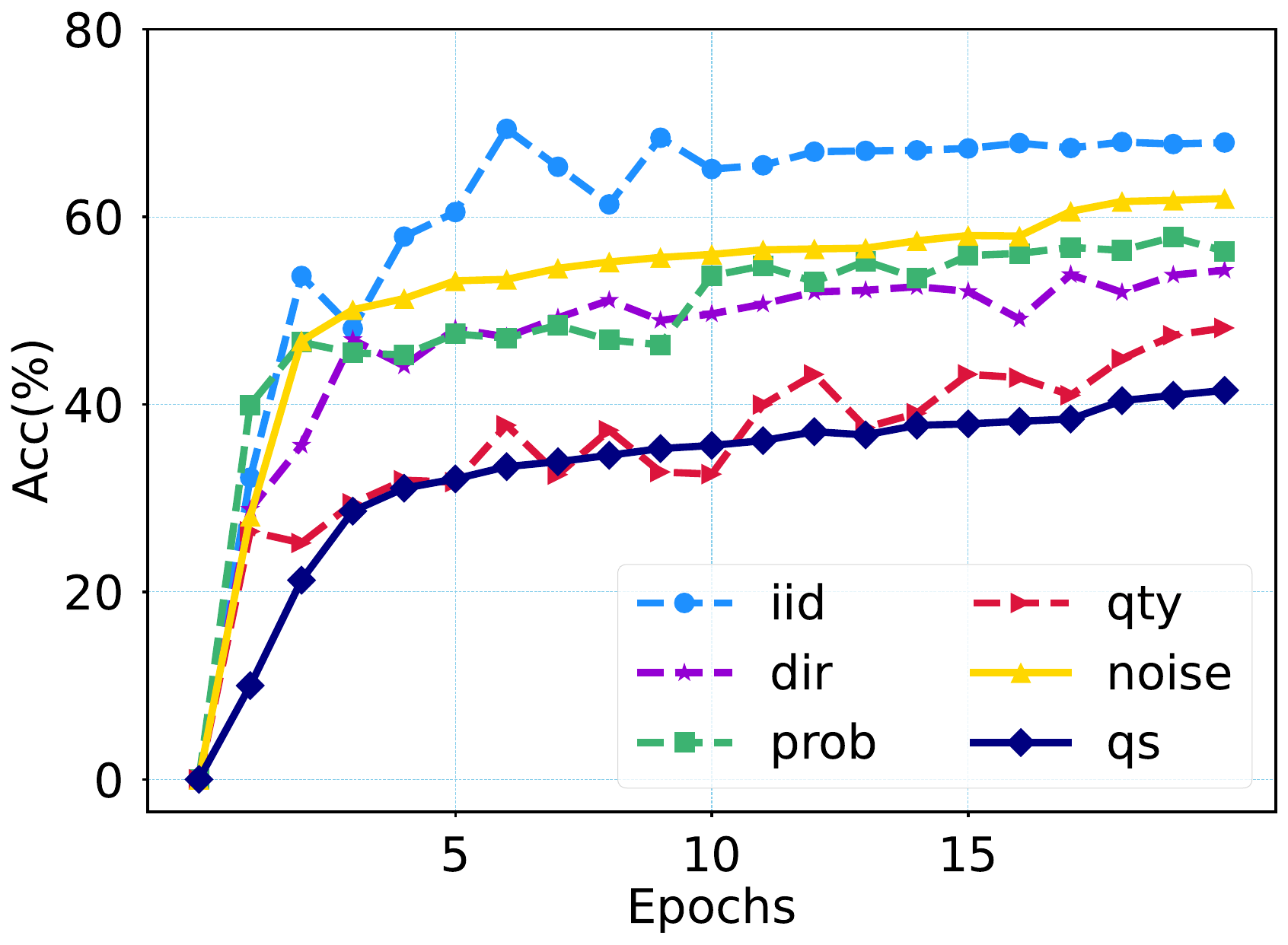}
        \caption{FLARE~\cite{wang2022flare}}
        \label{fig:noniid_degree_FLARE}
    \end{subfigure}
    \begin{subfigure}[b]{0.19\textwidth}
        \includegraphics[width=\textwidth]{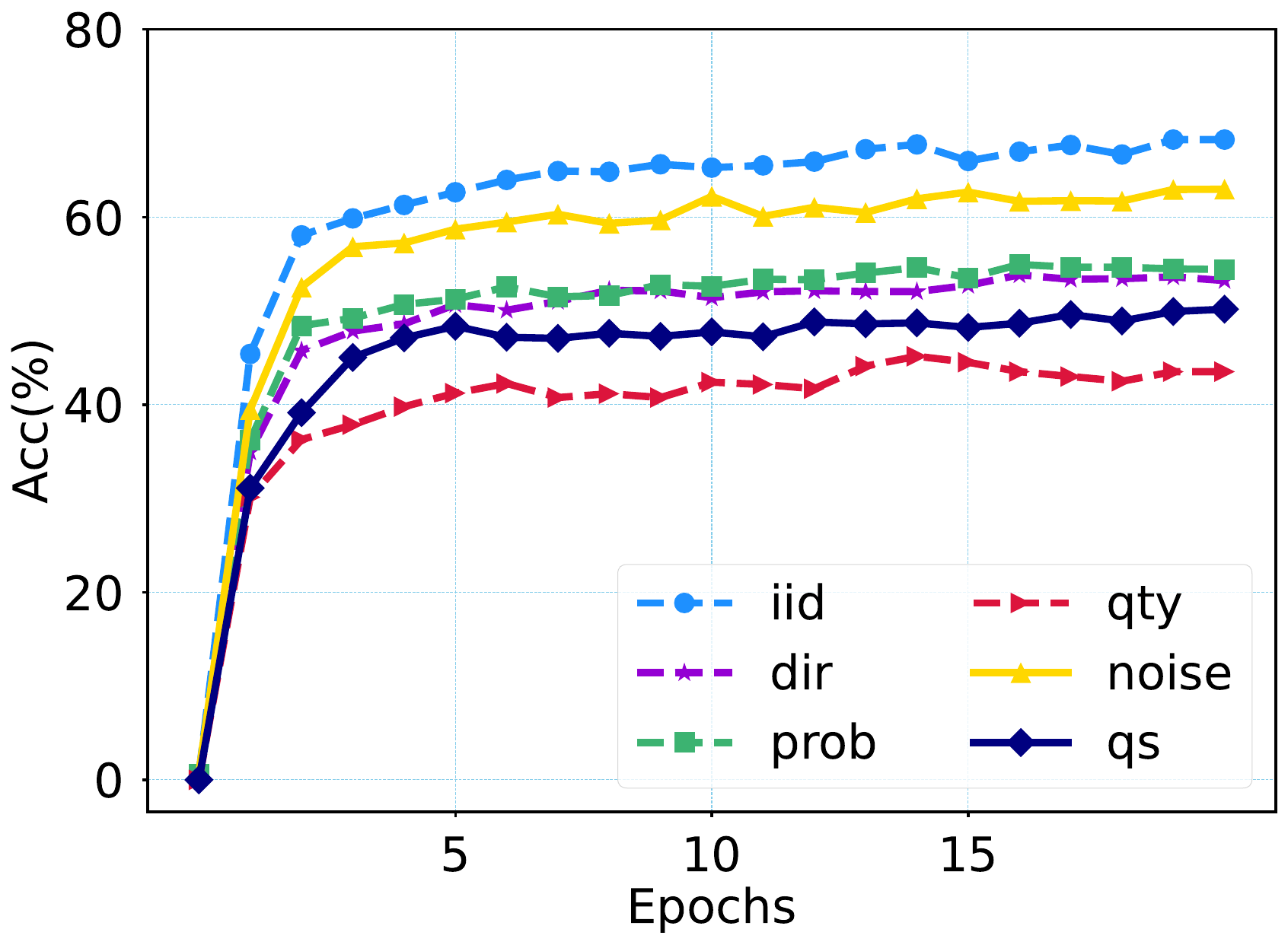}
        \caption{FLAME~\cite{nguyen2022flame}}
        \label{fig:noniid_degree_FLAME}
    \end{subfigure}
    \begin{subfigure}[b]{0.19\textwidth}
        \includegraphics[width=\textwidth]{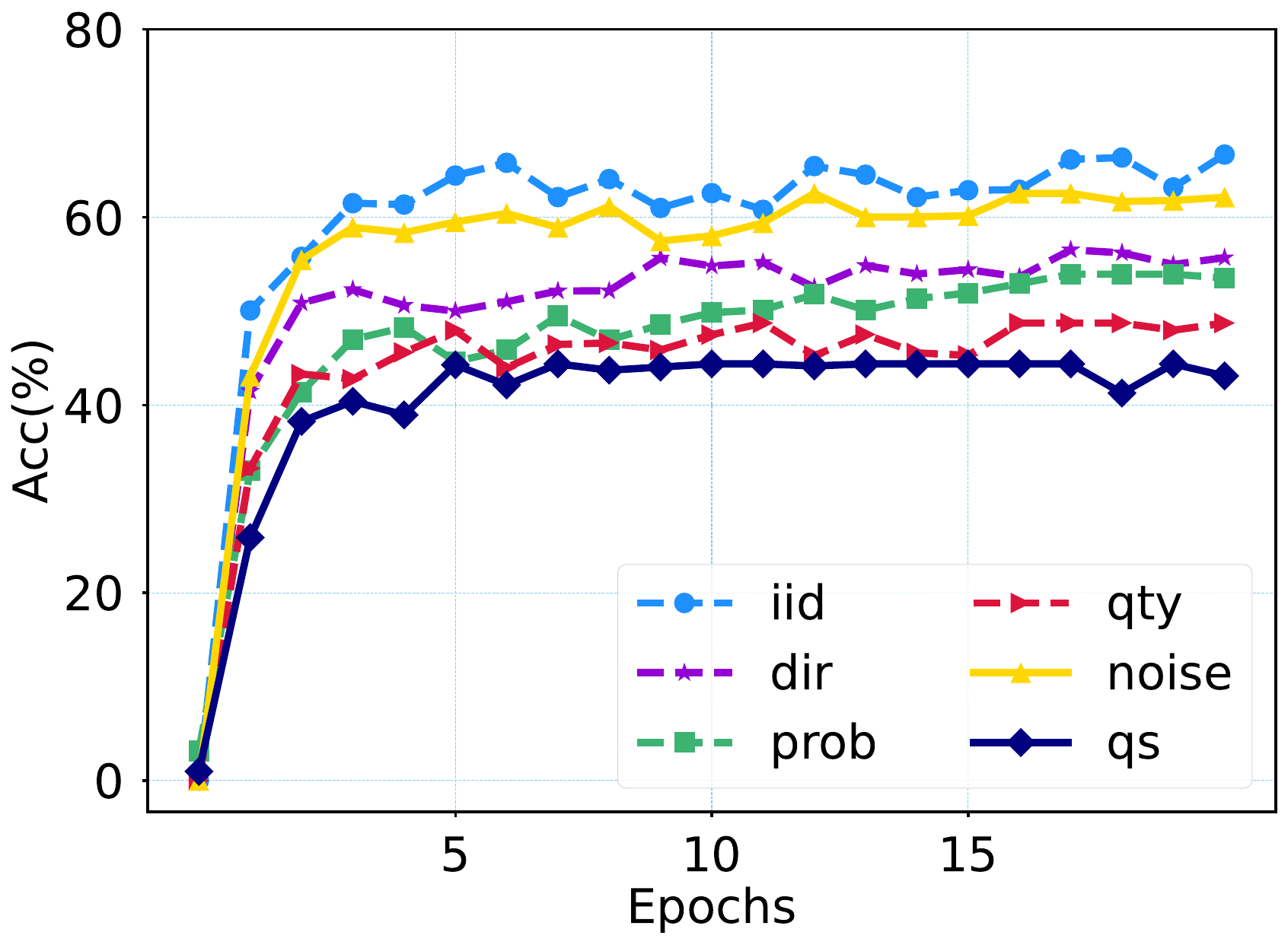}
        \caption{FLDetector~\cite{zhangfldetecotr}}
        \label{fig:noniid_degree_FLDetector}
    \end{subfigure}
    \begin{subfigure}[b]{0.19\textwidth}
        \includegraphics[width=\textwidth]{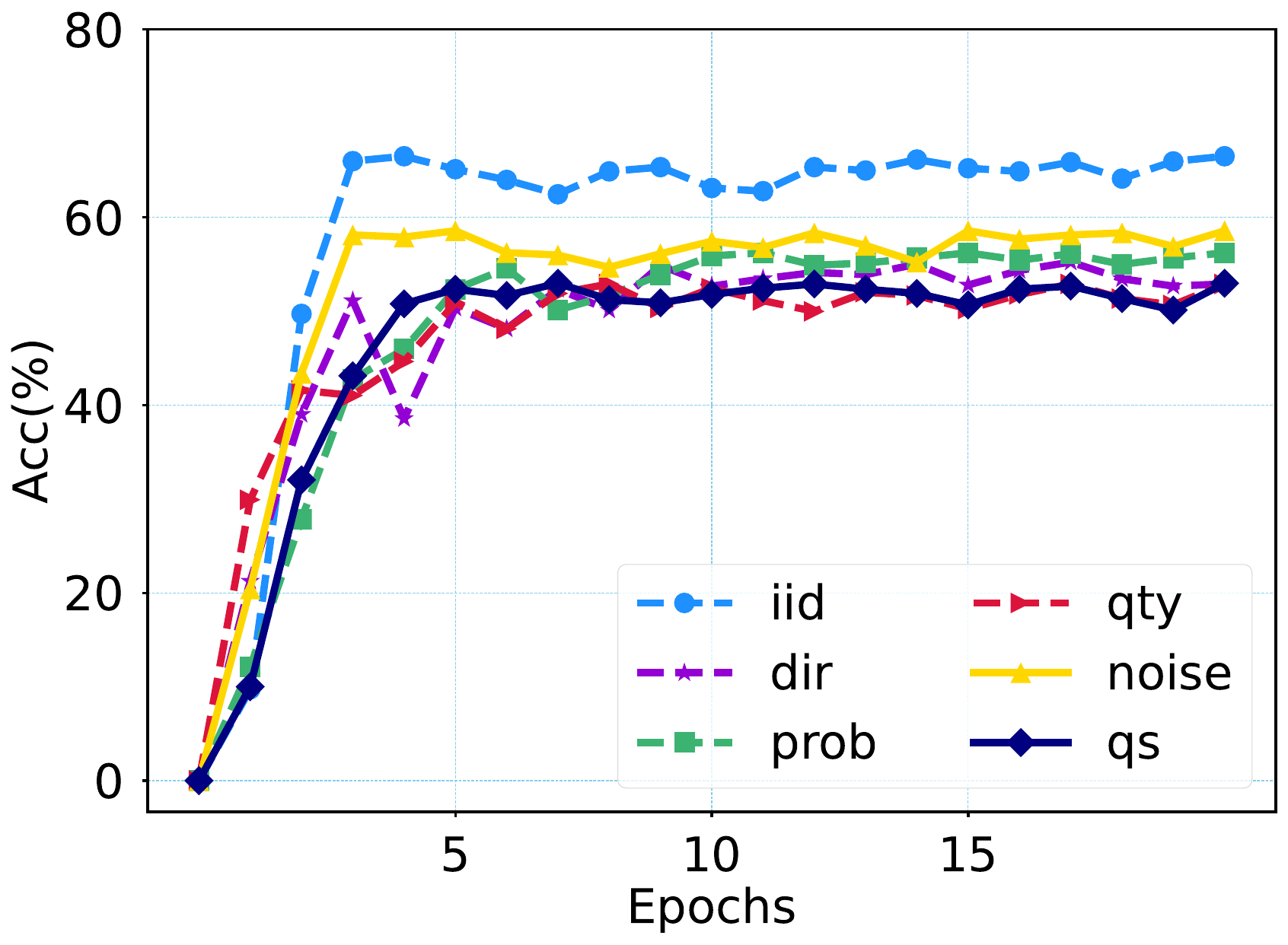}
        \caption{FLIP~\cite{zhang2023flipprovabledefenseframework}}
        \label{fig:noniid_degree_FLIP}
    \end{subfigure}
    
    \begin{subfigure}[b]{0.19\textwidth}
        \includegraphics[width=\textwidth]{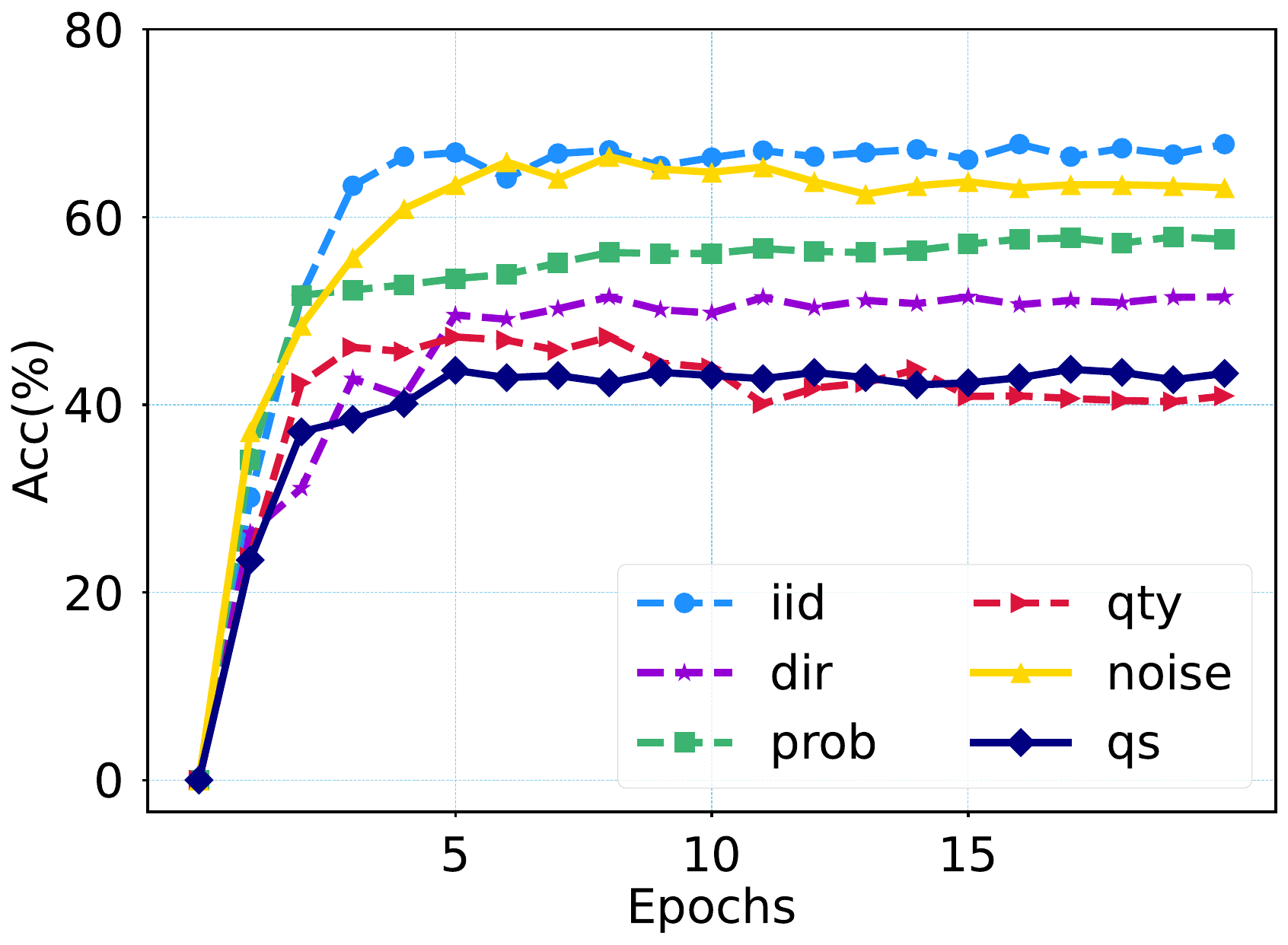}
        \caption{FLGuard~\cite{lee2023flguard}}
        \label{fig:noniid_degree_FLGuard}
    \end{subfigure}
    \begin{subfigure}[b]{0.19\textwidth}
        \includegraphics[width=\textwidth]{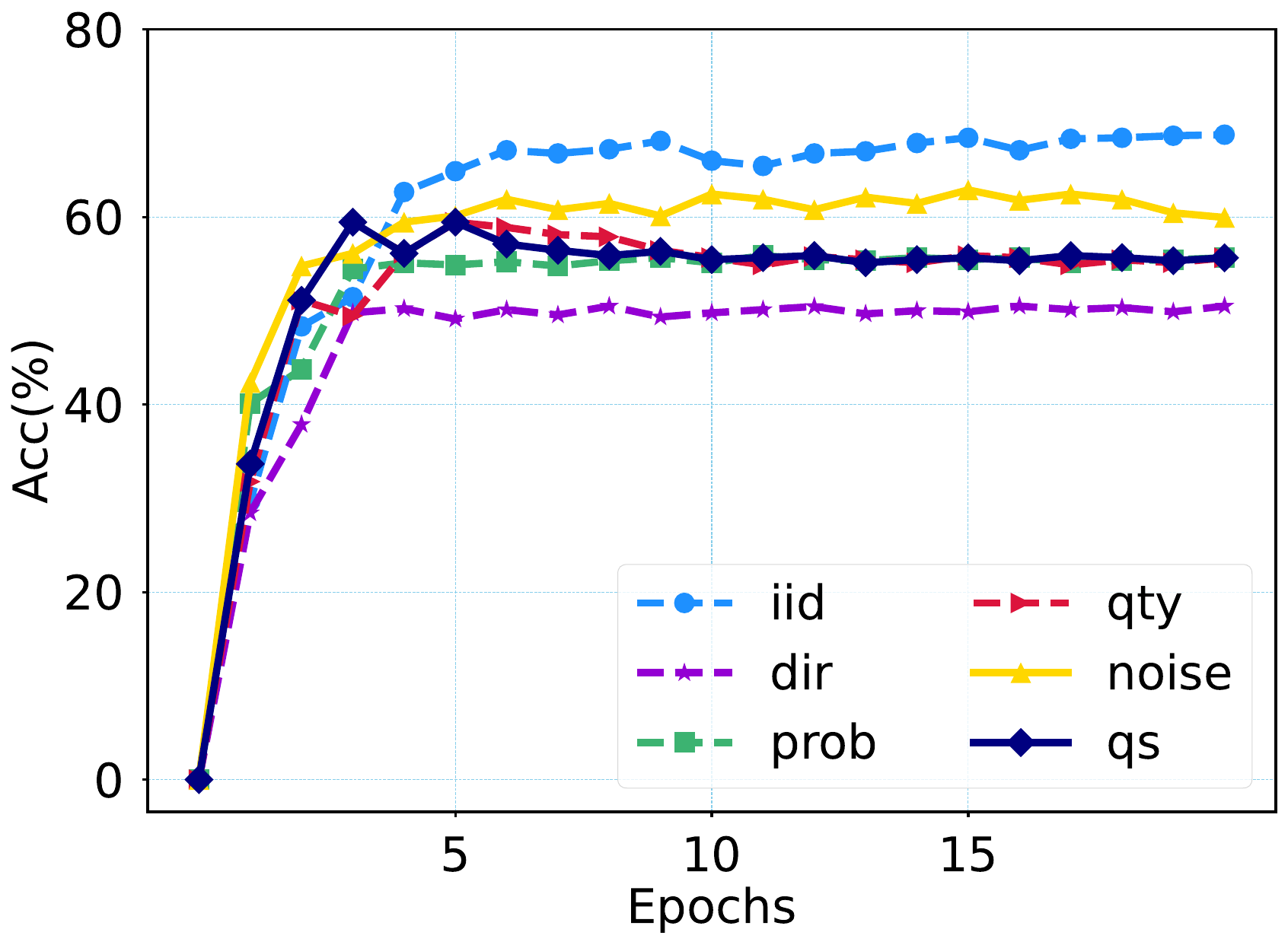}
        \caption{MESAS~\cite{MESAS}}
        \label{fig:noniid_degree_MESAS}
    \end{subfigure}
    \begin{subfigure}[b]{0.19\textwidth}
        \includegraphics[width=\textwidth]{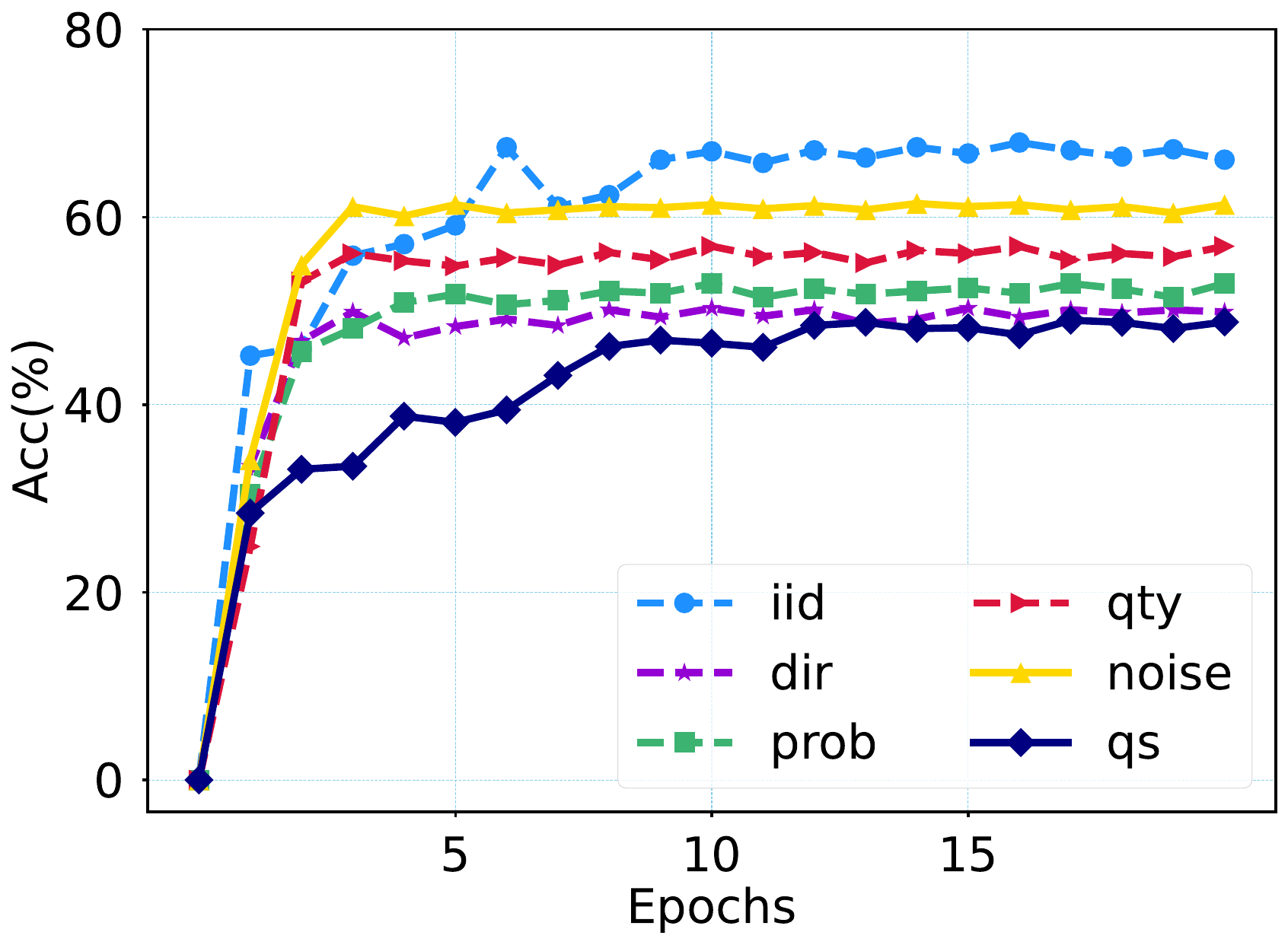}
        \caption{FreqFed~\cite{fereidooni2023freqfed}}
        \label{fig:noniid_degree_FreqFed}
    \end{subfigure}
    \begin{subfigure}[b]{0.19\textwidth}
        \includegraphics[width=\textwidth]{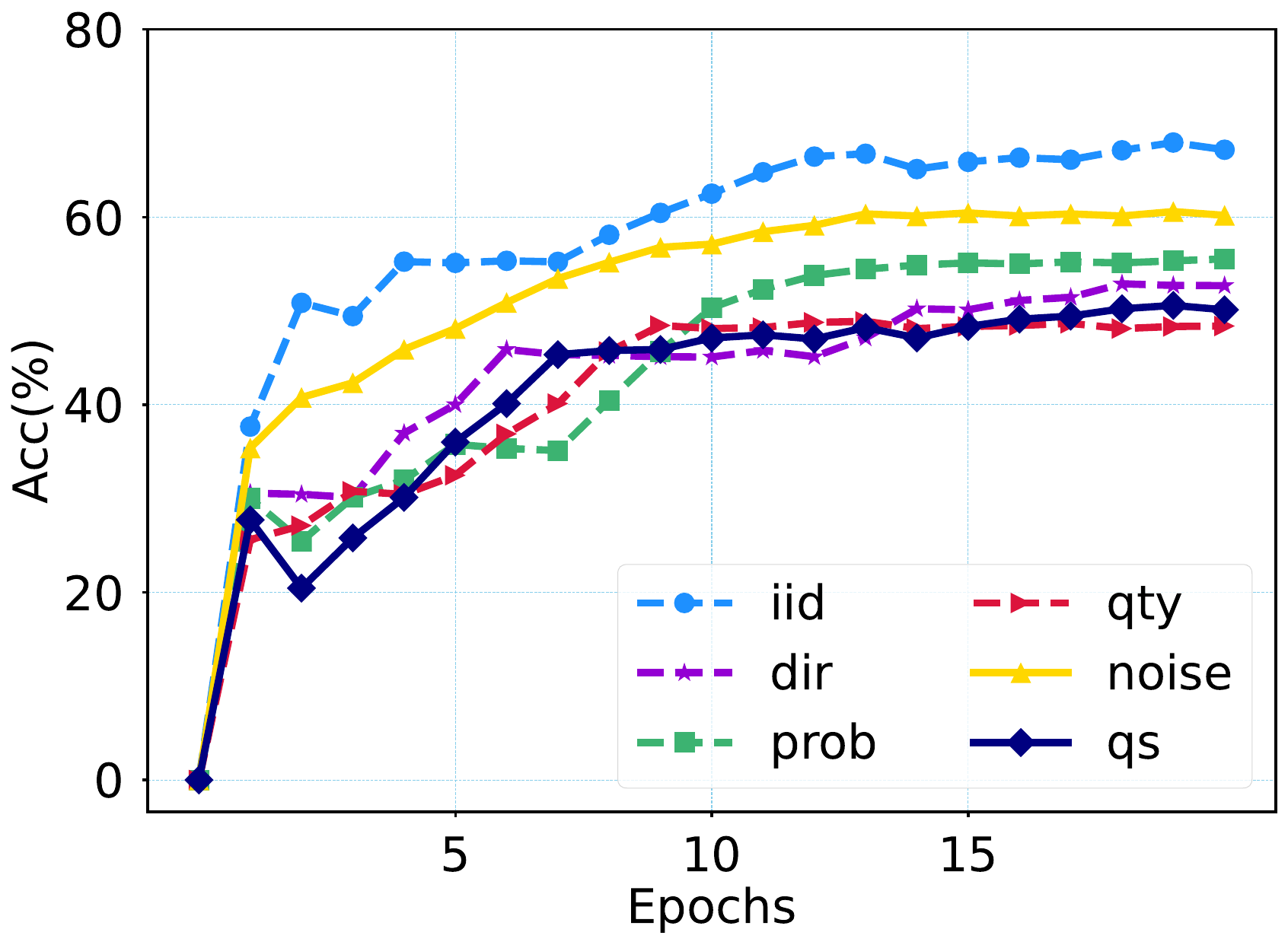}
        \caption{AGSD~\cite{ali2024adversarially}}
        \label{fig:noniid_degree_AGSD}
    \end{subfigure}
    \begin{subfigure}[b]{0.19\textwidth}
        \includegraphics[width=\textwidth]{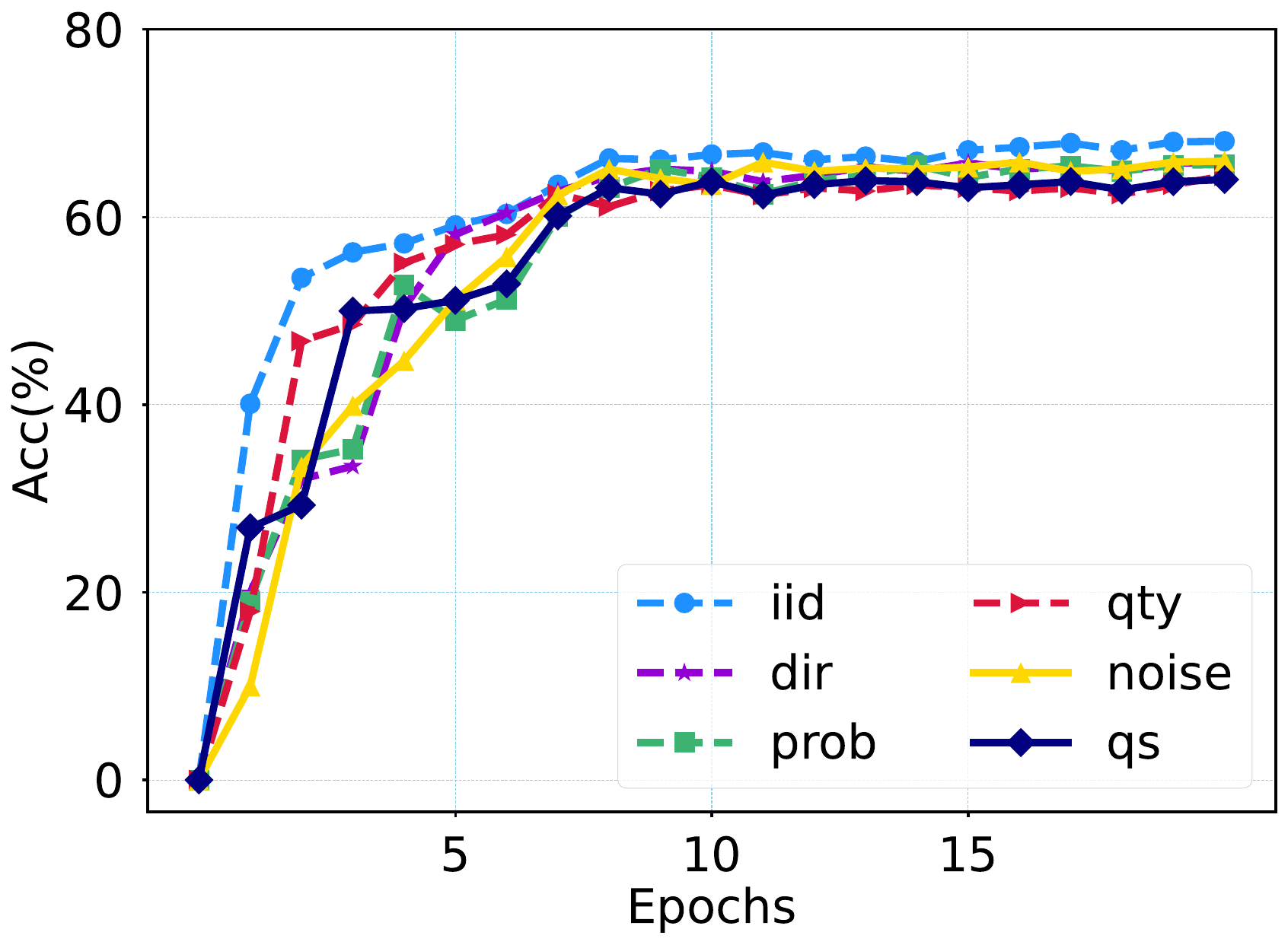}
        \caption{GeminiGuard (Ours)}
        \label{fig:noniid_degree_FLAG}
    \end{subfigure}

    \vspace{-5pt}
    \caption{Comparison of \NIID impacts on various FL defenses for CIFAR-10, Min-Max Attack~\cite{shejwalkar2021manipulating}.}
    \label{fig:umpa-baselines}
    \vspace{-5pt}
\end{figure*}

\textbf{Untrageted \MPAs.}
Fig.~\ref{fig:umpa-baselines} compares the model accuracy (Acc) of the SOTA \mpg defenses with those of \FG on the CIFAR-10 dataset under the Min-Max attack in both IID and \NIIDs settings. In the IID scenario, all defenses perform well, maintaining high Acc values, demonstrating their effectiveness in such settings. However, significant performance drops are observed in some \NIIDs  for certain defenses. For example, AGSD—a recently proposed defense—fails to preserve model performance in `qty' and `qs'. Similar trends can be observed in other SOTA defenses like FLGuard and FreqFed, as they assume IID data or only account for limited real-world data distribution scenarios.
However, it is evident that \FG effectively preserves the fidelity of the global model, maintaining a consistently high accuracy in all  settings. This demonstrates the robustness of \FG in defending against \mpa across various \NIIDs. 

Table~\ref{tab:untargeted-flag} further highlights the model accuracy of \FG on MNIST, FMNIST, CIFAR-10, and Sentiment-140 datasets under three \umpas in both IID and \NIIDs settings. In all cases, \FG shows only a minimal accuracy drop compared to the clean model (FedAvg, No Attack) reported in Table~\ref{tab:acc_all_benigh} and consistently resists different MPAs.

\begin{table*}[ht]
\centering
\renewcommand{\arraystretch}{1.2} 
\caption{\texttt{Model Acc} (\%) of untargeted \MPAs with \FG  under various Non-IID settings.}
\label{tab:untargeted-flag}
\resizebox{\textwidth}{!}{
\begin{tabular}{c|cccccc|cccccc|cccccc}
\hline
\multirow{2}{*}{\textbf{Dataset}} & \multicolumn{6}{c|}{\textbf{Krum-Attack}} & \multicolumn{6}{c|}{\textbf{Trimmean-Attack}} & \multicolumn{6}{c}{\textbf{Min-Sum}} \\
\cline{2-19}
 & \textbf{iid} & \textbf{dir} & \textbf{prob} & \textbf{qty} & \textbf{noise} & \textbf{qs} 
 & \textbf{iid} & \textbf{dir} & \textbf{prob} & \textbf{qty} & \textbf{noise} & \textbf{qs} 
 & \textbf{iid} & \textbf{dir} & \textbf{prob} & \textbf{qty} & \textbf{noise} & \textbf{qs} \\
\hline
MNIST & 97.88 & 92.83 & 92.15 & 92.98 & 93.84 & 92.79 
      & 96.13 & 93.12 & 92.23 & 93.84 & 93.69 & 93.68 
      & 94.63 & 92.83 & 92.98 & 93.97 & 93.76 & 93.25 \\
FMNIST & 90.95 & 81.05 & 80.78 & 80.62 & 86.15 & 80.94 
       & 90.73 & 81.58 & 80.33 & 82.06 & 85.04 & 80.27 
       & 90.58 & 82.28 & 83.05 & 82.88 & 84.00 & 80.79 \\
CIFAR-10 & 67.74 & 65.40 & 65.74 & 62.90 & 66.50 & 63.14 
        & 68.11 & 66.95 & 64.72 & 62.64 & 66.71 & 62.62 
        & 68.19 & 65.69 & 64.10 & 62.50 & 66.90 & 64.57 \\
Sentiment-140 & 77.42 & 72.57 & 71.28 & 70.57 & 71.40 & 70.68 
              & 77.85 & 72.71 & 70.63 & 69.85 & 71.04 & 70.19 
              & 76.50 & 71.85 & 70.33 & 71.18 & 69.41 & 70.85 \\
\hline
\end{tabular}
}
\vspace{-10pt}
\end{table*}

\textbf{Backdoor Attacks.}
Table~\ref{tab:noniid_backdoor} compares the attack success rate (\texttt{ASR}) of SOTA backdoor defenses with those of \FG across FMNIST, CIFAR-10, and Sentiment-140 datasets under five backdoor attack types in both IID and \NIIDs settings. Several key insights emerge from these results. Firstly, nearly all SOTA defenses perform well in defending against backdoor attacks in IID settings, with \texttt{ASR} values below 10\%. This demonstrates their effectiveness even against recent and sophisticated attacks such as Neurotoxin~\cite{zhang2022neurotoxin} and IBA~\cite{nguyen2024iba}. Secondly, while some SOTA defenses perform well in certain \NIIDs, their effectiveness diminishes in others. For instance, in the CIFAR-10 dataset under the IBA attack, the \texttt{ASR} of FreqFed is 6.58\% in `prob' and 8.11\% in `noise', but rises significantly to 13.58\% in `dir', 17.84\% in `qty', and 19.98\% in `qs', illustrating inconsistent performance across different data distributions. Furthermore, across all \NIIDs, the `noise' setting consistently has the least impact on model performance, suggesting it poses fewer challenges compared to other \NIIDs.

Table~\ref{tab:noniid_backdoor} also highlights that \FG consistently achieves the lowest \texttt{ASR} across all datasets and attack types, significantly outperforming other defenses. This underscores its robustness against diverse and challenging attack scenarios. Unlike other defenses, whose \texttt{ASR} often exceeds 10\% or even 20\% in most \NIIDs, \FG maintains exceptionally low \texttt{ASR} values. For instance, when defending against Edge-case attacks on FMNIST, \FG achieves \texttt{ASR} values of 0.09\% (IID), 0.12\% (`dir'), 0.30\% (`prob'), 0.32\% (`qty'), 0.30\% (`noise'), and 0.60\% (`qs'). Similar results can be observed for CIFAR-10 and Sentiment-140, further highlighting \FG's resilience under challenging \NIIDs. These results demonstrate that \FG effectively handles various \NIIDs, a challenge where most prior defenses fall short, providing strong evidence to support the core intuition presented in Sect.~\ref{sec:method}.

\begin{table*}[ht]
\centering
\Huge
\renewcommand{\arraystretch}{1.4} 
\caption{\texttt{ASR} (\%) of backdoor attacks with SOTA defenses under various Non-IID settings.}
\label{tab:noniid_backdoor}

\begin{subtable}[t]{\textwidth}
\centering
\caption{FMNIST}
\resizebox{\textwidth}{!}{
\begin{tabular}{c|cccccc|cccccc|cccccc|cccccc|cccccc}
\hline
\multirow{2}{*}{\textbf{Defense}} & \multicolumn{6}{c|}{\textbf{BadNets}} & \multicolumn{6}{c|}{\textbf{DBA}}  & \multicolumn{6}{c|}{\textbf{Edge-case}} & \multicolumn{6}{c|}{\textbf{Neurotoxin}} & \multicolumn{6}{c}{\textbf{IBA}} \\
\cline{2-31}
 & \textbf{iid} & \textbf{dir} & \textbf{prob} & \textbf{qty} & \textbf{noise} & \textbf{qs}  
 & \textbf{iid} & \textbf{dir} & \textbf{prob} & \textbf{qty} & \textbf{noise} & \textbf{qs} 
 & \textbf{iid} & \textbf{dir} & \textbf{prob} & \textbf{qty} & \textbf{noise} & \textbf{qs} 
 & \textbf{iid} & \textbf{dir} & \textbf{prob} & \textbf{qty} & \textbf{noise} & \textbf{qs} 
 & \textbf{iid} & \textbf{dir} & \textbf{prob} & \textbf{qty} & \textbf{noise} & \textbf{qs} \\ \hline

 \textbf{FLTrust} & \default{6.80} & \default{9.63} & \default{5.64} & \default{22.42} & \default{8.70} & \default{19.66}
                 & \default{2.30} & \default{13.22} & \default{9.19} & \default{29.19} & \default{7.43} & \default{26.77}
                 & \default{3.30} & \default{15.88} & \default{11.60} & \default{13.90} & \default{9.80} & \default{13.77}
                 & \default{5.80} & \default{12.75} & \default{10.10} & \default{20.57} & \default{6.79} & \default{15.66}
                 & \default{1.20} & \default{11.15} & \default{6.80} & \default{24.70} & \default{7.30} & \default{13.79} \\

\textbf{FLARE}  & \default{1.85} & \default{7.04} & \default{10.00} & \default{7.91} & \default{6.16} & \default{22.04}
                & \default{2.66} & \default{10.22} & \default{10.15} & \default{7.28} & \default{7.45} & \default{13.94}
                & \default{3.71} & \default{11.24} & \default{7.54} & \default{2.81} & \default{6.81} & \default{7.16}
                & \default{2.08} & \default{8.19} & \default{10.88} & \default{6.20} & \default{7.61} & \default{13.28}
                & \default{6.56} & \default{14.43} & \default{12.45} & \default{5.30} & \default{7.15} & \default{13.83} \\

\textbf{FLAME} & \default{0.25} & \default{1.75} & \default{1.25} & \default{9.01} & \default{4.88} & \default{12.55}
               & \default{0.42} & \default{1.50} & \default{7.90} & \default{10.27} & \default{3.48} & \default{7.33}
               & \default{3.64} & \default{3.50} & \default{5.26} & \default{4.06} & \default{7.44} & \default{7.26}
               & \default{0.24} & \default{2.00} & \default{0.76} & \default{41.66} & \default{4.75} & \default{18.22}
               & \default{0.36} & \default{4.61} & \default{3.63} & \default{11.92} & \default{10.54} & \default{10.64} \\

\textbf{FLDetector} & \default{2.02} & \default{15.10} & \default{7.76} & \default{13.90} & \default{6.63} & \default{29.05}
                    & \default{1.76} & \default{9.02} & \default{5.37} & \default{21.54} & \default{8.39} & \default{26.68}
                    & \default{8.14} & \default{8.08} & \default{4.38} & \default{11.52} & \default{10.13} & \default{36.95}
                    & \default{6.15} & \default{16.00} & \default{5.26} & \default{18.60} & \default{7.51} & \default{35.82}
                    & \default{0.12} & \default{24.01} & \default{7.01} & \default{12.72} & \default{8.39} & \default{37.69} \\

\textbf{FLIP} & \topone{0.00} & \default{10.68} & \default{4.01} & \default{8.34} & \default{5.16} & \default{10.57}
              & \default{1.13} & \default{7.83} & \default{6.42} & \default{8.06} & \default{8.35} & \default{18.01}
              & \default{0.94} & \default{12.58} & \default{7.34} & \default{21.48} & \default{5.04} & \default{22.92}
              & \default{0.37} & \default{14.87} & \default{6.70} & \default{20.06} & \default{5.12} & \default{11.64}
              & \default{0.62} & \default{6.86} & \default{4.43} & \default{11.94} & \default{5.83} & \default{22.48} \\

\textbf{FLGuard} & \default{3.56} & \default{10.32} & \default{18.14} & \default{23.05} & \default{10.58} & \default{41.40}
                 & \default{5.35} & \default{15.26} & \default{12.76} & \default{21.12} & \default{9.80} & \default{27.04}
                 & \default{4.12} & \default{13.12} & \default{16.16} & \default{10.53} & \default{9.60} & \default{32.40}
                 & \default{2.26} & \default{7.61} & \default{10.42} & \default{22.85} & \default{9.86} & \default{31.80}
                 & \default{4.91} & \default{21.43} & \default{28.68} & \default{25.92} & \default{8.59} & \default{24.12} \\

\textbf{MESAS} & \default{1.00} & \default{12.62} & \default{10.98} & \default{12.00} & \default{3.48} & \default{10.04}
               & \default{1.37} & \default{4.92} & \default{7.57} & \default{20.84} & \default{10.10} & \default{10.09}
               & \default{1.94} & \default{10.95} & \default{9.20} & \default{24.91} & \default{6.12} & \default{9.76}
               & \default{0.89} & \default{7.28} & \default{11.58} & \default{15.29} & \default{9.84} & \default{13.76}
               & \default{3.10} & \default{16.75} & \default{7.79} & \default{25.38} & \default{9.53} & \default{18.58} \\

\textbf{FreqFed} & \default{1.66} & \default{6.89} & \default{3.12} & \default{37.94} & \default{6.88} & \default{25.08}
                 & \default{1.69} & \default{14.52} & \default{7.38} & \default{27.36} & \default{8.37} & \default{27.12}
                 & \default{2.12} & \default{9.75} & \default{5.99} & \default{17.94} & \default{7.12} & \default{15.32}
                 & \default{4.81} & \default{15.14} & \default{5.49} & \default{20.92} & \default{3.88} & \default{21.49}
                 & \default{2.98} & \default{15.42} & \default{5.46} & \default{14.27} & \default{10.88} & \default{15.98} \\

\textbf{AGSD} & \default{0.48} & \default{4.88} & \default{0.88} & \default{17.04} & \default{2.82} & \default{10.90}
              & \default{1.19} & \default{3.77} & \default{0.49} & \default{13.88} & \default{5.47} & \default{19.44}
              & \default{0.62} & \default{4.01} & \default{0.48} & \default{12.60} & \default{2.52} & \default{9.78}
              & \default{0.12} & \default{2.12} & \topone{0.12} & \default{10.02} & \default{2.33} & \default{12.70}
              & \topone{0.00} & \default{6.24} & \topone{0.12} & \default{11.54} & \default{2.28} & \default{9.13} \\

\textbf{GeminiGuard} & \default{0.20} & \topone{0.36} & \topone{0.26} & \topone{0.13} & \topone{0.24} & \topone{0.46} 
                     & \topone{0.17} & \topone{0.33} & \topone{0.11} & \topone{0.30} & \topone{0.13} & \topone{0.11}
                     & \topone{0.09} & \topone{0.12} & \topone{0.30} & \topone{0.32} & \topone{0.30} & \topone{0.60}
                     & \topone{0.00} & \topone{0.41} & \default{0.60} & \topone{0.32} & \topone{0.30} & \topone{0.71}
                     & \default{0.13} & \topone{0.22} & \default{0.45} & \topone{0.21} & \topone{0.16} & \topone{0.58} \\

\hline
\end{tabular}}
\end{subtable}
\vspace{-20pt}

\begin{subtable}[t]{\textwidth}
\centering
\label{tab:non-iid-backdoors-cifar}
\caption{CIFAR-10}
\resizebox{\textwidth}{!}{
\begin{tabular}{c|cccccc|cccccc|cccccc|cccccc|cccccc}
\hline
\multirow{2}{*}{\textbf{Defense}} & \multicolumn{6}{c|}{\textbf{BadNets}} & \multicolumn{6}{c|}{\textbf{DBA}}  & \multicolumn{6}{c|}{\textbf{Edge-case}} & \multicolumn{6}{c|}{\textbf{Neurotoxin}} & \multicolumn{6}{c}{\textbf{IBA}} \\
\cline{2-31}
 & \textbf{iid} & \textbf{dir} & \textbf{prob} & \textbf{qty} & \textbf{noise} & \textbf{qs}  
 & \textbf{iid} & \textbf{dir} & \textbf{prob} & \textbf{qty} & \textbf{noise} & \textbf{qs} 
 & \textbf{iid} & \textbf{dir} & \textbf{prob} & \textbf{qty} & \textbf{noise} & \textbf{qs} 
 & \textbf{iid} & \textbf{dir} & \textbf{prob} & \textbf{qty} & \textbf{noise} & \textbf{qs} 
 & \textbf{iid} & \textbf{dir} & \textbf{prob} & \textbf{qty} & \textbf{noise} & \textbf{qs} \\ \hline
\textbf{FLTrust} & \default{8.91} & \default{11.14} & \default{8.72} & \default{15.07} & \default{10.10} & \default{11.91}
                 & \default{6.26} & \default{11.37} & \default{8.58} & \default{12.92} & \default{9.64} & \default{17.56}
                 & \default{4.42} & \default{10.57} & \default{8.60} & \default{18.91} & \default{7.47} & \default{13.06}
                 & \default{6.00} & \default{15.17} & \default{7.09} & \default{13.41} & \default{8.70} & \default{16.97}
                 & \default{8.80} & \default{13.98} & \default{10.58} & \default{15.21} & \default{9.12} & \default{16.40} \\

\textbf{FLARE} & \default{4.30} & \default{9.15} & \default{13.00} & \default{10.28} & \default{8.01} & \default{28.65}
               & \default{5.90} & \default{13.29} & \default{13.20} & \default{11.46} & \default{9.69} & \default{18.12}
               & \default{9.82} & \default{14.61} & \default{12.80} & \default{3.65} & \default{8.85} & \default{9.31}
               & \default{10.70} & \default{34.30} & \default{14.14} & \default{8.06} & \default{9.89} & \default{17.26}
               & \default{8.53} & \default{18.76} & \default{10.18} & \default{15.00} & \default{11.29} & \default{17.98} \\

\textbf{FLAME} & \default{5.32} & \default{14.70} & \default{7.56} & \default{11.27} & \default{10.11} & \default{25.69}
               & \default{3.60} & \default{10.30} & \default{8.87} & \default{18.84} & \default{8.35} & \default{19.17}
               & \default{6.54} & \default{14.38} & \default{9.57} & \default{23.90} & \default{9.30} & \default{19.07}
               & \default{7.30} & \default{15.65} & \default{8.95} & \default{52.08} & \default{9.44} & \default{22.77}
               & \default{5.45} & \default{26.60} & \default{9.29} & \default{22.50} & \default{16.92} & \default{18.29} \\

\textbf{FLDetector} & \default{2.52} & \default{16.38} & \default{5.96} & \default{17.37} & \default{9.30} & \default{36.31}
                    & \default{0.95} & \default{13.78} & \default{4.21} & \default{26.92} & \default{10.48} & \default{33.34}
                    & \default{10.17} & \default{21.70} & \default{5.47} & \default{14.40} & \default{8.91} & \default{46.19}
                    & \default{7.93} & \default{20.19} & \default{6.57} & \default{14.09} & \default{9.39} & \default{44.78}
                    & \default{4.15} & \default{15.01} & \default{8.76} & \default{20.90} & \default{10.48} & \default{47.12} \\

\textbf{FLIP} & \default{3.50} & \default{12.10} & \default{10.01} & \default{14.40} & \default{9.45} & \default{23.21}
              & \default{1.41} & \default{37.70} & \default{8.02} & \default{32.19} & \default{10.44} & \default{22.52}
              & \default{3.17} & \default{13.22} & \default{9.18} & \default{26.85} & \default{6.55} & \default{16.16}
              & \default{0.46} & \default{21.34} & \default{8.37} & \default{25.08} & \default{5.15} & \default{14.55}
              & \default{0.78} & \default{18.33} & \default{5.54} & \default{14.92} & \default{10.54} & \default{28.09} \\

\textbf{FLGuard} & \default{10.10} & \default{96.50} & \default{84.80} & \default{28.82} & \default{16.98} & \default{51.75}
                 & \default{6.69} & \default{8.57} & \default{38.45} & \default{26.40} & \default{10.25} & \default{21.30}
                 & \default{6.15} & \default{12.66} & \default{17.70} & \default{20.21} & \default{14.50} & \default{40.50}
                 & \default{6.58} & \default{10.76} & \default{13.02} & \default{28.56} & \default{12.33} & \default{39.75}
                 & \default{6.13} & \default{10.24} & \default{25.00} & \default{29.91} & \default{10.74} & \default{30.15} \\

\textbf{MESAS} & \default{3.27} & \default{9.50} & \default{13.73} & \default{15.00} & \default{8.35} & \default{17.54}
               & \default{4.21} & \default{16.15} & \default{9.46} & \default{26.05} & \default{12.63} & \default{16.36}
               & \default{9.92} & \default{8.69} & \default{11.50} & \default{31.14} & \default{7.65} & \default{10.95}
               & \default{3.11} & \default{9.11} & \default{14.48} & \default{19.11} & \default{7.80} & \default{24.70}
               & \default{5.60} & \default{10.63} & \default{9.73} & \default{31.72} & \default{11.91} & \default{20.73} \\

\textbf{FreqFed} & \default{2.07} & \default{17.50} & \default{6.66} & \default{47.43} & \default{8.60} & \default{31.35}
                 & \default{3.11} & \default{23.50} & \default{10.97} & \default{71.70} & \default{10.21} & \default{71.40}
                 & \default{2.66} & \default{9.69} & \default{5.74} & \default{17.43} & \default{10.90} & \default{22.90}
                 & \default{6.01} & \default{16.42} & \default{8.61} & \default{26.14} & \default{4.84} & \default{26.87}
                 & \default{5.00} & \default{13.78} & \default{6.58} & \default{17.84} & \default{8.11} & \default{19.98} \\

\textbf{AGSD} & \default{5.60} & \default{16.11} & \default{7.09} & \default{21.30} & \default{8.53} & \default{19.50}
              & \default{1.48} & \default{14.71} & \default{3.61} & \default{17.36} & \default{6.84} & \default{34.20}
              & \default{0.78} & \default{15.01} & \default{2.60} & \default{15.75} & \default{6.15} & \default{12.23}
              & \default{0.95} & \default{12.66} & \default{4.15} & \default{12.52} & \default{6.91} & \default{15.87}
              & \topone{0.60} & \default{7.80} & \default{7.15} & \default{24.43} & \default{6.85} & \default{13.70} \\

\textbf{GeminiGuard} & \topone{0.17} & \topone{0.66} & \topone{0.18} & \topone{0.59} & \topone{0.17} & \topone{0.62}
                     & \topone{0.15} & \topone{0.25} & \topone{0.08} & \topone{0.22} & \topone{0.39} & \topone{0.09}
                     & \topone{0.06} & \topone{0.07} & \topone{0.62} & \topone{0.27} & \topone{0.23} & \topone{0.80}
                     & \topone{0.00} & \topone{0.25} & \topone{0.46} & \topone{0.25} & \topone{0.21} & \topone{0.46}
                     & \default{0.98} & \topone{0.16} & \topone{0.31} & \topone{0.18} & \topone{0.63} & \topone{1.00} \\

\hline
\end{tabular}}
\end{subtable}
\vspace{-20pt}

\begin{subtable}[t]{\textwidth}
\centering
\caption{Sentiment-140}
\resizebox{\textwidth}{!}{
\begin{tabular}{c|cccccc|cccccc|cccccc|cccccc|cccccc}
\hline
\multirow{2}{*}{\textbf{Defense}} & \multicolumn{6}{c|}{\textbf{BadNets}} & \multicolumn{6}{c|}{\textbf{DBA}}  & \multicolumn{6}{c|}{\textbf{Edge-case}} & \multicolumn{6}{c|}{\textbf{Neurotoxin}} & \multicolumn{6}{c}{\textbf{IBA}} \\
\cline{2-31}
 & \textbf{iid} & \textbf{dir} & \textbf{prob} & \textbf{qty} & \textbf{noise} & \textbf{qs}  
 & \textbf{iid} & \textbf{dir} & \textbf{prob} & \textbf{qty} & \textbf{noise} & \textbf{qs} 
 & \textbf{iid} & \textbf{dir} & \textbf{prob} & \textbf{qty} & \textbf{noise} & \textbf{qs} 
 & \textbf{iid} & \textbf{dir} & \textbf{prob} & \textbf{qty} & \textbf{noise} & \textbf{qs} 
 & \textbf{iid} & \textbf{dir} & \textbf{prob} & \textbf{qty} & \textbf{noise} & \textbf{qs} \\ \hline
 \textbf{FLTrust} & \default{3.33} & \default{46.66} & \default{16.66} & \default{50.83} & \default{19.16} & \default{74.16} & \default{2.30} & \default{13.22} & \default{7.50} & \default{41.66} & \default{10.83} & \default{34.16} & \default{6.30} & \default{85.00} & \default{31.66} & \default{48.33} & \default{20.83} & \default{90.80} & \default{8.80} & \default{77.50} & \default{16.66} & \default{60.00} & \default{55.00} & \default{31.66} & \default{10.20} & \default{36.66} & \default{30.80} & \default{49.16} & \default{21.60} & \default{45.80} \\

\textbf{FLARE} & \default{6.52} & \default{12.00} & \default{44.16} & \default{52.50} & \default{10.00} & \default{40.00} & \default{3.81} & \default{15.66} & \default{55.00} & \default{24.00} & \default{45.80} & \default{20.00} & \default{7.50} & \default{20.00} & \default{60.00} & \default{10.00} & \default{20.00} & \default{16.80} & \default{8.80} & \default{20.00} & \default{19.00} & \default{12.00} & \default{25.00} & \default{11.66} & \default{8.77} & \default{16.66} & \default{20.00} & \default{10.00} & \default{19.00} & \default{20.00} \\

\textbf{FLAME} & \topone{0.76} & \default{20.00} & \default{11.35} & \default{15.29} & \default{11.85} & \default{19.44} & \default{0.94} & \default{15.06} & \default{11.00} & \default{20.00} & \default{10.00} & \default{14.60} & \default{0.78} & \default{13.74} & \default{12.00} & \default{17.00} & \default{9.00} & \default{14.83} & \default{0.74} & \default{19.71} & \default{6.70} & \default{20.00} & \default{10.50} & \default{18.00} & \default{0.82} & \default{12.00} & \default{4.00} & \default{25.00} & \default{22.80} & \default{19.60} \\

\textbf{FLDetector} & \default{4.40} & \default{32.15} & \default{40.29} & \default{11.92} & \default{40.88} & \default{41.58} & \default{4.39} & \default{11.36} & \default{25.78} & \default{45.16} & \default{22.55} & \default{34.88} & \default{7.11} & \default{10.26} & \default{14.99} & \default{33.81} & \default{17.10} & \default{13.70} & \default{7.29} & \default{29.00} & \default{16.94} & \default{30.00} & \default{24.00} & \default{45.61} & \default{8.28} & \default{27.18} & \default{19.28} & \default{26.01} & \default{10.11} & \default{38.87} \\

\textbf{FLIP} & \default{4.83} & \default{9.35} & \default{34.04} & \default{17.41} & \default{9.12} & \default{14.47} & \default{2.81} & \default{10.00} & \default{13.71} & \default{17.50} & \default{33.08} & \default{24.10} & \default{6.01} & \default{16.90} & \default{24.47} & \default{12.67} & \default{16.99} & \default{60.83} & \default{4.66} & \default{18.00} & \default{37.90} & \default{20.48} & \default{12.88} & \default{30.39} & \default{6.39} & \default{10.60} & \default{18.86} & \default{23.36} & \default{10.30} & \default{27.50} \\

\textbf{FLGuard} & \default{9.64} & \default{97.50} & \default{57.76} & \default{65.13} & \default{12.01} & \default{91.60} & \default{5.23} & \default{53.62} & \default{38.86} & \default{36.51} & \default{29.31} & \default{41.59} & \default{10.45} & \default{41.27} & \default{58.27} & \default{30.02} & \default{36.71} & \default{55.00} & \default{7.71} & \default{38.79} & \default{27.92} & \default{41.47} & \default{24.88} & \default{40.02} & \default{9.06} & \default{27.90} & \default{13.93} & \default{38.44} & \default{24.28} & \default{67.30} \\

\textbf{MESAS} & \default{2.52} & \default{9.32} & \default{10.36} & \default{30.66} & \default{7.65} & \default{13.07} & \default{2.12} & \default{10.97} & \default{18.71} & \default{10.52} & \default{20.56} & \default{23.10} & \default{5.00} & \default{12.73} & \default{18.41} & \default{20.00} & \default{10.88} & \default{36.66} & \default{5.42} & \default{14.61} & \default{20.78} & \default{22.68} & \default{11.14} & \default{27.93} & \default{9.92} & \default{16.64} & \default{14.39} & \default{27.83} & \default{10.00} & \default{54.16} \\

\textbf{FreqFed} & \default{6.24} & \default{18.45} & \default{10.73} & \default{50.36} & \default{15.00} & \default{71.03} & \default{3.57} & \default{23.45} & \default{14.00} & \default{12.31} & \default{12.66} & \default{15.70} & \default{20.86} & \default{63.58} & \default{23.75} & \default{50.77} & \default{20.01} & \default{27.97} & \default{3.33} & \default{22.64} & \default{16.93} & \default{37.01} & \default{11.65} & \default{13.69} & \default{7.86} & \default{19.00} & \default{13.29} & \default{25.47} & \default{11.20} & \default{18.70} \\

\textbf{AGSD} & \default{2.39} & \default{8.08} & \default{10.30} & \default{17.61} & \default{24.12} & \default{19.24} & \default{7.31} & \default{23.79} & \default{18.84} & \default{15.31} & \default{18.17} & \default{21.27} & \default{7.89} & \default{12.38} & \default{17.14} & \default{19.61} & \default{12.00} & \default{33.88} & \default{4.31} & \default{24.24} & \default{16.31} & \default{23.79} & \default{10.19} & \default{39.16} & \default{4.18} & \default{11.16} & \default{19.48} & \default{17.04} & \default{10.17} & \default{20.00} \\

\textbf{GeminiGuard} & \default{0.85} & \topone{0.96} & \topone{0.88} & \topone{0.97} & \topone{0.76} & \topone{1.01} & \topone{0.65} & \topone{0.82} & \topone{0.88} & \topone{0.90} & \topone{1.11} & \topone{1.06} & \topone{0.04} & \topone{0.59} & \topone{0.65} & \topone{0.83} & \topone{0.50} & \topone{0.72} & \topone{0.30} & \topone{0.71} & \topone{0.86} & \topone{0.57} & \topone{1.08} & \topone{1.00} & \topone{0.36} & \topone{0.50} & \topone{0.79} & \topone{1.10} & \topone{0.52} & \topone{0.94} \\

\hline
\end{tabular}}
\end{subtable}

\end{table*}

\subsection{Ablation Study}
To verify the robustness of \FG to FL Hyperparameters, we conduct ablation studies with \NIID degree, the number of malicious clients, data poisoning rate, auxiliary datasets, metrics and the number of layers.

\begin{figure*}[ht]
    \centering
    \begin{subfigure}[b]{0.19\textwidth}
        \includegraphics[width=\textwidth]{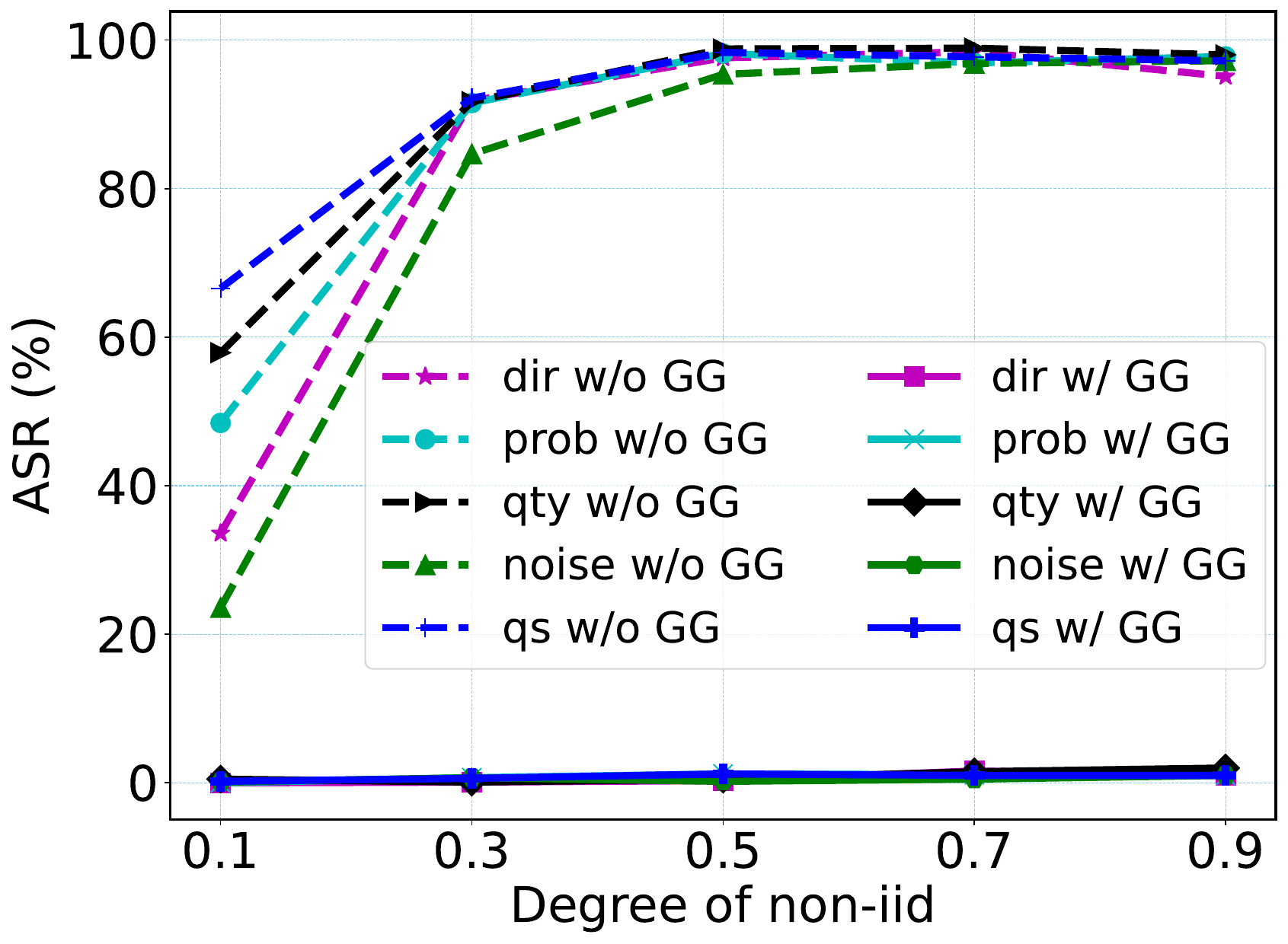}
        \caption{BadNets}
        \label{fig:noniid_degree_FLTrust}
    \end{subfigure}
    \begin{subfigure}[b]{0.19\textwidth}
        \includegraphics[width=\textwidth]{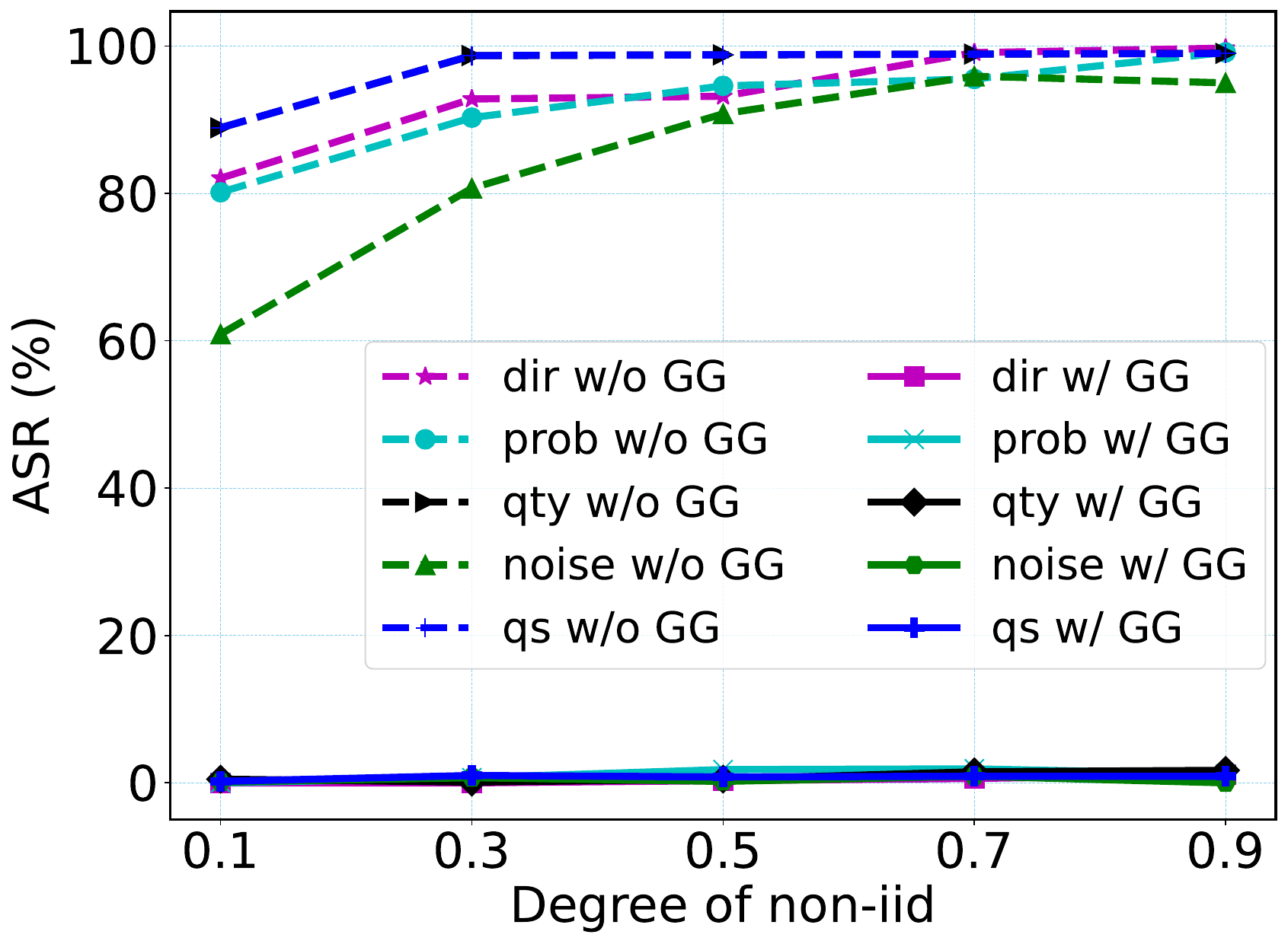}
        \caption{DBA}
        \label{fig:noniid_degree_FLDetector}
    \end{subfigure}
    \begin{subfigure}[b]{0.19\textwidth}
        \includegraphics[width=\textwidth]{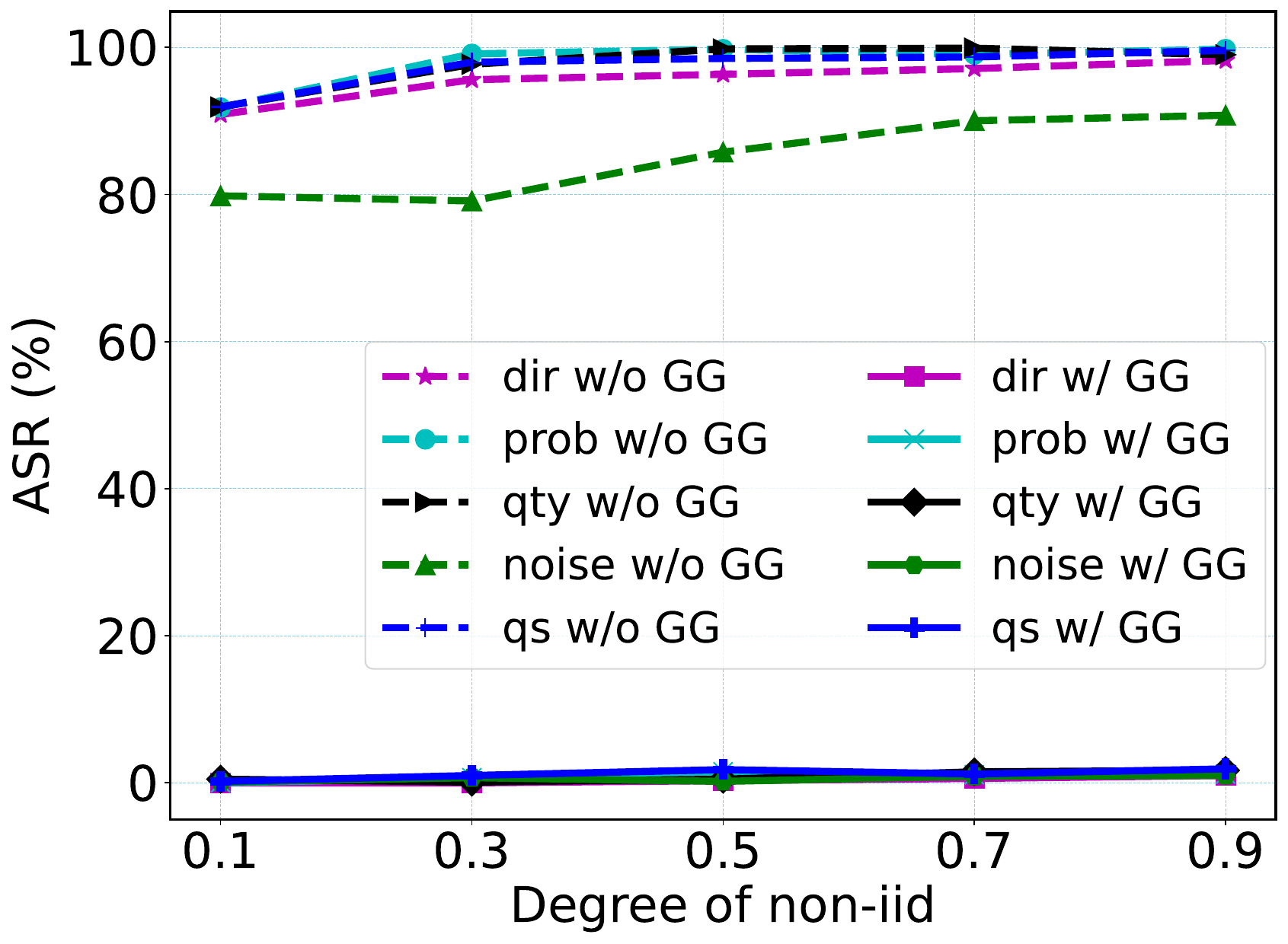}
        \caption{Edge-case}
        \label{fig:noniid_degree_FLARE}
    \end{subfigure}
    \begin{subfigure}[b]{0.19\textwidth}
        \includegraphics[width=\textwidth]{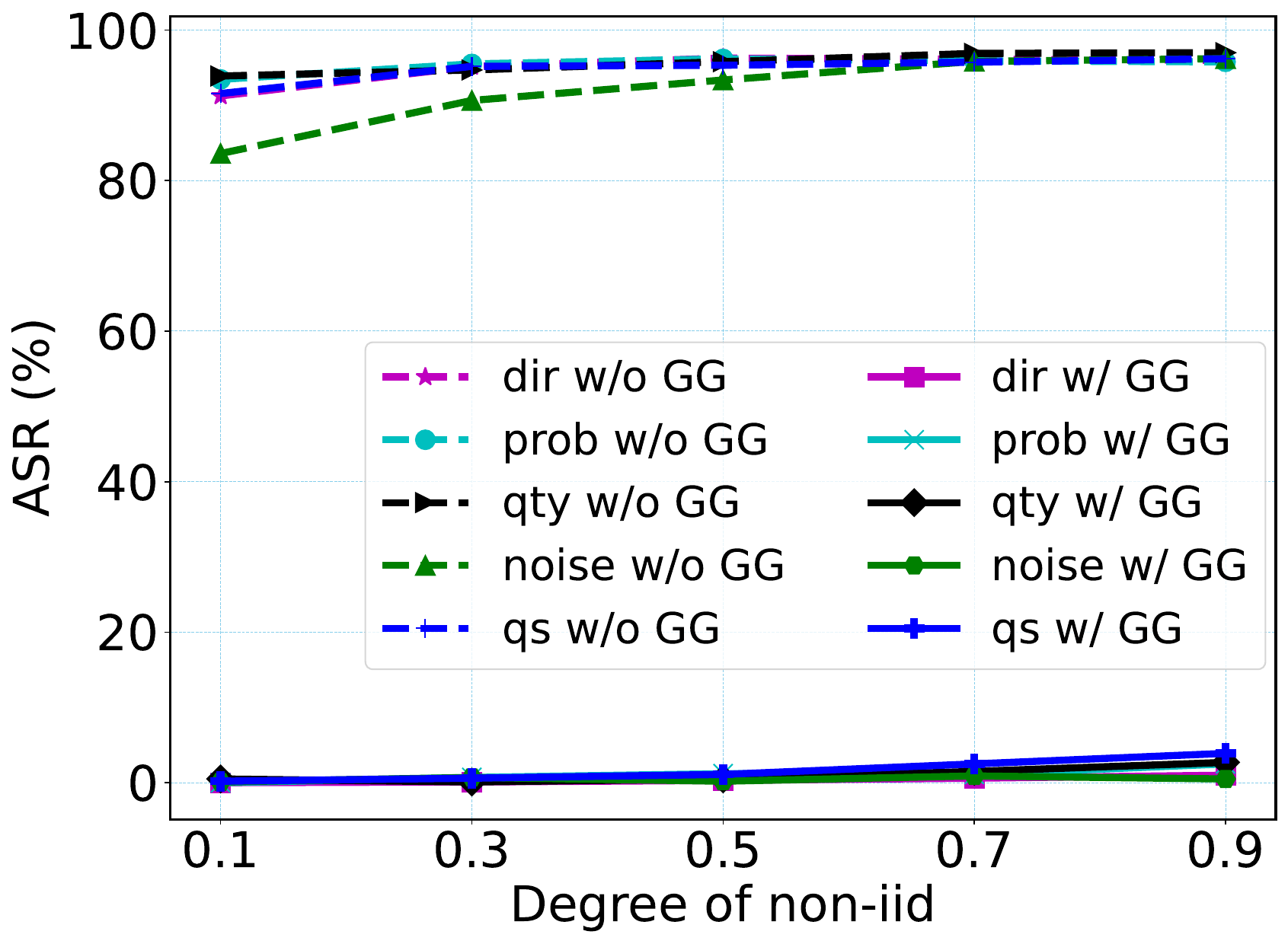}
        \caption{Neurotoxin}
        \label{fig:noniid_degree_FLAME}
    \end{subfigure}
    \begin{subfigure}[b]{0.19\textwidth}
        \includegraphics[width=\textwidth]{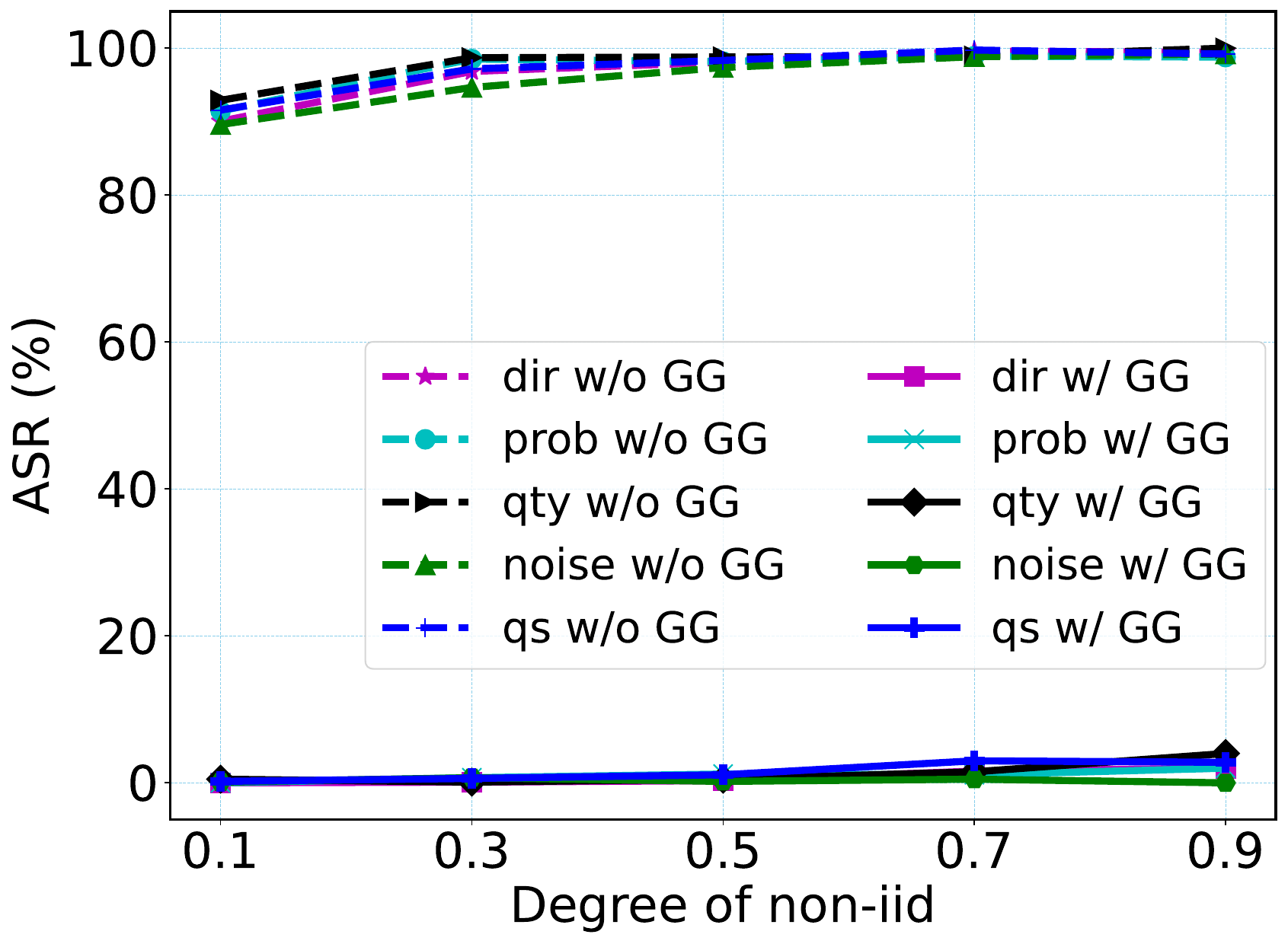}
        \caption{IBA}
        \label{fig:noniid_degree_FLIP}
    \end{subfigure}
    \caption{Impact of the degree of Non-IID on \FG(GG) on CIFAR-10. A higher Non-IID degree indicates a more severe level of Non-IID..}
    \label{fig:noniid-degree-impact}
    \vspace{-5pt}
\end{figure*}

\textbf{Impact of the non-iid degree.}
To facilitate comparison, we normalize all \NIID degrees to a unified scale ranging from 0.1 to 0.9 (only in this section). Intuitively, a large \NIID degree leads to severe expansion of \NIID from IID, therefore increasing the probability that the resulted benign cluster overlaps with the malicious cluster. See Fig.~\ref{fig:intuition-weights} for further details. Fig.~\ref{fig:noniid-degree-impact} illustrates the impact of \NIID degree on \FG, from which we have two observations.
First, the \texttt{ASR} without \FG increases as the \NIID degree rises, indicating that higher \NIID degrees cause more severe harm to the FL model. Second, when \NIID degree increases, it has only a slight influence on the defense performance of \FG, hence rendering \FG a robust defense towards \NIID degree.

\begin{figure*}[ht]
    \centering
    \begin{subfigure}[b]{0.19\textwidth}
        \includegraphics[width=\textwidth]{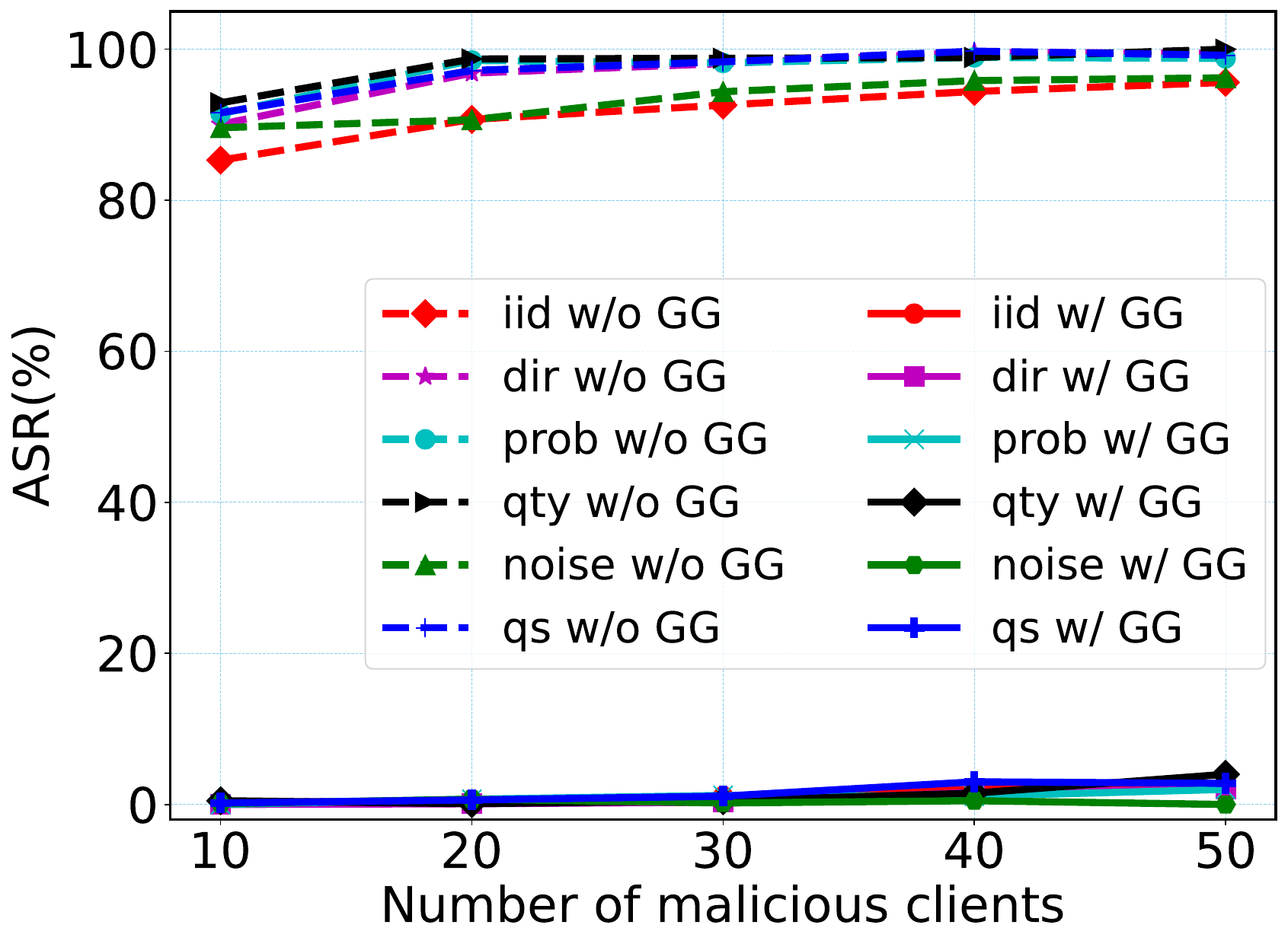}
        \caption{BadNets}
        \label{fig:noniid_degree_FLTrust}
    \end{subfigure}
    \begin{subfigure}[b]{0.19\textwidth}
        \includegraphics[width=\textwidth]{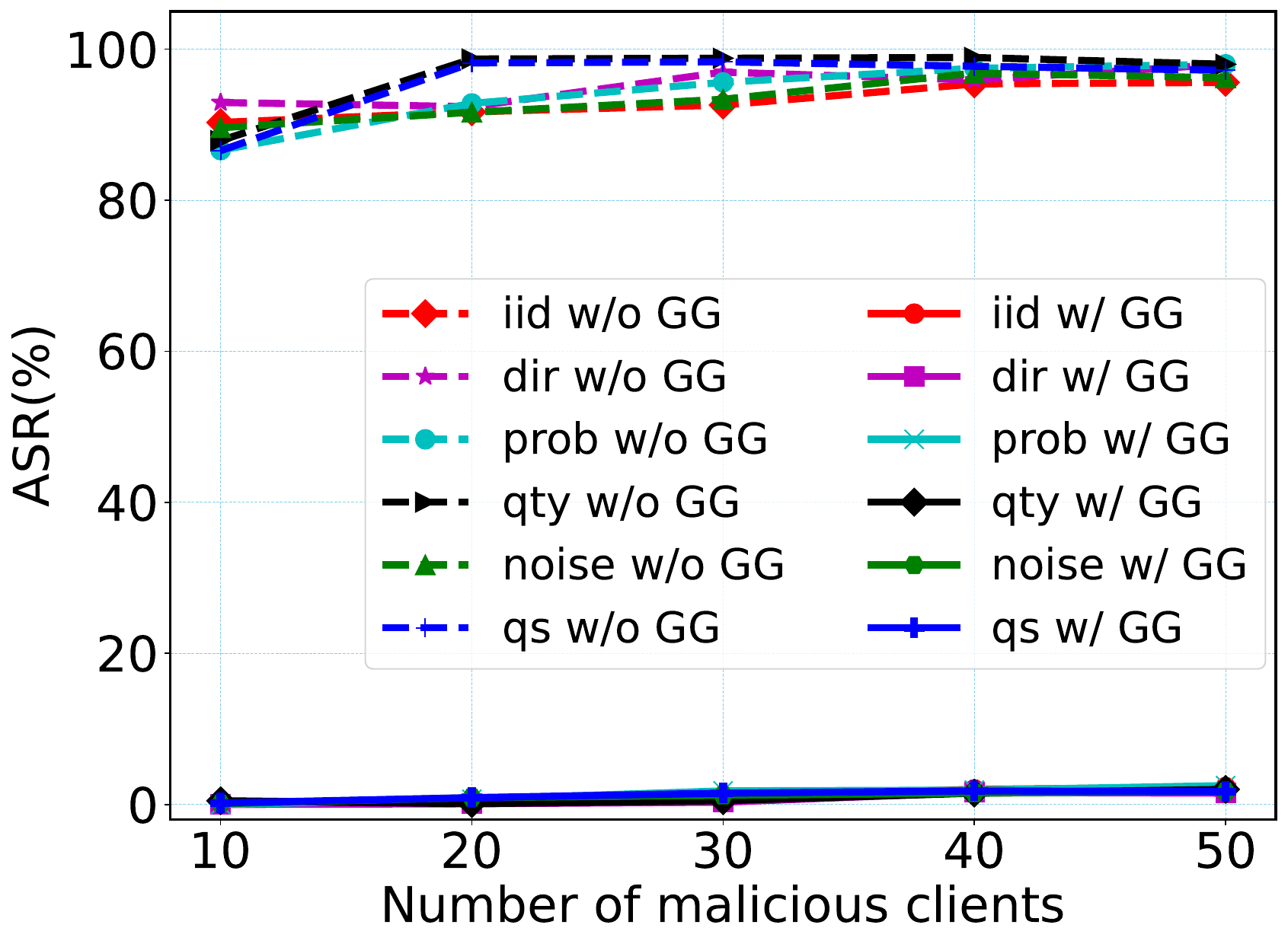}
        \caption{DBA}
        \label{fig:noniid_degree_FLDetector}
    \end{subfigure}
    \begin{subfigure}[b]{0.19\textwidth}
        \includegraphics[width=\textwidth]{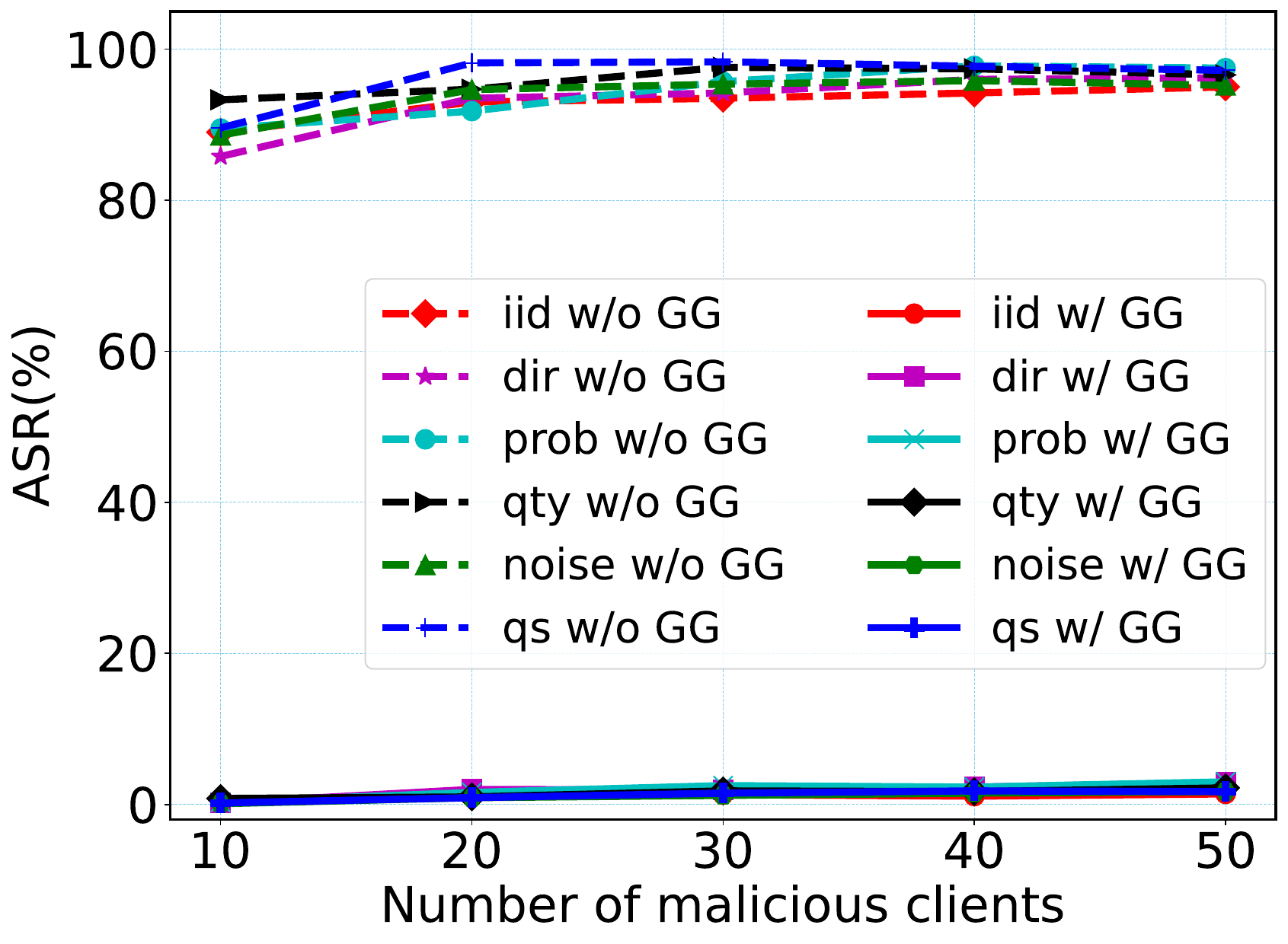}
        \caption{Edge-case}
        \label{fig:noniid_degree_FLARE}
    \end{subfigure}
    \begin{subfigure}[b]{0.19\textwidth}
        \includegraphics[width=\textwidth]{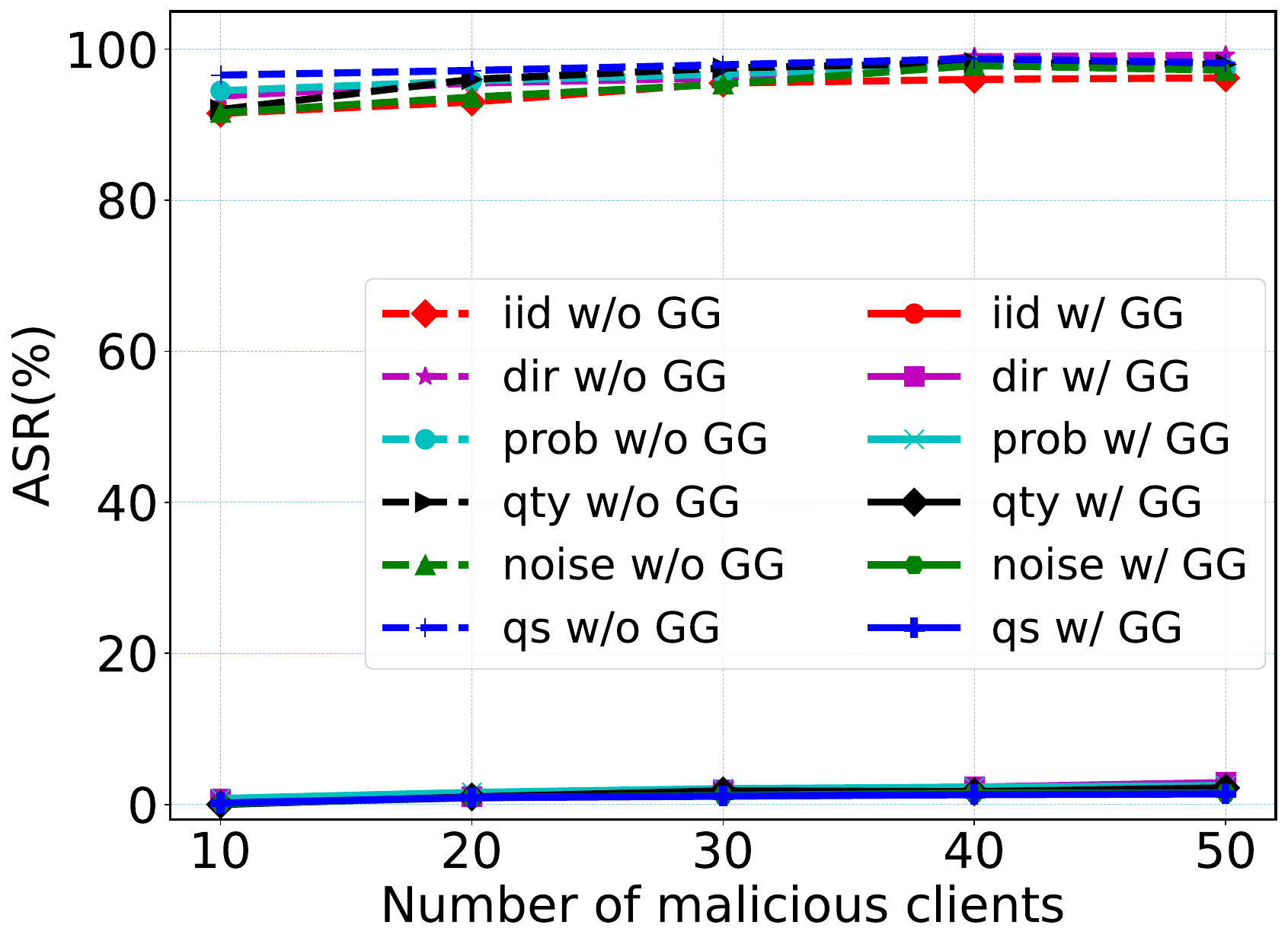}
        \caption{Neurotoxin}
        \label{fig:noniid_degree_FLAME}
    \end{subfigure}
    \begin{subfigure}[b]{0.19\textwidth}
      \includegraphics[width=\textwidth]{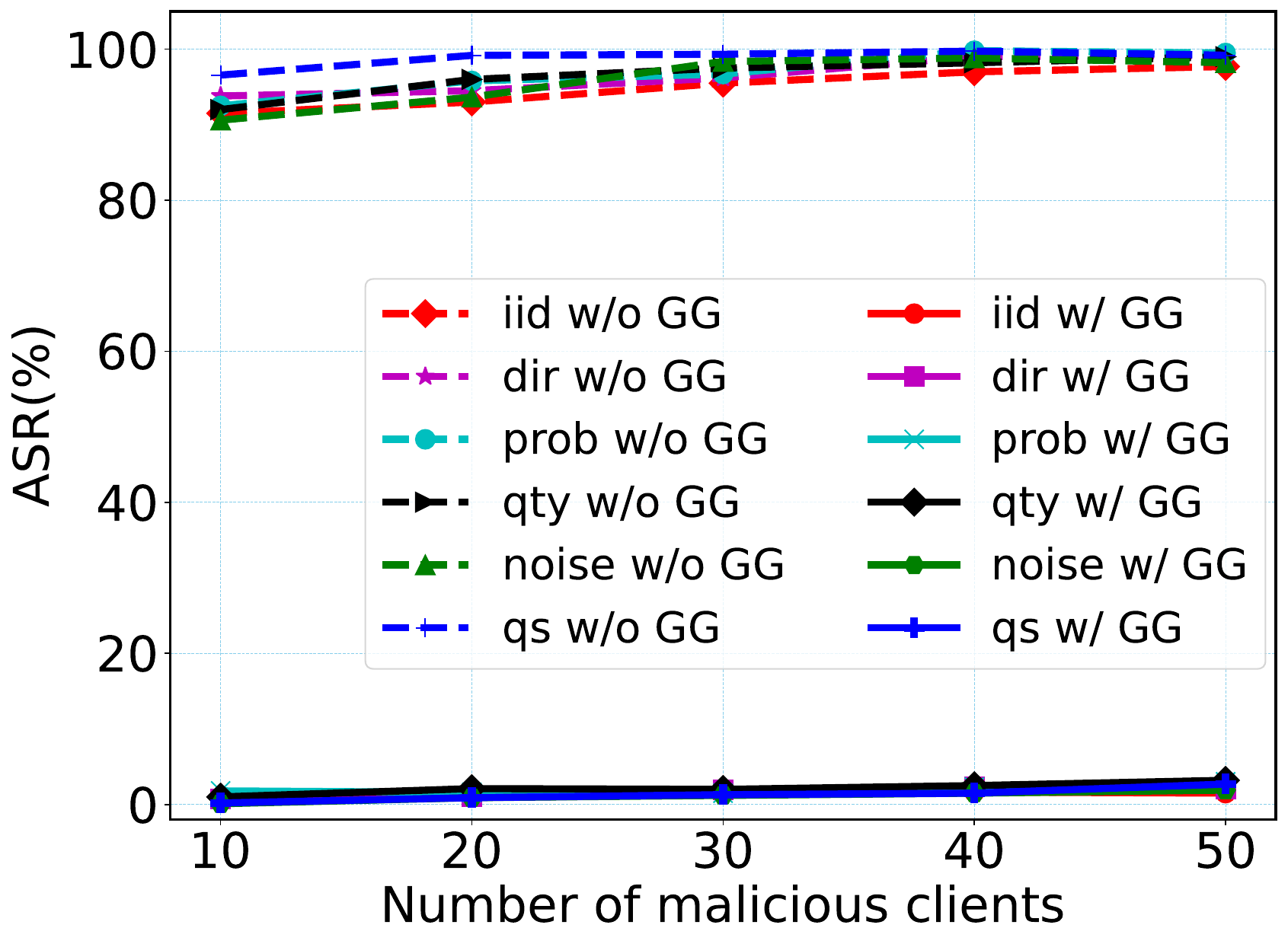}
        \caption{IBA}
        \label{fig:noniid_degree_FLIP}
    \end{subfigure}
    \caption{ Impact of the number of malicious clients on \FG (GG) on CIFAR-10.}
    \label{fig:malicious-impact}
    \vspace{-15pt}
\end{figure*}

\textbf{Impact of the number of malicious clients.}
Intuitively, a larger malicious rate corresponds to a stronger backdoor attack on the chosen FL model. Fig.~\ref{fig:malicious-impact} shows the impact of the number of malicious clients on \FG. First, we observe that malicious models can achieve high \texttt{ASR} without \FG under both IID and \NIID settings, confirming the effectiveness of these backdoors. Secondly, \FG effectively mitigates backdoors, maintaining low \texttt{ASR} values even as the number of malicious clients increases.

\begin{figure}[t]
    \centering
    \begin{subfigure}[b]{0.24\textwidth}
        \includegraphics[width=\textwidth]{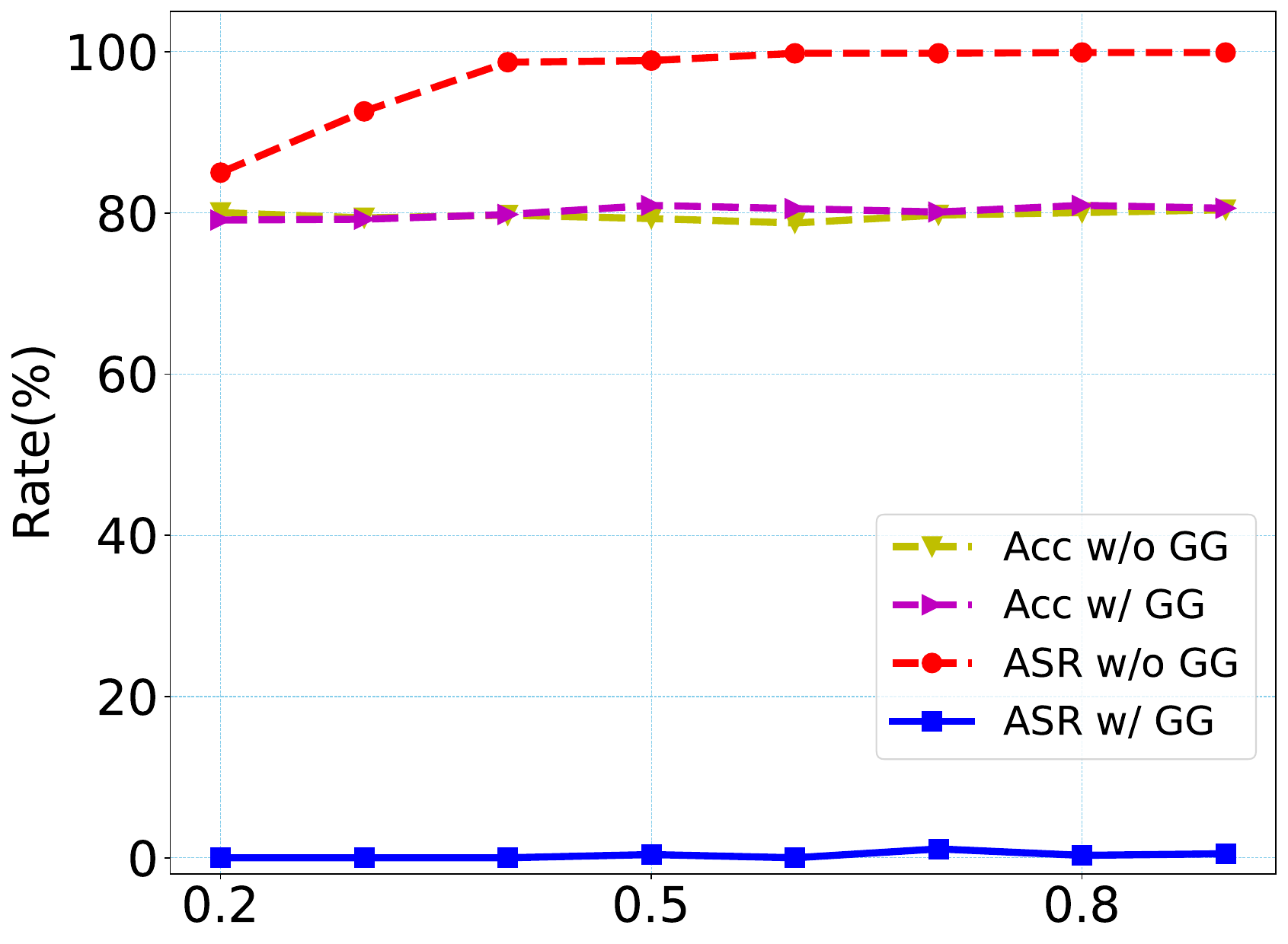}
        \caption{FMNIST}
        \label{fig:data_poisoning_rate_fmnist}
    \end{subfigure}
    \begin{subfigure}[b]{0.24\textwidth}
        \includegraphics[width=\textwidth]{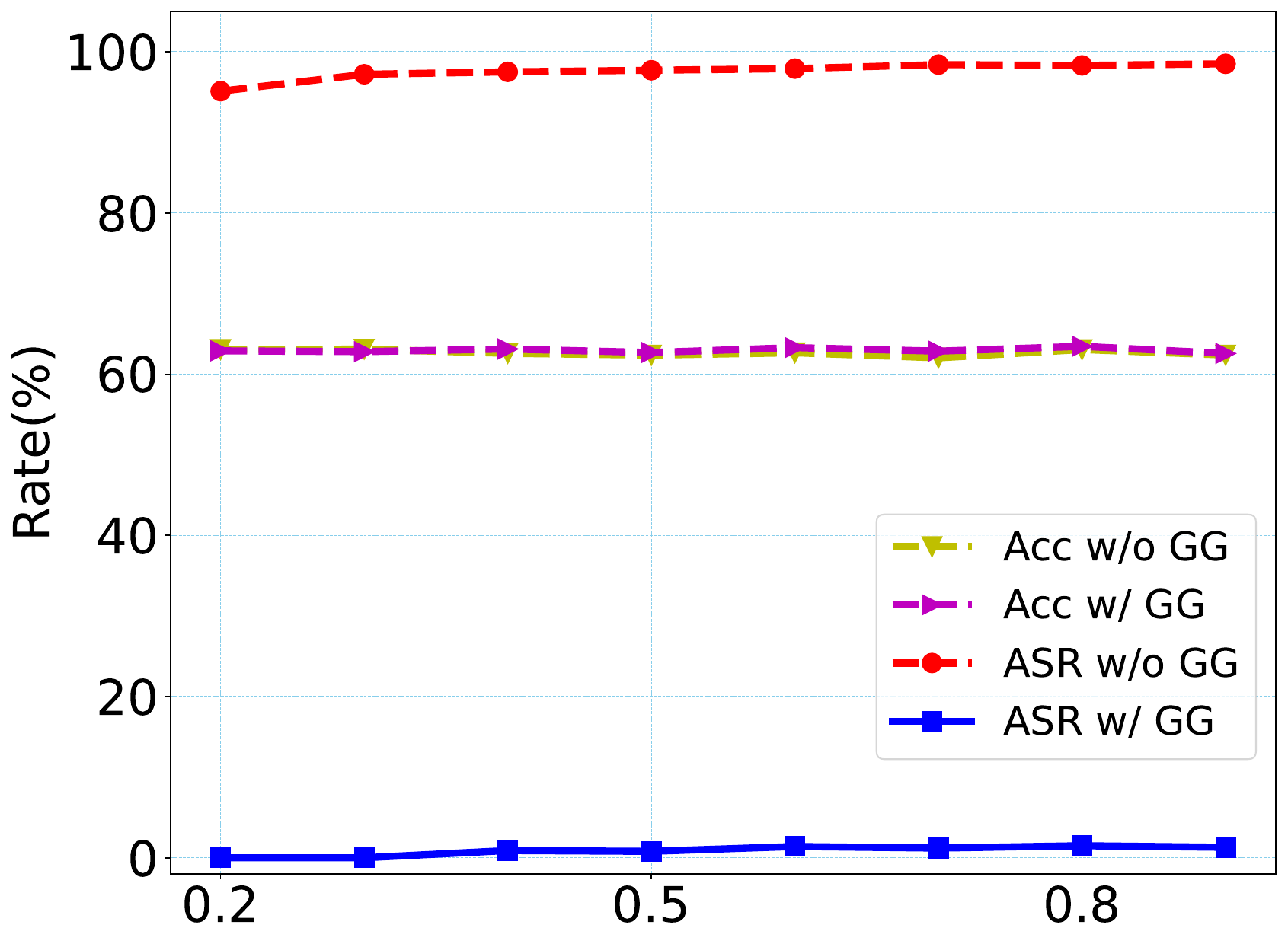}
        \caption{CIFAR-10}
        \label{fig:data_poisoning_rate_cifar}
    \end{subfigure}
    \vspace{-5pt}
 \caption{Impact of data poisoning rate for \FG(GG). \NIID: qty-3 classes; Backdoor Attack: IBA~\cite{nguyen2024iba}.}
    \label{fig:pdr_impact}
\end{figure}

\textbf{Impact of the data poisoning rate.} 
Similarly, a higher data poisoning rate corresponds to a stronger backdoor attack on the targeted FL model. Fig.~\ref{fig:pdr_impact} illustrates the effect of varying data poisoning rates on \FG across two datasets under the recent IBA backdoor attack, where the \NIID setting is configured as qty-3 classes per client. As the poisoning rate increases, the \texttt{ASR} without defense rises, indicating that each malicious model update becomes more impactful on the global FL model. However, this increasing rate has only a minimal effect on \FG, resulting in consistently low \texttt{ASR}s even under severe poisoning scenarios.

\textbf{Impact of the auxiliary data.}
To evaluate the robustness of \FG's validation mechanism, we replaced the original auxiliary datasets with alternative but similar datasets to compare their ASRs. We follow the same method to construct the replaced datasets as Sec.~\ref{subsec:design-flag}.
Specifically, EMNIST~\cite{cohen2017emnist} was used to replace MNIST, KMNIST~\cite{clanuwat2018deep} replaced FMNIST, CIFAR100 replaced CIFAR-10, and IMDB~\cite{maas2011learning} replaced Sentiment-140 to create four dataset pairs. We only  shows the \texttt{ASR}  for these pairs under the IBA attack in Table~\ref{tab:other-auxiliary} due to space limit. 
The results show that using alternative auxiliary datasets leads to only minor changes in \texttt{ASR} values compared to the original datasets. This confirm that the server-side validation mechanism in \FG is robust and does not rely on specific auxiliary datasets.

\begin{table}[!ht]
    \renewcommand{\arraystretch}{1.3}
    \centering
    \scriptsize 
    \resizebox{\columnwidth}{!}{ 
    \begin{tabular}{p{0.8cm}|p{1.45cm}|p{0.5cm}p{0.5cm}p{0.5cm}p{0.5cm}p{0.5cm}p{0.5cm}}
    \hline
    \textbf{\#Seq} & \textbf{TrainSet \& Auxiliary-Set} & \textbf{iid} & \textbf{dir} & \textbf{prob} & \textbf{qty} & \textbf{noise} & \textbf{qs} \\ 
    \hline
    \multirow{2}{*}{Pair1} & MNIST & 0.10 & 0.31 & 0.35 & 0.19 & 0.70 & 0.50 \\
                           & EMNIST & 0.13 & 0.28 & 0.40 & 0.26 & 0.84 & 0.38 \\
    \hline
    \multirow{2}{*}{Pair2} & FMNIST & 0.09 & 0.50& 0.58 & 0.36 & 0.40 & 0.73 \\
                           & KMNIST & 0.15 & 0.48 & 0.60 & 0.42 & 0.52 & 1.00 \\
    \hline
    \multirow{2}{*}{Pair3} & CIFAR-10 & 0.10 & 0.30 & 0.52 & 0.22 & 0.55 & 0.60 \\
                           & CIFAR100  & 0.25 & 0.40 & 0.68 & 0.47  &  0.70  & 0.91 \\
    \hline
    \multirow{2}{*}{Pair4} & Sentiment-140 & 0.29 & 0.80 & 0.85 & 0.60 & 0.90 & 1.05 \\
                           & IMDB & 0.40 & 1.12 & 0.90 & 0.68 & 1.10 & 1.20 \\
    \hline
    \end{tabular}
    }
    \caption{Impact of the auxiliary datasets on \FG, IBA attack.}
    \label{tab:other-auxiliary}
    \vspace{-10pt}
\end{table}

\textbf{Impact of the choices of metrics.}
Table~\ref{tab:impact_metrics} shows the impact of metric choices in model-weight module on \FG. Due to space limit, we only present the defense results for the most recent IBA attack. Clustering on model weights achieves the lowest ASRs, while using only cosine similarity or Euclidean distance performs worse. This highlights the need for a comprehensive approach combining direction and magnitude, especially in \NIIDs where ASRs are much higher than in iid.

\begin{table}[!ht]
    \renewcommand{\arraystretch}{1.3}
    \centering
    \footnotesize
    \caption{Impact of the choices of metrics on GeminiGuard, CIFAR-10, IBA Attack}
    \label{tab:impact_metrics}
    \begin{tabular}{p{2.2cm}|p{0.5cm}p{0.5cm}p{0.5cm}p{0.5cm}p{0.5cm}p{0.5cm}}
        \hline
         \centering
      \textbf{Metrics} & \textbf{iid} & \textbf{dir} & \textbf{prob} & \textbf{qty} & \textbf{noise} & \textbf{qs} \\ 
        \hline
         \centering
     Model weights & 0.73 & 0.95 & 0.80 & 1.62 & 0.70 & 1.23  \\
         \centering
    Cosine similarity  & 1.00 & 2.70 & 0.90 & 3.02 & 1.00 & 2.35 \\
         \centering
    Euclidean distance  & 1.60 & 2.20 & 1.70 & 4.95 & 1.90 & 1.95  \\
        \hline
    \end{tabular}
\end{table}


\textbf{Impact of the number of layers.} 
\label{sec:evaluation:layers}
We evaluate the effect of using 1 layer (penultimate layer, PLR), 2 layers (PLR and the second-to-last layer), and 3 layers (PLR and the last two layers) in the latent-space module of \FG. Due to space limit, we only present the defense results for the most recent IBA attack in Table~\ref{tab:impact_layers}. The results show that using only one layer leads to a higher ASR compared to two or three ones, indicating the need to consider multiple layers to mitigate attacks from unconsidered layers.



\begin{table}[!ht]
    \renewcommand{\arraystretch}{1.3}
    \centering
    \footnotesize
    \caption{Impact of the number of layers on GeminiGuard, CIFAR-10, IBA Attack}
    \label{tab:impact_layers}
    \begin{tabular}{p{1.5cm}|p{0.6cm}p{0.6cm}p{0.6cm}p{0.6cm}p{0.6cm}p{0.6cm}}
        \hline
         \centering
      \textbf{\#Layers} & \textbf{iid} & \textbf{dir} & \textbf{prob} & \textbf{qty} & \textbf{noise} & \textbf{qs} \\ 
        \hline
        \centering
    1  & 1.30 & 1.60 & 1.83 & 2.17 & 1.62 & 1.50  \\
         \centering
    2 & 1.10 & 1.20 & 1.31 & 1.58 & 0.95 & 1.10 \\
         \centering
    3  & 0.50 & 0.73 & 0.68 & 0.80 & 0.60 & 0.87  \\
        \hline
    \end{tabular}
    \vspace{-15pt}
\end{table}


\subsection{Computation time analysis of \FG} 

Table~\ref{tab:time-memory} compares the average computation time (in seconds) of GeminiGuard and FedAvg~\cite{mcmahan2017communication} on FMNIST and CIFAR-10 across various non-iids. While GeminiGuard incurs slightly higher computation time due to clustering and trust score calculations, the overhead remains minimal compared to FedAvg. This highlights GeminiGuard as a lightweight and efficient solution for real-world deployments.

\begin{table}[!ht]
    \renewcommand{\arraystretch}{1.3}
    \centering
    \caption{Computation Time (s) -- GeminiGuard vs. FedAvg}
    \scriptsize 
    \resizebox{\columnwidth}{!}{ 
    \begin{tabular}{p{1.3cm}|p{1.10cm}|p{0.55cm}p{0.55cm}p{0.55cm}p{0.55cm}p{0.55cm}p{0.55cm}}
    \hline
    \textbf{Defense} & \textbf{Dataset} & \textbf{iid} & \textbf{dir} & \textbf{prob} & \textbf{qty} & \textbf{noise} & \textbf{qs} \\ 
    \hline
    \multirow{2}{*}{FedAvg} & FMNIST & 0.0008 &  0.0024  & 0.0009 & 0.0027 & 0.0008 & 0.0011 \\
                           & CIFAR-10 & 0.0112  & 0.0106 & 0.0108 & 0.0124  & 0.0105 & 0.0193 \\
    \hline
    \multirow{2}{*}{GeminiGuard} & FMNIST  & 0.0028 & 0.0038 & 0.0022 & 0.0045  & 0.0024 & 0.0032 \\
                           & CIFAR-10 & 0.0351 & 0.0299 & 0.0309 & 0.0362 & 0.0360 & 0.0242 \\
    \hline
    \end{tabular}
    }
    \label{tab:time-memory}
    \vspace{-15pt}
\end{table}

\subsection{Resilience to Adaptive Attacks under \NIIDs} 
As in prior defenses, we expect that resilience to adaptive attacks is critical for a defense. When it comes to backdoors, given knowledge about \FG, a \mpg attacker's goal is to adaptively change the attack strategy for bypassing \FG while successfully attacking the victim FL model. 

We design an adaptive attack and enhance it by (1) including five \NIIDs that are not evaluated in previous work, and (2) optimizing the malicious updates. In fact, \NIID settings are necessary for evaluating defenses against adaptive attacks because \NIIDs effectively facilitate adaptive attacks, i.e., reducing distinguishability between benign and malicious updates. To generate optimal malicious updates, we include multiple attack objectives and normalize the loss terms, which will be discussed in detail below.

The adaptive attacker crafts her malicious model updates and corresponding layers to resemble those of benign model updates. As a result, the formulation of such an adaptive attack is to minimize a combination of loss from (1) targeted inputs and clean data, (2) the distance between a crafted malicious update and average benign ones, and (3) the distance between the hidden layers of a crafted malicious update and that of average benign model updates. In the end, we formulate the adaptive attack as follows: 

\begin{equation}
    \begin{aligned}
        \arg\min_{\delta_{mal}} \quad L(\mathcal{D}_{mal}) + \lambda L(\mathcal{D}_{train}) + \rho \Tilde{\Delta d_{\delta}} + \eta \Tilde{d_{layers}},
    \end{aligned}
    \label{eq:adaptive-attack}
\end{equation}

\noindent where $L(\mathcal{D}_{mal})$ denotes the loss on targeted inputs, $L(\mathcal{D}_{train})$ the loss on clean dataset, $\Tilde{\Delta d_{\delta}} = \frac{\lVert \delta_{mal}-\bar{\delta}_{ben}\rVert}{\lVert \delta_{mal}-\bar{\delta}_{ben}\rVert_{max}}$ the normalized distance between malicious model updates and average benign ones, and finally \(\Tilde{d_{layer}}=\frac{d_{layer}}{d_{layer,max}}\) the normalized distance between the layers of malicious model updates and the average one of benign models. $\lambda$, $\rho$, and $\eta$ are the corresponding coefficients for different loss terms we obtain from the greedy search. 

\begin{figure}[ht]
    \centering
    \begin{subfigure}[b]{0.24\textwidth}
        \includegraphics[width=\textwidth]{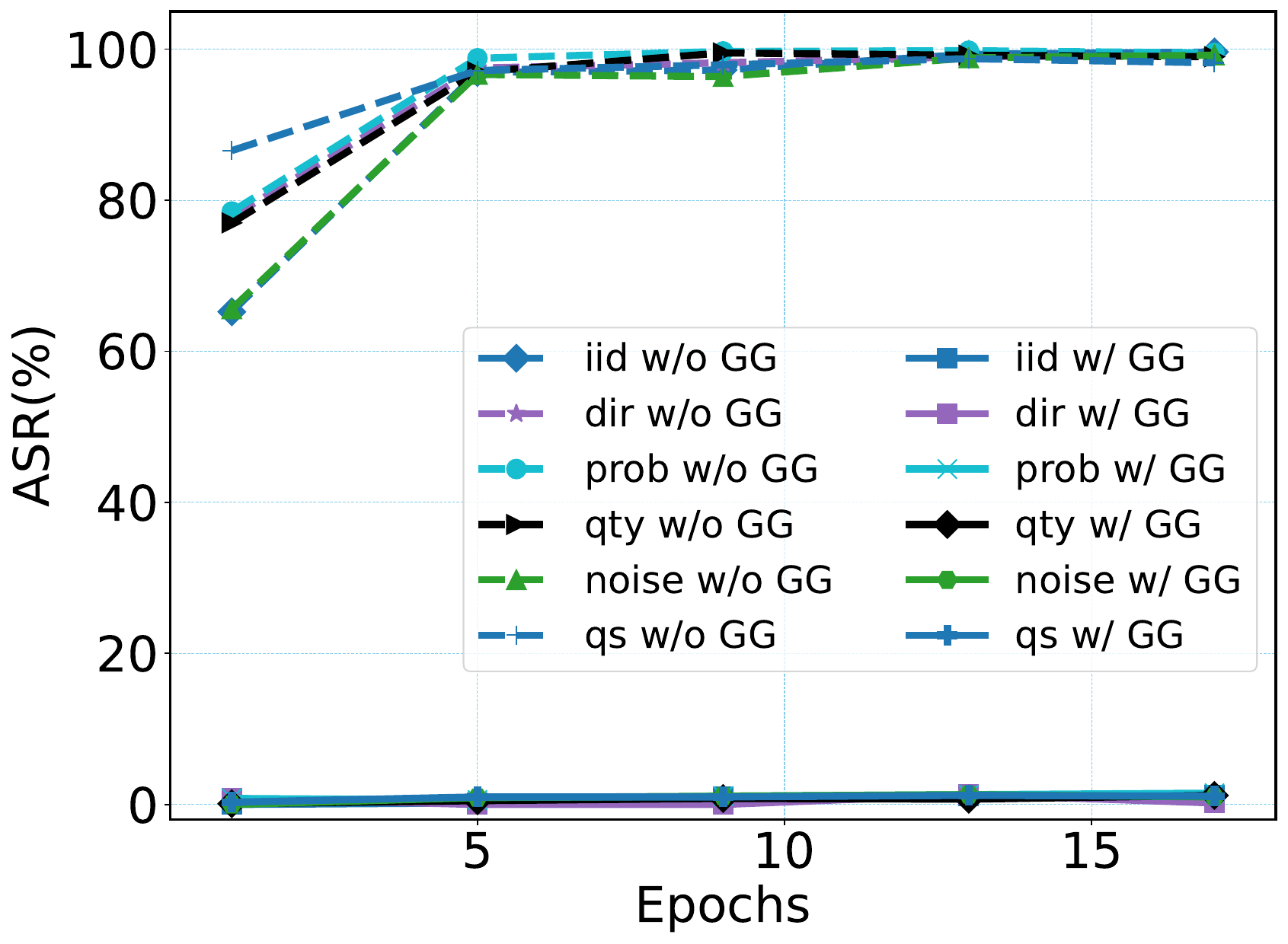}
        \caption{MNIST}
        \label{fig:adaptive_mnist}
    \end{subfigure}
    \begin{subfigure}[b]{0.24\textwidth}
        \includegraphics[width=\textwidth]{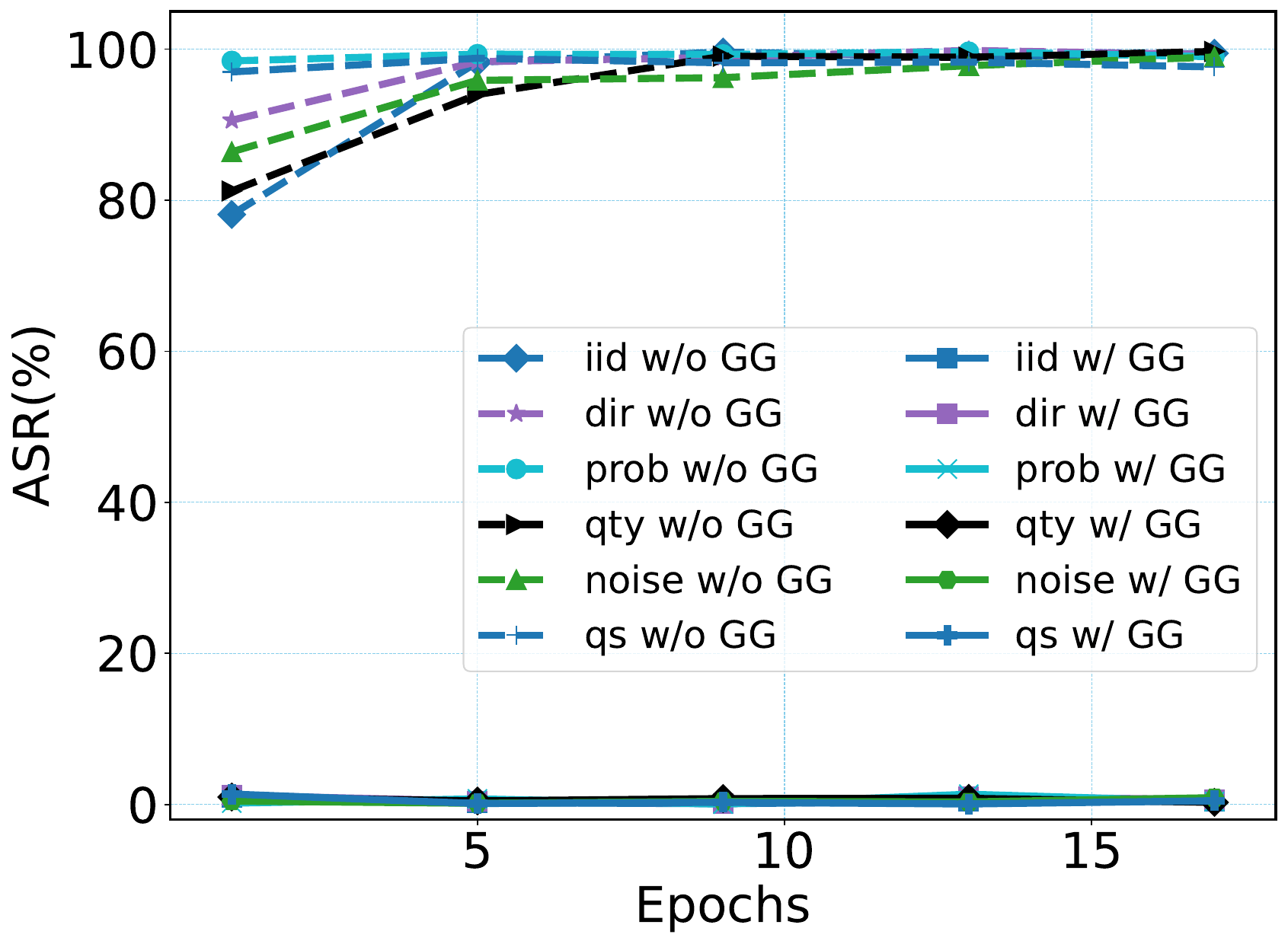}
        \caption{FMNIST}
        \label{fig:adaptive_fmnist}
    \end{subfigure}
    \begin{subfigure}[b]{0.24\textwidth}
        \includegraphics[width=\textwidth]{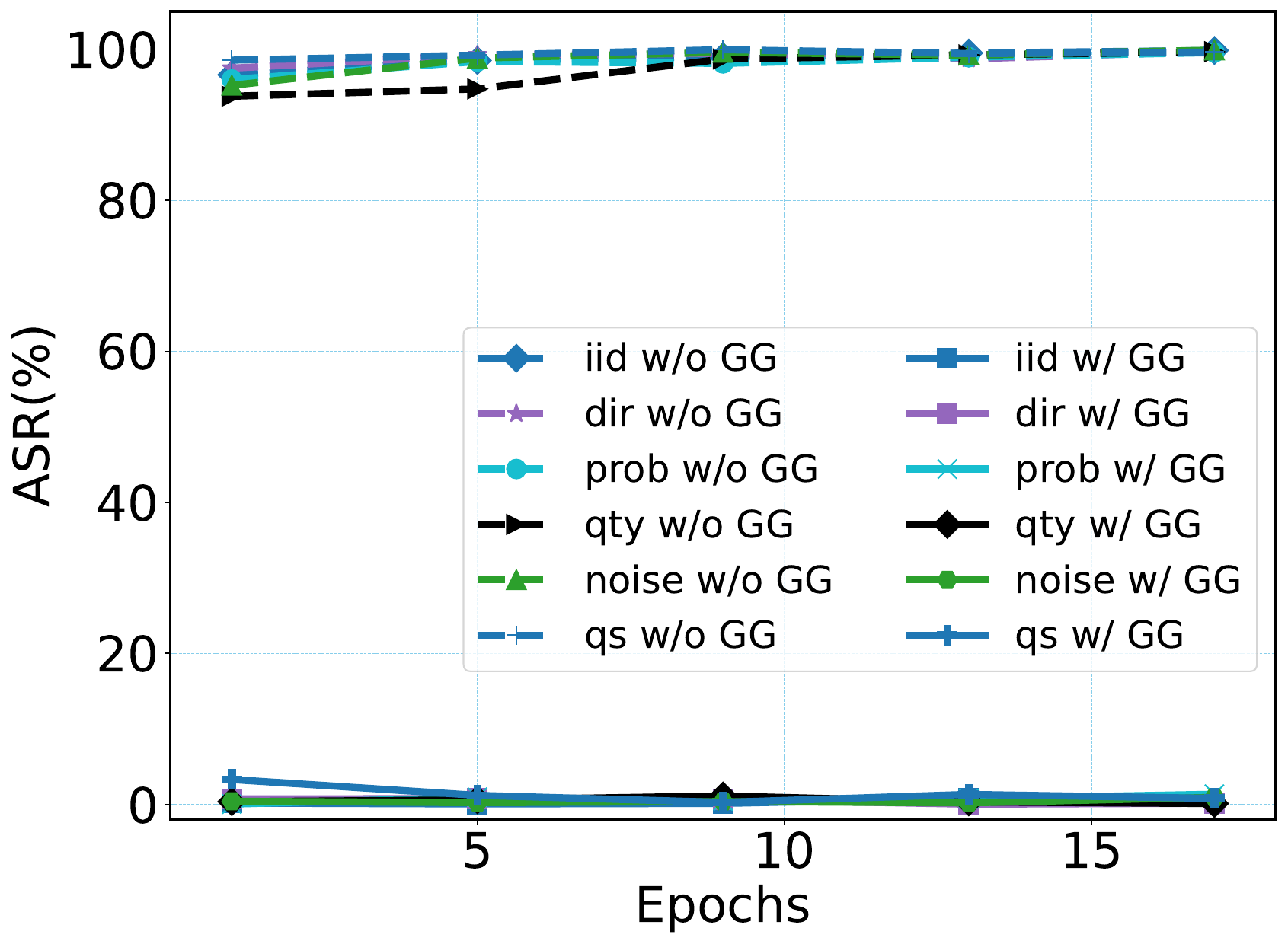}
        \caption{CIFAR-10}
        \label{fig:adaptive_cifar}
    \end{subfigure}
    \begin{subfigure}[b]{0.24\textwidth}
        \includegraphics[width=\textwidth]{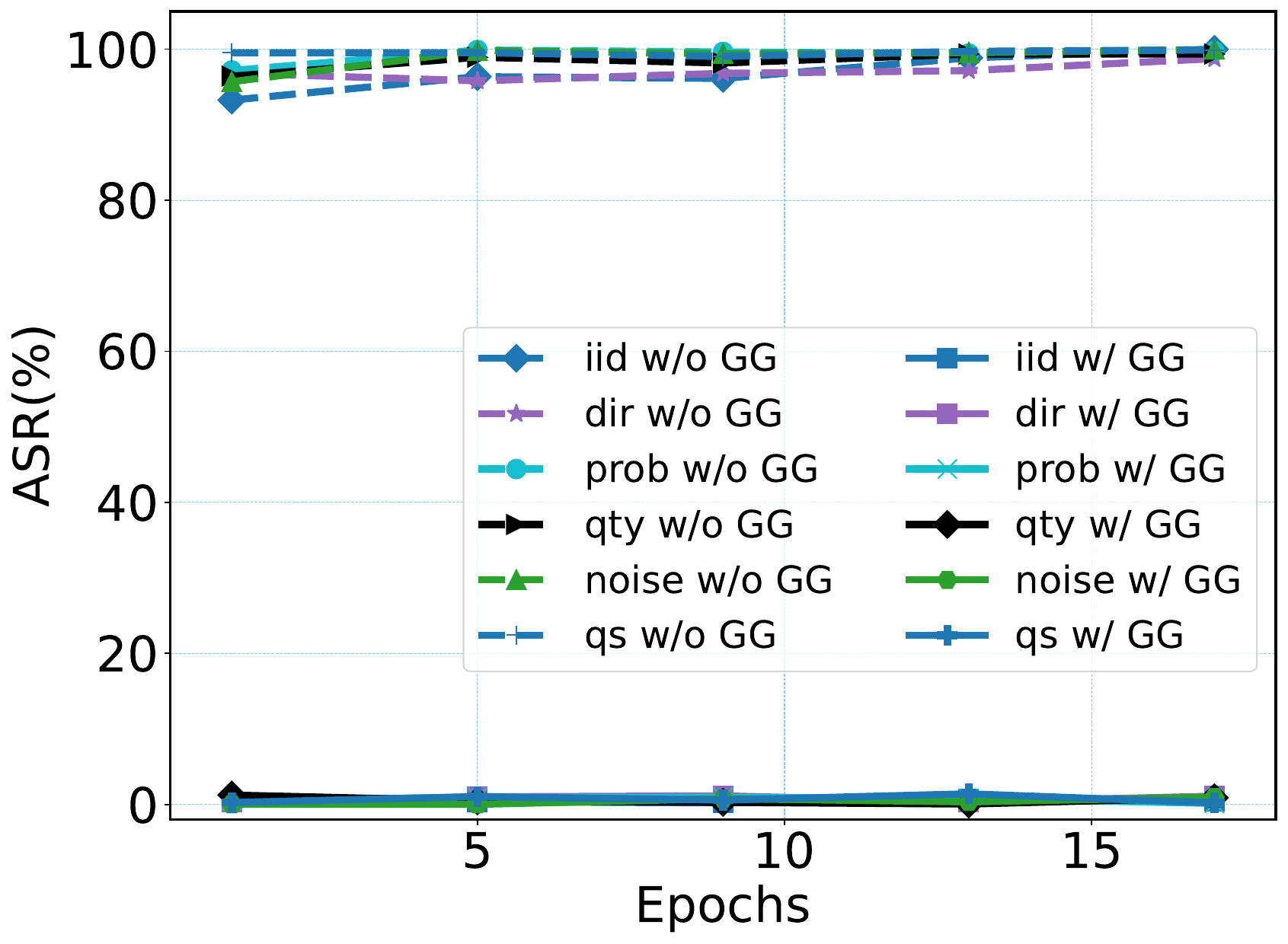}
        \caption{Sentiment-140}
        \label{fig:adaptive_sentiment}
    \end{subfigure}
    \caption{Performance of \FG(GG) against adaptive attacks.}
    \label{fig:flag_adaptive_performance}
    \vspace{-10pt}
\end{figure}

We use two \texttt{ASRs} to evaluate the performance of \FG against adaptive attacks---\texttt{ASR} without \FG and \texttt{ASR} with \FG. Obviously, the larger the differences between the two, the more resilient \FG is towards adaptive attacks. 
Fig.~\ref{fig:flag_adaptive_performance} confirms that \FG works well against adaptive attacks across various datasets under \NIIDs. In this figure, we can see that the adaptive attack achieves a high \texttt{ASR} without \FG, meaning that the proposed adaptive attack succeeds in attacking the victim FL model. When \FG is deployed, the performance of the adaptive attack drops significantly. Such results confirm that it is challenging for adaptive attacks to simultaneously achieve their malicious objectives and bypass \FG.

%% file: 6-related.tex
\section{Related Work}
\label{sec:related}

\subsection{\MPAs}
Recent years have witnessed the development of various novel model poisoning attacks (MPAs) in FL, each characterized by unique strategies to manipulate local model updates. These attacks are broadly classified into untargeted and targeted (backdoor) attacks. Untargeted MPAs aim to disrupt global model performance by techniques like crafting malicious updates in the opposite direction of benign ones, as explored in Krum-Attack~\cite{fang2020local} and Trimmean-Attack~\cite{fang2020local}, or by optimizing specific objectives, such as in Min-Max~\cite{shejwalkar2021manipulating} and Min-Sum attacks~\cite{shejwalkar2021manipulating}.
More recently, several MPAs explored backdoors on FL, which were first introduced in BadNet~\cite{gu2017badnets} and used an embedded trigger to enforce specific model output. Xie et al.~\cite{xie2019dba} proposed DBA, extending the backdoor attack to distributed settings. Edge-case~\cite{wang2020attack} targets tail-end inputs in the data distribution to evade detection. Neurotoxin~\cite{zhang2022neurotoxin} exploits parameters with minimal updates during training to maintain backdoor persistence. IBA~\cite{nguyen2024iba} advances further by embedding stealthy and irreversible backdoors, offering strong resistance to existing defenses.






\subsection{Defenses against \MPAs}


The efforts to defend against them mainly focus on two directions: (i) robust aggregation and (ii) detection.

\textbf{Aggregation-based Defenses.} The early explorations along this direction were referred to as byzantine resilient aggregation. Among them, Krum~\cite{blanchard2017machine} proposed to select the local model most similar to others as the global model to reduce the impact of outliers. Trimmed Mean~\cite{yin2018byzantine} used a coordinate-wise aggregation rule to evaluate and aggregate each model parameter independently. However, these methods have been shown to be ineffective for backdoor attacks in~\cite{fang2020local,wang2022flare}, not to mention further challenges under \NIIDs.

Recent aggregation-based defenses have shown better performance on strong MPAs, particularly in stealthy backdoor attacks. Specifically, FLTrust in~\cite{cao2021fltrust} proposed to compute a trust score for a submitted model update with a joint assessment of direction similarity and magnitude and use the score for model aggregation. FLARE~\cite{wang2022flare} instead focused on exploiting penultimate layer representations (PLRs) for better characterizing malicious model updates and then calculating trust scores for model update aggregation. The main limitation of these defenses is that they adopted none or at most one \NIID setting, which is unlikely to hold in practice.


\textbf{Detection-based Defenses.} 
FLAME~\cite{nguyen2022flame} used clustering to filter out malicious model updates and applied the minimal amount of noise necessary to eliminate backdoors. Zhang et al.~\cite{zhang2023flipprovabledefenseframework} proposed FLIP, which focuses on reverse engineering backdoor triggers to enhance defense.
FLDetector~\cite{zhangfldetecotr} detected and filtered malicious model updates using historical data and consistency checks for identifying malicious ones. CrowdGuard~\cite{CrowdGuard}, MESAS~\cite{MESAS}, FreqFed~\cite{fereidooni2023freqfed}, FLGuard~\cite{lee2023flguard}, FLShield~\cite{kabir2024flshield} and AGSD~\cite{ali2024adversarially} are the few most recent defenses aiming to address \NIID challenges explicitly. CrowdGuard involved clustering for hidden layer features of model updates for attack detection. MESAS used a multi-faceted detection strategy, statistical tests, clustering, pruning, and filtering for backdoor detection. FreqFed tackled backdoor attacks in a unique way where it transforms model updates into the frequency domain and identify unique components from malicious model updates. FLGuard~\cite{lee2023flguard} used contrastive learning to detect and remove malicious updates. FLShield~\cite{kabir2024flshield} proposed a novel validation method using benign participant data to defend against model poisoning attacks. AGSD~\cite{ali2024adversarially} utilized a trust index history with FGSM optimization to resist backdoor attacks.
Aside from specific assumptions on hardware (\eg, TEE requirement in CrowdGuard), additional training (\eg, contrastive learning in FLGuard and FGSM training in AGSD), and specific strategies (\eg, FreqFed assumed backdoor attacks exhibit unique patterns in the frequency domain. FLShield assumes that validators are selected from the local clients pool to perform the validation), these methods still lack comprehensiveness regarding \NIIDs, as shown in Table~\ref{table:flag-position}.

\textbf{Differences in \FG.} The main differences in \FG when compared to the above defenses consist of the following. Firstly, \FG considered the most comprehensive and important set of \NIIDs. Secondly, the design of \FG is an unsupervised and easy to employed in real-world scenarios. Thirdly, \FG considered a comprehensive set of adaptive attacks as well. Finally, our evaluations for \FG consisted of a comprehensive combination of datasets, attacks, baselines, and ML tasks. The details are summarized in Table~\ref{table:flag-position}.

%% file: 7-conclusion.tex
\section{Conclusions}
\label{sec:conclude}

In this paper, we introduce \FG, a lightweight, versatile, and unsupervised framework designed to defend Federated Learning (FL)  against model poisoning attacks across five challenging \NIID settings. Unlike existing methods that are often limited to iid  or only specific \NIID scenarios, \FG bridges this gap by employing model-weight analysis, coupled with a novel latent-space analysis module to effectively distinguish between benign and malicious updates.
Through comprehensive experiments, \FG demonstrates exceptional defense performance against both untargeted and targeted (backdoor) attacks, including adaptive strategies, across diverse datasets and FL tasks. 
Compared to nine state-of-the-art defenses, \FG consistently achieves superior robustness, especially in non-iid scenarios. These findings highlight \FG as a reliable and practical solution for mitigating \mpas in realistic \NIID settings.